# Yanyun-3: Enabling Cross-Platform Strategy Game Operation with Vision–Language Models

Guoyan Wang[1], Yanyan Huang[†1], Chunlin Chen[2], Lifeng Wang[3], Yuxiang Sun[†2]

(1. School of Automation, Nanjing University of Science and Technology, Nanjing, Jiangsu 210094, China; 2. School of Robotics and Automation, Nanjing University, Suzhou, Jiangsu 215163, China; 3. Sichuan Teng dun Liangyuan Intelligent Technology Co., Ltd., Chengdu, Sichuan 610037, China)

**Abstract**: Cross-platform strategy game automation remains a challenge due to diverse user interfaces and dynamic battlefield environments. Existing Vision–Language Models (VLMs) struggle with generalization across heterogeneous platforms and lack precision in interface understanding and action execution. We introduce Yanyun-3, a VLM-based agent that integrates Qwen2.5-VL for visual reasoning and UI-TARS for interface execution. We propose a novel data organization principle—*combination granularity*—to distinguish intra-sample fusion and inter-sample mixing of multimodal data (static images, multi-image sequences, and videos). The model is fine-tuned using QLoRA on a curated dataset across three strategy game platforms. The optimal strategy (M*V+S) achieves a **12.98× improvement** in BLEU-4 score and a ***63% reduction*** in inference time compared to full fusion. Yanyun-3 successfully executes core tasks (e.g., target selection, resource allocation) across platforms without platform-specific tuning. Our findings demonstrate that structured multimodal data organization significantly enhances VLM performance in embodied tasks. Yanyun-3 offers a generalizable framework for GUI automation, with broader implications for robotics and autonomous systems.

# Introduction

Advances in large multimodal models have spurred the application of vision–language collaborative reasoning in domains such as visual question answering (VQA) and robotic interaction [1–7]. However, when tasked with high-precision interactive operations within complex and variable user interfaces, these models still suffer from insufficient generalization—particularly in the dynamic battlefield environments of cross-platform strategy game systems, where they must interpret diverse interface elements (e.g., maps, weapons, target buttons) and execute precise actions across heterogeneous platforms.

Strategy gaming serves as a pivotal technical methodology for strategic analysis and validation, playing a key role in training simulation, plan demonstration, and effectiveness evaluation [8]. From early board strategy games to modern computer-based systems, its core objective has remained the simulation of complex scenarios to support decision optimization and action planning [9].

Modern digital strategy game systems integrate maps, unit deployment, toolkits, and real-time battlefield updates to simulate multi-branch coordination, multi-phase operations, and dynamic battlefield evolution [10]. Users must make high-precision decisions under time pressure based on the screen state and execute them via a sequence of mouse or keyboard commands—an ability that places extreme demands on interface comprehension, continuous action planning, and cross-platform adaptability [11,12]. Current AI approaches for strategy games can be broadly categorized into three types: rule-driven AI, which relies on predefined rules and is effective in fixed environments but lacks adaptability [13]; reinforcement learning (RL) AI, which can learn complex strategies through prolonged interaction yet exhibits limited cross-platform generalization [14,15]; and large language model (LLM)-based AI, which demonstrates considerable promise in strategic analysis and contextual reasoning [16,17]. Indeed, LLM-based agents have recently achieved win rates exceeding 80% in platoon-level confrontations, outperforming both traditional RL and rule-based counterparts and signaling rapid maturation. Nevertheless, they have not yet effectively tackled core challenges inherent to strategy games, such as asymmetric force allocation, high-stochasticity adjudication mechanisms, and multi-agent asynchronous collaboration [18]. In recent years, the convergence of visual perception and language reasoning has positioned vision–language models (VLMs) as a leading candidate for achieving cross-platform strategy game automation [19–21], although a performance gap relative to human experts remains [22].

However, in real-world strategy game tasks, VLMs still confront several prominent challenges: insufficient cross-platform UI understanding [23], primarily stemming from substantial differences in layout, appearance, and interaction logic across platforms; vulnerability of multi-step reasoning to interface update latency [24,25]; a tension between generalization and knowledge retention; and low precision in recognizing subtle interface actions [26,27].

To address these challenges, we introduce a novel data organization paradigm—*Combination Granularity* [28]—and construct a multimodal training set that integrates high signal-to-noise-ratio static snapshots with structured dynamic sequences to unlock the potential of VLMs in cross-platform gaming tasks. Building upon this, we propose Yanyun-3, a cross-platform automation framework centered on the vision–language model Qwen2.5-VL and the executor UI-TARS (from ByteDance). We curate a multimodal dataset spanning three heterogeneous strategy game platforms, encompassing single images, multi-image sequences, videos, and their combined variants, and conduct systematic ablation studies on both modality and *combination granularity*. Yanyun-3 performs core tasks—including attack, resource allocation, movement, and area control—across diverse platforms using a single set of shared weights. To our knowledge, this work is the first to demonstrate that VLMs, via structured multimodal fine-tuning, can achieve cross-platform automated strategy game operation in a human-like [29] interactive manner, i.e., through "observing the screen and controlling the mouse/keyboard."

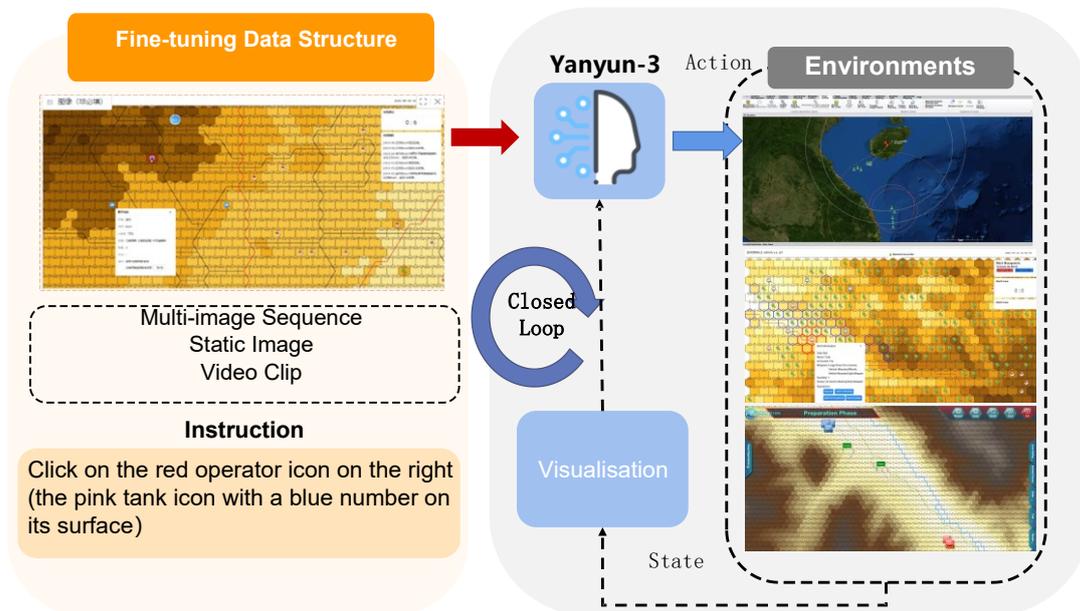

**Fig. 1**. Yanyun-3 Fine-tuning Stage. Yanyun-3 perceives the screen state [30] across diverse strategy game environments and outputs natural language action commands [3], which are then parsed and executed by the integrated UI-TARS module.

# Results

The performance details of the large models fine-tuned on various datasets formed in the ablation experiments on different test sets are provided in Appendix B12.

## Modal Ablation Experiment

### Prediction Quality Ablation Experiment

This subsection divides the prediction quality ablation experiment into two parts: overall modality prediction quality comparison, and quantitative analysis of modality contributions with task-specific mechanisms, to strengthen its systematicity and interpretability.

*Overall Modality Prediction Quality Comparison*

This part focuses on the overall prediction quality comparison of the modality ablation experiment, providing a global perspective.

From Table B1, the prediction quality metrics corresponding to different modality datasets are extracted to form Table.1:

**Table.1**. Prediction Quality of Modality Ablation Experiment for Each Dataset

| Test_Set | Type | Symbol | Dataset | BLEU-4 ↑% | ROUGE-1 ↑% | ROUGE-2 ↑% | ROUGE-L ↑% |
|---|---|---|---|---|---|---|---|
| val_sum | Base Model | Base | 0 | 0.78 | 7.24 | 0.54 | 3.53 |
| | Single Image | S | annotions_new2.1 | **34.82** | **51.50** | **37.64** | **48.39** |
| | Multi-Image | M | MI2.8.3 | 11.69 | *33.93* | *13.83* | 20.70 |
| | Video | V | my_video_data | 1.90 | 16.87 | 3.39 | 7.56 |
| | Single Image*Multi-Image | S*M | combo_C1.2 | 9.24 | 31.07 | 11.17 | 18.70 |
| | Single Image*Video | S*V | combo_A1.1 | *12.29* | 31.73 | 11.12 | *27.40* |
| | Multi-Image * Video | M*V | combo_B1.7 | 3.13 | 16.01 | 3.52 | 8.96 |
| | Full Fusion | S*V*M | combo_D1.2 | 8.42 | 28.36 | 9.53 | 17.66 |
| val_S | Base Model | Base | 0 | 0.57 | 5.49 | 0.43 | 2.27 |
| | Single Image | S | annotions_new2.1 | **45.84** | **60.34** | **46.72** | **60.12** |
| | Multi-Image | M | MI2.8.3 | 6.32 | *28.13* | 7.54 | 14.66 |
| | Video | V | my_video_data | 1.25 | 12.97 | 1.78 | 4.71 |
| | Single Image*Multi-Image | S*M | combo_C1.2 | 5.33 | 26.63 | 6.77 | 13.21 |
| | Single Image*Video | S*V | combo_A1.1 | *13.04* | 28.85 | *7.819* | *27.40* |
| | Multi-Image * Video | M*V | combo_B1.7 | 1.12 | 11.05 | 1.23 | 4.51 |
| | Full Fusion | S*V*M | combo_D1.2 | 4.81 | 23.46 | 4.91 | 12.84 |

To more intuitively observe the performance of different datasets, Fig. 2 is formed from the above table:

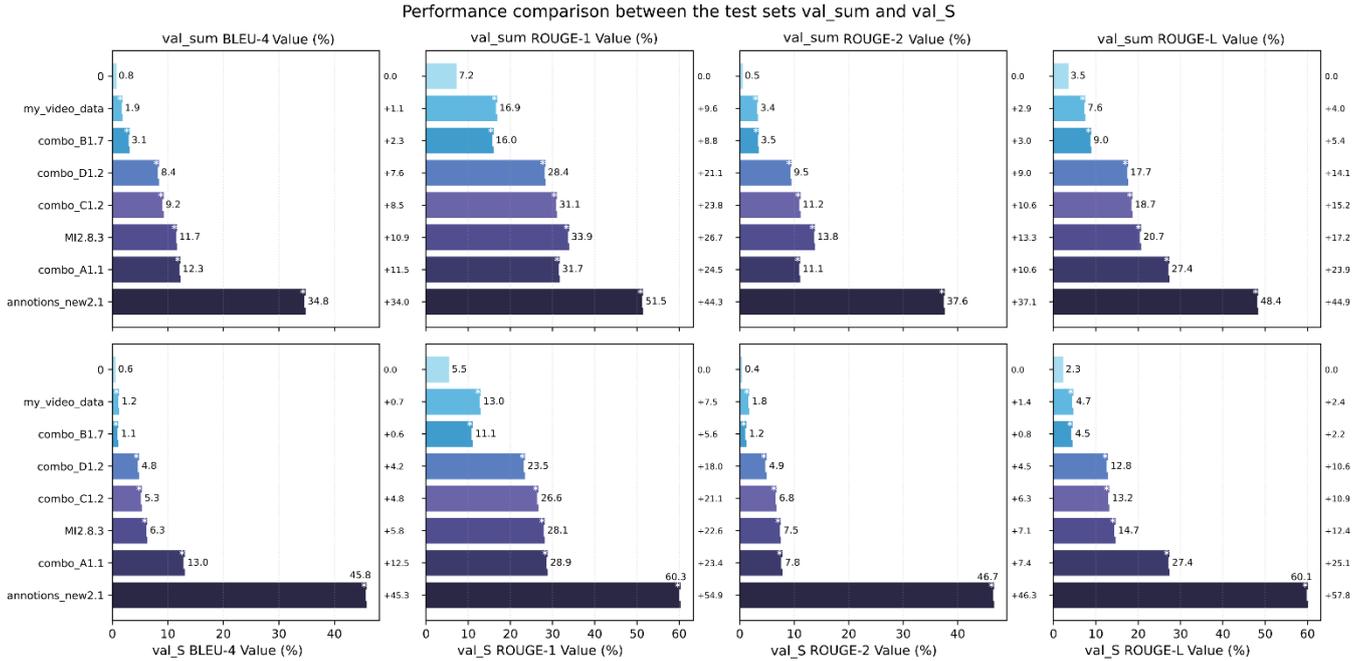

**Fig. 2.** Prediction Quality Performance Chart of the Modality Ablation Experiment. This chart shows the performance of large models fine-tuned on different datasets across four prediction quality metrics on two test sets. It more intuitively shows that the single-image dataset's performance is significantly better than other datasets.

As shown in Table.1 and Fig. 2, the single-image modality (S) shows a dominant lead in performance on the comprehensive test set (val_sum), with its BLEU-4 score (34.82%) being approximately three times that of the closest competing modality (multi-image modality M, 11.69%). This phenomenon indicates that the atomic operations in strategy games are highly dependent on precise perception of the instantaneous interface state. The single-image modality captures the spatial configuration of the interface with the highest signal-to-noise ratio, providing a visual anchor for action primitives. In contrast, the multi-image (M) and video (V) modalities, due to containing temporal redundant information or because the model is not optimized for long sequences, fail to effectively utilize their dynamic information, which even becomes a source of interference.

Notably, the performance advantage of the single-image modality is further amplified on the val_S test set, which is specialized for static interface element recognition (BLEU-4 increases from 34.82% to 45.84%). This reinforces our conclusion: the characteristics of static tasks perfectly align with the advantages of the single-image modality. Conversely, the performance of the multi-image and video modalities decreases on val_S, as their dynamic information becomes redundant in static tasks, confirming the principle that modality selection must be precisely aligned with task requirements.

*Quantitative Analysis of Modality Contributions and Task-Specific Mechanisms*

To rigorously quantify the contribution of each modality and elucidate its underlying mechanism, we introduce the Performance Decline (PD) metric for systematic analysis. Table.2 comprehensively details the performance shifts on both the multi-task (val_sum) and single-task (val_S) test sets resulting from the ablation of individual modalities. This global analysis unveils a clear hierarchy of modality value.

We define PD to precisely characterize the role of each modality within the dataset. It quantifies the relative change in a given metric upon the removal of a specific modality, calculated as follows:

$$PD = \frac{Baselinevalue - Finalvalue}{Baselinevalue} \times 100\%$$

where *Baseline* denotes the metric value of the model fine-tuned on the full fusion dataset, and Metric without modality is the score obtained after removing a specific modality. A positive *PD* indicates that the omitted modality was beneficial (its removal harms performance), whereas a negative *PD* implies that excluding the modality actually improves results—suggesting interference or redundancy.

Table.2. Performance Impact of Modality Ablation Across Task Settings

| Test_Set | Exp.NO. | S | M | V | BLEU-4 ↑% | PD | ROUGE-1 ↑% | PD | ROUGE-2 ↑% | PD | ROUGE-L ↑% | PD |
|---|---|---|---|---|---|---|---|---|---|---|---|---|
| val_sum | 1 | √ | √ | √ | 8.42 | 0% | 28.36 | 0% | 9.53 | 0% | 17.66 | 0% |
| | 2 | √ | √ | × | 9.24 | -9.74% | 31.07 | -9.56% | 11.17 | -17.21% | 18.70 | -5.89% |
| | 3 | √ | × | √ | 12.29 | -45.96% | 31.73 | -11.88% | 11.12 | -16.68% | 27.40 | -55.15% |
| | 4 | × | √ | √ | 3.13 | 62.83% | 16.01 | 43.55% | 3.52 | 63.06% | 8.96 | 49.26% |
| | 5 | √ | × | × | 34.82 | -313.54% | 51.50 | -81.59% | 37.64 | -294.96% | 48.39 | -174.01% |
| | 6 | × | √ | × | 11.69 | -38.84% | 33.93 | -19.64% | 13.83 | -45.12% | 20.70 | -17.21% |
| | 7 | × | × | √ | 1.90 | 77.43% | 16.87 | 40.51% | 3.39 | 64.43% | 7.56 | 57.19% |
| | 8 | × | × | × | 0.78 | 90.74% | 7.24 | 74.47% | 0.54 | 94.33% | 3.53 | 80.01% |
| val_S | 9 | √ | √ | √ | 4.81 | 0% | 23.46 | 0% | 4.91 | 0% | 12.84 | 0% |
| | 10 | √ | √ | × | 5.33 | -10.81% | 26.63 | -13.51% | 6.77 | -37.88% | 13.21 | -2.88% |
| | 11 | √ | × | √ | 13.04 | -171.10% | 28.85 | -22.97% | 7.819 | -59.25% | 27.40 | -113.40% |
| | 12 | × | √ | √ | 1.12 | 76.72% | 11.05 | 52.90% | 1.23 | 74.95% | 4.51 | 64.88% |
| | 13 | √ | × | × | 45.84 | -853.01% | 60.34 | -157.20% | 46.72 | -851.53% | 60.12 | -368.22% |
| | 14 | × | √ | × | 6.32 | -31.39% | 28.13 | -19.91% | 7.54 | -53.56% | 14.66 | -14.17% |
| | 15 | × | × | √ | 1.25 | 74.01% | 12.97 | 44.72% | 1.78 | 63.75% | 4.71 | 63.32% |
| | 16 | × | × | × | 0.57 | 88.15% | 5.49 | 76.60% | 0.43 | 91.24% | 2.27 | 82.32% |

The analysis confirms that the single-image modality (S) serves as an indispensable cornerstone. Its absence precipitates a severe performance collapse (e.g., Experiment 4: val_sum BLEU-4 PD = +62.83%), while its isolated use yields an order-of-magnitude performance gain (Experiment 5: PD = -313.54%). In stark contrast, the dynamic modalities (multi-image M, video V) offer limited utility when deployed in isolation [31]; however, their value emerges in specific synergistic combinations. Fig. 3 visually encapsulates this trend and underscores the critical regulatory role of the task environment: the supremacy of the single-image modality is even more pronounced in static tasks (val_S).

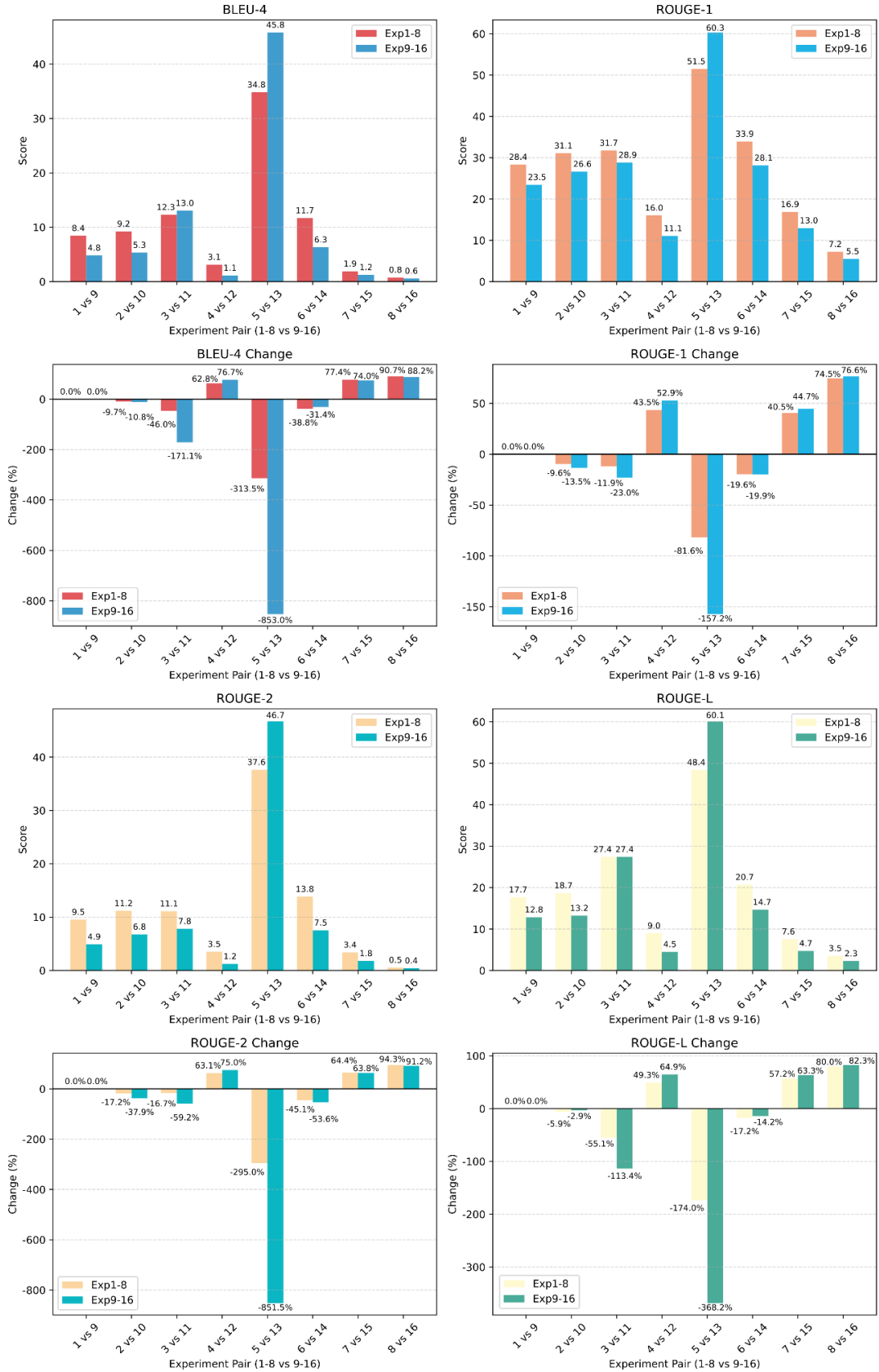

**Fig. 3**. Modality Ablation Analysis Across Multi-task and Single-task Scenarios.

(i) The Cornerstone Role and Independence of the Single-Image Modality

To dissect the foundational role of the single-image modality, we conducted a controlled comparative analysis. Table.3 quantifies its contribution across various configurations. The isolated control (Experiment 5 vs. 8) demonstrates that the single-image modality alone encapsulates the majority of the task-relevant information. Crucially, the combination control reveals that fusing the single image with video (Experiment 3, PD = -45.96%) yields a substantially greater performance gain than fusing it with multi-images (Experiment 2, PD = -9.74%), indicating a stronger synergistic relationship between static snapshots and continuous video dynamics.

Table.3. Isolated and Combined Control Analysis for the Single-image Modality

| Test_Set | | Isolated Control | | Combined Control | | | | | | |
|---|---|---|---|---|---|---|---|---|---|---|
| | Experiment ID | 5 | 8 | 1 | 4 | 2 | 6 | 3 | 7 |
| val_sum | PD | -313.54% | 90.74% | 0% | 62.83% | -9.74% | -38.84% | -45.96% | 77.43% |
| | Difference Value | -404.28% | | -62.83% | | 29.10% | | -123.39% | |
| val_S | Experiment ID | 13 | 16 | 9 | 12 | 10 | 14 | 11 | 15 |
| | PD | -853.01% | 88.15% | 0% | 76.72% | -10.81% | -31.39% | -171.10% | 74.01% |
| | Difference Value | -941.16% | | -76.72% | | 20.58% | | -245.11% | |

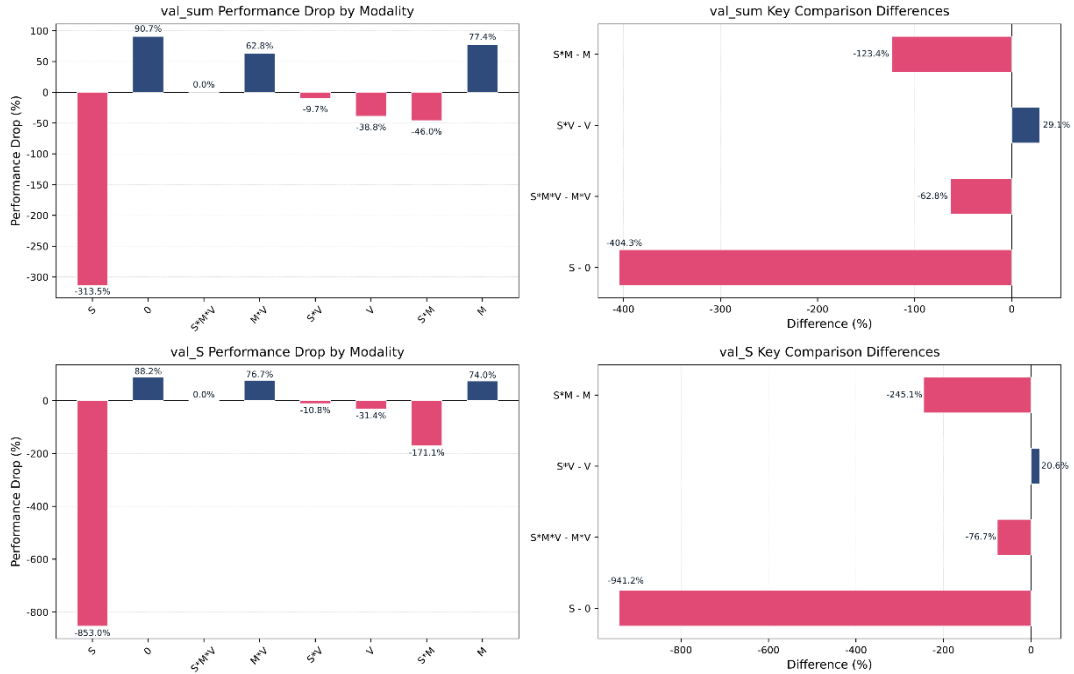

**Fig. 4**. Ablation Performance of the Single-image Modality Across Combination Strategies.

Fig. 4 visually corroborates this finding. The single-image modality achieves its peak performance in isolation, and this performance is further amplified when fused with video. Conversely, its efficacy is diminished when fused with multi-images. This underscores that the core value of the single-image modality stems from its high signal-to-noise ratio and functional independence. Acting as a stable static anchor, it exhibits maximal complementarity with the video modality, which supplies rich, continuous dynamic context.

(ii) Synergy and Dependency of Dynamic Modalities

The dynamic modalities (multi-image M, video V) exhibit a distinct behavioral pattern. As Table.4 illustrates, the multi-image modality is not only ineffective but actively detrimental when used in isolation (Experiment 6, PD < 0). Its utility emerges only when fused with video (Experiment 4), highlighting a critical dependency. This finding signifies that the phased contextual information from multi-images must be intra-sample aligned with the continuous dynamics from video to enable mutual compensation and effective synergy.

Table.4. Isolated and Combined Control Analysis for the Multi-image Modality.

| Test_Set | Isolated Control | | | Combined Control | | | | | | |
|---|---|---|---|---|---|---|---|---|---|---|
| val_sum | Experiment ID | 6 | 8 | 1 | 4 | 2 | 6 | 3 | 7 |
| | PD | -38.84% | 90.74% | 0% | -45.96% | -9.74% | -313.54% | 62.83% | 77.43% |
| | Difference Value | -129.58% | | 45.96% | | 303.80% | | -14.60% | |
| val_S | Experiment ID | 14 | 16 | 9 | 11 | 10 | 13 | 12 | 15 |
| | PD | -31.39% | 88.15% | 0% | -171.10% | -10.81% | -853.01% | 76.72% | 74.01% |
| | Difference Value | -119.54% | | 171.10% | | 842.20 % | | 2.71% | |

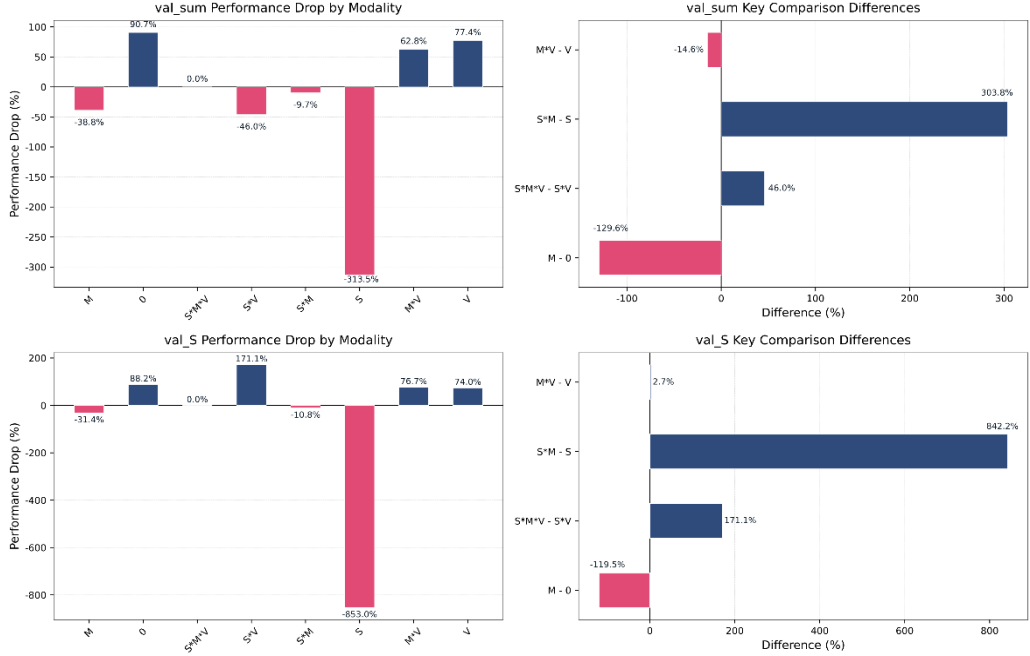

Fig. 5. Ablation Performance of the Multi-image Modality Across Combination Strategies

Fig. 5 further elucidates this dependency: the multi-image modality's value is contingent upon the integration strategy. In isolation, it degrades model performance, yet it forms a potent and meaningful complementarity when fused with video. This provides a robust empirical foundation for our subsequent adoption of the M*V fusion strategy.

An analysis of the video modality (Table.5 and Fig. 6) consolidates these insights. The video modality is the weakest contributor among all single modalities [31], as it lacks sufficient standalone informational content [32] to effectively support generation and comprehension tasks. Nonetheless, it can partially enhance performance [33] when fused with multi-images (Experiment 4), once again underscoring the necessity of synergy between dynamic modalities. This fundamental limitation is rooted in the base model's inherent difficulty in modeling long-range temporal dependencies within video sequences.

Table.5. Isolated and Combined Control Analysis for the Video Modality

| Test_Set | Isolated Control | | | Combined Control | | | | | | |
|---|---|---|---|---|---|---|---|---|---|---|
| val_sum | Experiment ID | 7 | 8 | 1 | 2 | 3 | 5 | 4 | 6 |
| | PD | 77.43% | 90.74% | 0% | -9.74% | -45.96% | -313.54% | 62.83% | -38.84% |
| | Difference Value | -13.31 % | | 9.74 % | | 267.58 % | | 101.67 % | |
| val_S | Experiment ID | 15 | 16 | 9 | 10 | 11 | 13 | 12 | 14 |
| | PD | 74.01% | 88.15% | 0% | -10.81% | -171.10% | -853.01% | 76.72% | -31.39% |
| | Difference Value | -14.14 % | | 10.81 % | | 681.91 % | | 108.11 % | |

For visual clarity, Fig. 6 presents the performance trends derived from this table.

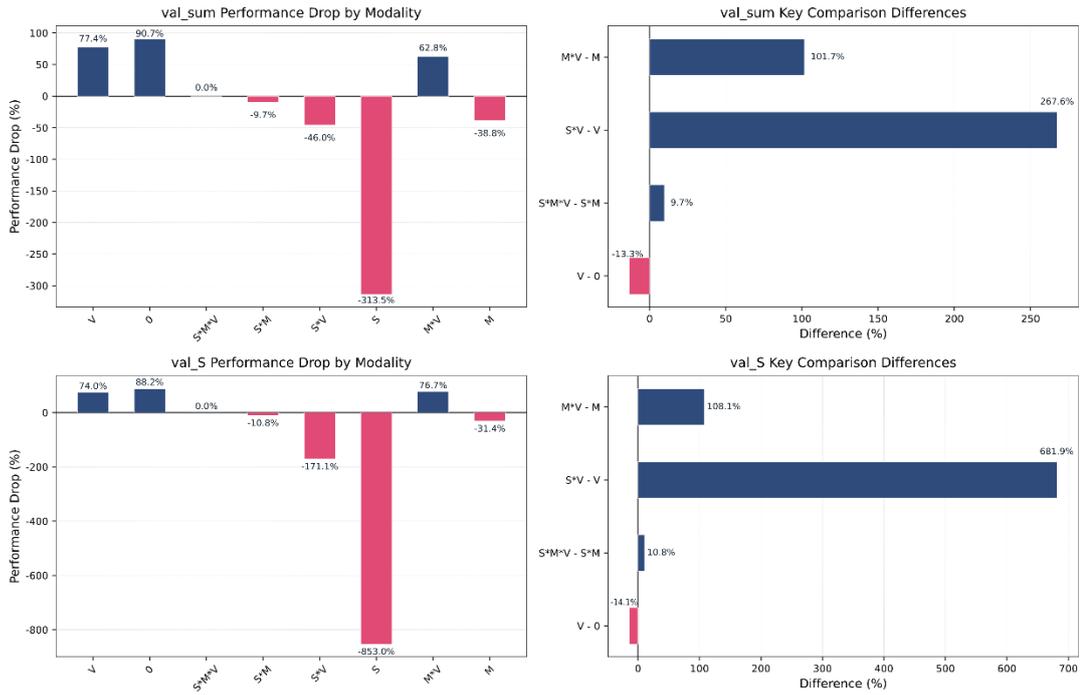

**Fig. 6**. Ablation Performance of the Video Modality Across Combination Strategies

(iii) Summary

In summary, our systematic quantitative analysis (Tables 4–7, Figures 4–7) reveals the fundamental mechanisms governing modality roles:

- Independence: The single-image modality serves as the performance cornerstone, with its informational characteristics being exquisitely aligned with the core task requirements.
- Synergy: The value of dynamic modalities (multi-image, video) is not intrinsic but emergent, contingent upon intra-sample fusion to achieve effective complementarity.
- Task-Dependent Regulation: The relative importance of each modality is profoundly governed by the nature of the task—be it dynamic or static.

Collectively, these findings establish that the modality organization strategy—namely, *combination granularity*—is of equal importance to the choice of modality type itself. This insight naturally propels us to the central question of the next section: given a full set of modalities, which *combination granularity* strategy can optimally orchestrate their inherent independence and synergistic potential?

**Efficiency Ablation Experiment**

Computational efficiency is a critical factor for real-world agent deployment. Here, we analyze the inference latency, throughput, and computational overhead of models trained on each modality configuration, using the comprehensive metrics in Table. 6, to elucidate the intrinsic relationship between modality-specific information characteristics and computational cost.

The prediction quality metrics for all datasets extracted from Appendix Table B1 are shown in Table. 6:

**Table. 6**. Efficiency Metrics Across Modality Configurations in the Ablation Study

| Test_Set | Type | Symbol | Dataset | MPT ↓ s | RT ↓ s | SAMPLE/s ↑ | STEPS/s ↑ |
|---|---|---|---|---|---|---|---|
| val_sum | Base Model | Base | 0 | 0.012 | 2828.3704 | 0.094 | 0.012 |

| | Single Image | S | annotions_new2.1 | 0.0094 | **729.108** | **0.363** | 0.0094 |
| | Multi-Image | M | MI2.8.3 | 0.0095 | 2306.4003 | 0.115 | 0.0095 |
| | Video | V | my_video_data | 0.0203 | 2360.4538 | 0.112 | 0.0203 |
| | Single Image*Multi-Image | S*M | combo_C1.2 | **0.007** | 1178.9005 | 0.225 | **0.113** |
| | Single Image*Video | S*V | combo_A1.1 | 0.0134 | 786.4008 | 0.337 | 0.0134 |
| | Multi-Image * Video | M*V | combo_B1.7 | 0.0185 | 2480.9239 | 0.107 | 0.0185 |
| | Full Fusion | S*V*M | combo_D1.2 | 0.0095 | 1109.5285 | 0.239 | 0.12 |
| val_S | Base Model | Base | 0 | 0.0093 | 1924.7092 | 0.104 | 0.052 |
| | Single Image | S | annotions_new2.1 | 0.011 | 177.7917 | **1.131** | **0.568** |
| | Multi-Image | M | MI2.8.3 | 0.0093 | 1542.4098 | 0.13 | 0.065 |
| | Video | V | my_video_data | 0.0123 | 1414.4485 | 0.142 | 0.071 |
| | Single Image*Multi-Image | S*M | combo_C1.2 | **0.0071** | 747.9945 | 0.269 | 0.135 |
| | Single Image*Video | S*V | combo_A1.1 | 0.0142 | 225.4404 | 0.892 | 0.448 |
| | Multi-Image * Video | M*V | combo_B1.7 | 0.0115 | 1133.4459 | 0.177 | 0.089 |
| | Full Fusion | S*V*M | combo_D1.2 | 0.0069 | 596.618 | 0.337 | 0.169 |

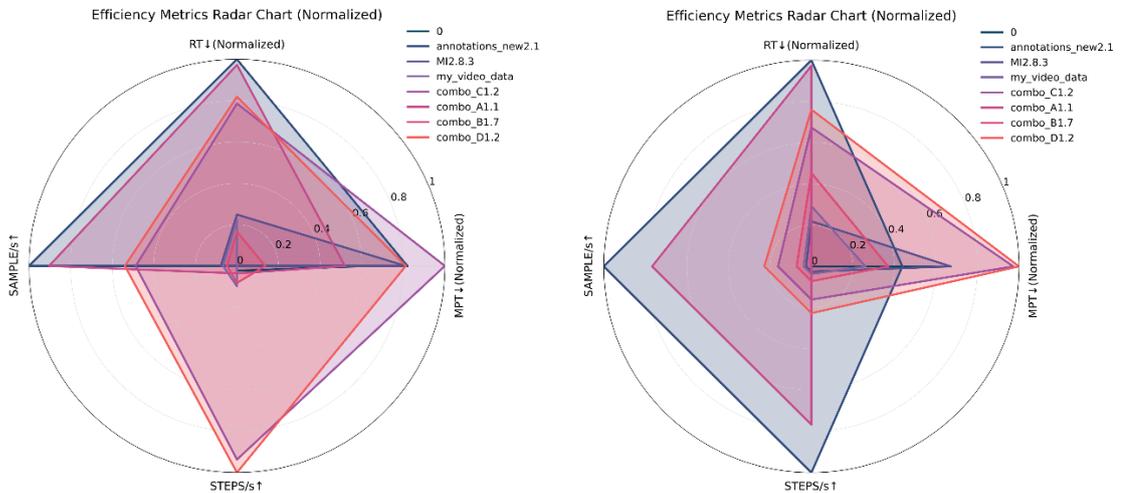

**Fig. 7**. Multidimensional Efficiency Comparison via Radar Charts. (a) Multi-task (val_sum) and (b) single-task (val_S) test set performance across MPT, Runtime, SAMPLE/s, and STEPS/s. A larger enclosed area indicates higher overall efficiency.

Our analysis reveals that computational efficiency is primarily governed by a modality's information density and encoding complexity. The single-image modality (S, annotations_new2.1) exhibits the highest efficiency across both test sets. On the static-focused val_S test set, its inference time (RT = 177.79 s) is merely 24.4% of that on the dynamic val_sum test set (RT = 729.11 s), while its throughput (SAMPLE/s = 1.131) increases by nearly an order of magnitude. This superior efficiency stems from the fact that a single, static image provides high signal-to-noise ratio instantaneous state information, enabling the vision encoder (e.g., Qwen2.5-VL's ViT) to extract features rapidly without the burden of temporal modeling.

In contrast, dynamic modalities suffer from inherent inefficiencies due to redundant information processing. The video modality (V, my_video_data) incurs the highest computational overhead, with model preparation time (MPT) of 0.0203 s and 0.0123 s on val_sum and val_S, respectively. Its limited RT reduction (16.5%–26.5%) underscores the base model's challenges in efficient long-range temporal modeling, where inter-frame redundancy dilutes key signals and hinders dynamic information extraction. The multi-image modality (M, MI2.8.3), while more efficient than video, still shows suboptimal gains (RT and throughput improvements < 25%) due to sequence redundancy (e.g., high similarity between consecutive frames), indicating a need for fine-grained integration to unlock its contextual value without incurring excessive cost.

Crucially, multimodal combination strategies exert a profound regulatory effect on efficiency. The S*V fusion strategy (combo_A1.1) achieves a dual optimization, dramatically reducing RT by 72.2% and 88.3% on val_sum and val_S, respectively, while boosting throughput (SAMPLE/s) to 0.337 and 0.892. This synergy arises because the static image acts as an anchor to suppress redundancy, while the video enriches context, and shared encoding pathways minimize overhead. Conversely, the M*V strategy (combo_B1.7) is the least efficient, with RT on val_sum decreasing by only 12.3%, as the compounded redundancy of two dynamic modalities leads to a superposition of temporal encoding burdens. The full-fusion baseline (S*M*V) demonstrates moderate efficiency—outperforming individual dynamic modalities but falling short of the optimal, thereby confirming that naive modality stacking introduces significant feature overlap and computational waste.

Fig. 7 provides a multidimensional visualization of these trade-offs. The near-circular profile of the single-image modality reflects its balanced and comprehensive efficiency, while the robust shape of the S*V strategy and the markedly "dented" contour of the M*V strategy offer intuitive, visual confirmation of our quantitative findings.

In summary, the efficiency ablation study establishes that: (i) the single-image modality is the cornerstone of computational efficiency, particularly well-suited for static tasks; (ii) dynamic modalities require structured integration—such as the S*V fusion strategy—to mitigate redundancy; and (iii) the efficiency gain from static data is far more pronounced in static-dominant tasks (val_S: 8.7× speedup) than in dynamic tasks (val_sum: 3.1× speedup). This insight provides a clear engineering guideline for practical deployment: prioritize single-image data and integrate dynamic information via intra-sample fusion, while avoiding simplistic mixing strategies.

**Summary of Modality Ablation**

The modality ablation study conclusively demonstrates that the contributions of distinct visual modalities to model performance are highly heterogeneous, a finding robustly corroborated by both prediction quality (e.g., BLEU-4) and computational efficiency (e.g., inference latency) metrics. The single-image (Static) modality is established as the highest-value atomic data source [24,34]: on the multi-task val_sum test set, it achieves a BLEU-4 score of 34.82%, substantially outperforming the multi-image (12.45%) and video (0.96%) modalities. Concurrently, it exhibits the best computational efficiency, with an inference time approximately 70% shorter than that of the video modality. This dual advantage stems from its high signal-to-noise ratio, as single images provide a stable, instantaneous snapshot that serves as a precise anchor for interface element recognition and atomic action generation. The multi-image modality, when used in isolation, demonstrates some contextual perception capability (e.g., elevated ROUGE-L scores), but its utility is critically dependent on synergistic combination with other modalities (e.g., fusion with video enhances temporal continuity); its standalone performance is hampered by redundant phase information. The video modality exhibits the lowest information utility in this domain; its integration frequently introduces temporal noise and alignment instabilities (e.g., high variance in the feature extractor f(v) can cause the fusion function g to fail), not only failing to compensate for missing modalities but actively degrading model stability [31,35]. Synthesizing quality and efficiency metrics, we establish a definitive performance hierarchy: single-image > multi-image > video. This hierarchy encapsulates a fundamental principle: static data provides essential grounding, whereas dynamic data requires structured, purposeful integration to mitigate redundancy and unlock its value.

# Combination Granularity Ablation Experiment

The preceding modality ablation study did not isolate the effect of *combination granularity* under a fixed modality composition. To directly probe this critical factor, we compare the performance of three datasets that incorporate the full set of modalities (S, M, V) but employ distinct *combination granularity* strategies: full intra-sample fusion (S*M*V), partial fusion with mixing (S*V+M), and our proposed hybrid strategy (M*V+S). The performance of the latter two strategies across both test sets is also documented in Appendix Table B1.

## Prediction Quality Ablation Experiment

### Global Comparison of Combination Strategies

To dissect the impact of modality organization, we benchmark our proposed hybrid strategy (M*V+S) against the full-fusion baseline (S*M*V) and an alternative mixed strategy (S*V+M).

Table. 7. Prediction Quality under Different *Combination Granularity* Strategies

| Test_Set | Symbol | Dataset | BLEU-4 ↑ % | ROUGE-1 ↑ % | ROUGE-2 ↑ % | ROUGE-L ↑ % |
|---|---|---|---|---|---|---|
| val_sum | S*M*V | combo_D1.2 | 8.42 | 28.36 | 9.53 | 17.66 |
|  | S*V+M | combo_D1SVM | 10.54 | 30.41 | 13.71 | 20.28 |
|  | M*V+S | combo_D1MVS | **47.79** | **65.23** | **53.27** | **61.01** |
| val_S | S*M*V | combo_D1.2 | 4.81 | 23.46 | 4.91 | 12.85 |
|  | S*V+M | combo_D1SVM | 4.37 | 21.90 | 6.14 | 11.74 |
|  | M*V+S | combo_D1MVS | **62.41** | **74.86** | **64.15** | **74.60** |

As shown in Table. 7, the M*V+S strategy yields a quantum leap in prediction quality, achieving BLEU-4 scores of 47.79 and 62.41 on the val_sum and val_S test sets, respectively—representing an approximately 12.98-fold improvement over the full-fusion baseline. In contrast, the S*V+M strategy exhibits inconsistent performance, while the full-fusion approach delivers only mediocre results, thereby exposing the fundamental limitations of naive modality stacking.

For visual clarity, Fig. 8 presents these results as a comparative radar chart.

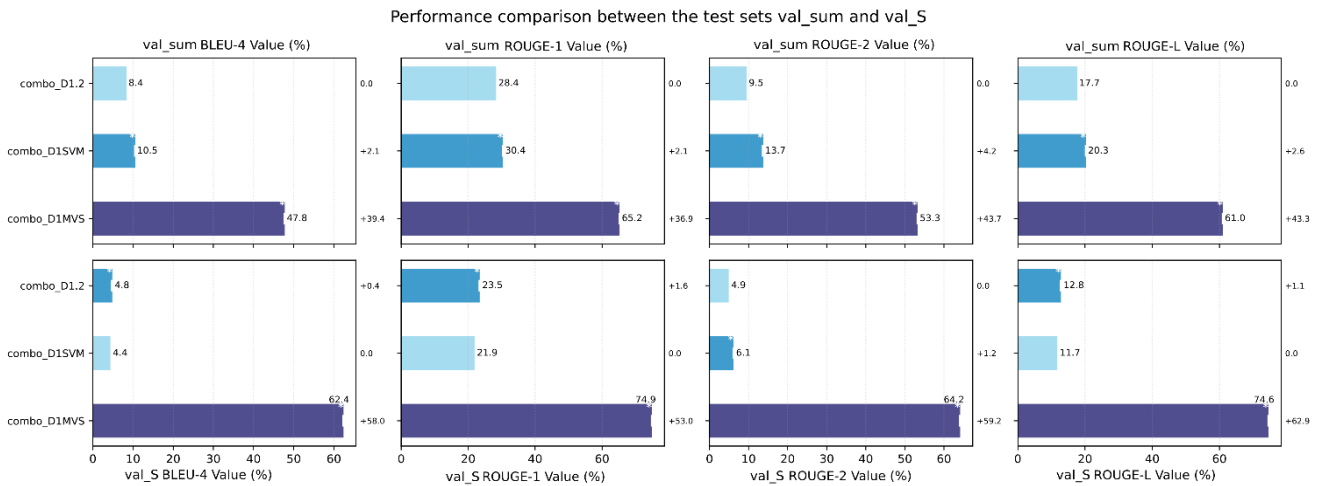

**Fig. 8**. Prediction Quality of the *Combination Granularity* Ablation Atudy. Performance across four metrics on both val_sum and val_S test sets.

This stark performance divergence stems from the principled alignment of modality characteristics with their organizational strategy. The M*V+S strategy's success is underpinned by its "dynamic fusion, static isolation" philosophy: intra-sample fusion of multi-image (M) and video (V) enables the model to learn the complementary alignment between discrete phase contexts and continuous dynamics, whereas inter-sample mixing of static images (S) preserves their high signal-to-noise ratio, preventing feature redundancy and interference from dynamic modalities (as S is often a cropped frame from M or V). This architecture effectively mimics the human cognitive process of functional specialization for complex tasks. Conversely, the S*V+M strategy severs the crucial temporal linkage between M and V, and the S*M*V strategy injects redundant noise by forcing an ill-suited fusion of all modalities within a single sample. Fig. 8 visually encapsulates this "discontinuous" performance advantage.

To ensure the robustness and statistical significance of the reported PD metric for the M*V+S strategy, we conducted 10 independent inference runs on the val_S test set, varying only the random seed for the text generation process while keeping all other hyperparameters (e.g., temperature=0.7, top_p=0.9) fixed. The strategy achieved a mean BLEU-4 score of $62.89 \pm 0.93$, yielding a mean PD of -1206.6% relative to the full-fusion baseline (4.81%). A one-sample t-test confirmed this performance gain is highly statistically significant ($t(9) = 198.56$, $p < 0.0001$). This rigorous statistical validation confirms that the observed superiority of the M*V+S strategy is not due to random chance but reflects a robust and reproducible phenomenon.

*Mechanistic Analysis of Combination Granularity*

Moving beyond aggregate performance, we conduct a mechanistic analysis to quantify the efficacy of the two fundamental operations—intra-sample fusion and inter-sample mixing—and to elucidate their interplay with task-specific demands.

**Table.8**. Performance Impact of *Combination Granularity* Strategies

| Test_Set | Exp. NO. | S | M | V | BLEU-4 ↑ % | PD | ROUGE-1 ↑ % | PD | ROUGE-2 ↑ % | PD | ROUGE-L ↑ % | PD |
|---|---|---|---|---|---|---|---|---|---|---|---|---|
| val_sum | 1 | * | * | * | 8.42 | 0% | 28.36 | 0% | 9.53 | 0% | 17.66 | 0% |
|  | 17 | * | + | * | 10.54 | -25.18% | 30.41 | -7.22% | 13.71 | -43.86 % | 20.28 | -14.84 % |
|  | 18 | + | * | * | 47.79 | -467.58% | 65.22 | -129.97 % | 53.27 | -458.97% | 61.01 | -245.47% |
| val_S | 9 | * | * | * | 4.81 | 0% | 23.46 | 0% | 4.91 | 0% | 12.84 | 0% |
|  | 19 | * | + | * | 4.37 | 9.15% | 21.90 | 6.65% | 6.14 | -25.05% | 11.74 | 8.57% |
|  | 20 | + | * | * | 62.41 | -1197.51% | 74.86 | -219.09 % | 64.15 | -1206.52% | 74.60 | -480.99% |

Table.8 quantifies the relative performance change (PD) of each strategy against the full-fusion (S*M*V) baseline. A critical finding is the profound regulatory role of task context. The M*V+S strategy demonstrates a vastly superior gain in the static val_S setting (PD = -1197.51%) compared to the dynamic val_sum setting (PD = -467.58%), confirming that its design is exquisitely tailored for static interface recognition. Conversely, the S*V+M strategy degrades performance on val_S (PD = +9.15%) while offering marginal gains on val_sum, highlighting its inherent instability and task misalignment. This reinforces our earlier conclusion (Modal Ablation Experiment) that the optimal modality combination is contingent upon the task's reliance on temporal information.

From Table.8, we obtain Fig. 9:

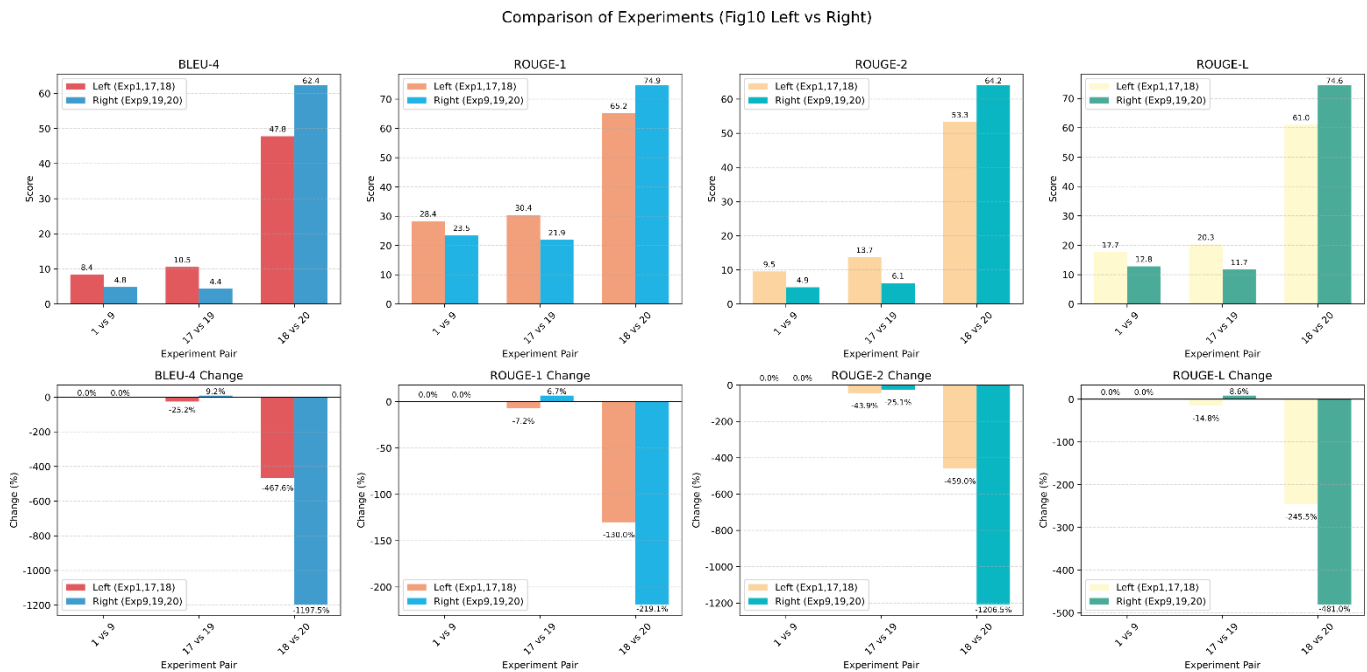

**Fig. 9**. Analysis of *Combination Granularity* Across Task Settings.

Fig. 9 visually reinforces the conclusions of Table 10:

(i) Cross-Task Superiority of M*V+S: As shown in the upper part of Fig. 9(metric values), the bar corresponding to the M*V+S strategy significantly outperforms others on both val_sum and val_S. In the lower part (metric change), its PD bar is the lowest, exhibiting the largest negative deviation from the baseline and creating a clear "performance trough". This pattern demonstrates the strategy's strong generalization prowess across multiple tasks. This cross-task robustness arises from its structured modality organization: intra-sample fusion of temporal modalities (M*V) enables the model to learn dynamic continuity [28,36], while inter-sample mixing of static images (+S) provides high-fidelity, independent anchors for state perception. This "dynamic fusion + static anchoring" paradigm mirrors human cognitive division of labor, optimizing performance for both contextual reasoning and precise atomic actions.

(ii) Task-Specific Failure of S*V+M: The S*V+M strategy exhibits a critical flaw: its PD for BLEU-4 becomes positive on val_S (Fig. 9), signaling performance degradation. This occurs because, while S*V fusion captures some static-dynamic associations, the multi-image modality (M) is introduced via mixing, which severs its essential temporal alignment with the video (V). In static tasks, this orphaned M information becomes redundant noise. This underscores the principle of "alignment consistency": dynamic modalities must be co-organized to realize their synergistic potential.

(iii) The Full-Fusion Baseline as a Negative Control: The full-fusion strategy consistently occupies an intermediate performance tier, serving as a compelling negative control that demonstrates the suboptimality of indiscriminate modality fusion.

**Computational Efficiency of *Combination Granularity***

The choice of *combination granularity* exerts a profound influence not only on prediction quality but also on the agent's real-time operational efficiency. As quantified in Table. 9, our optimal M*V+S strategy emerges as the most computationally efficient configuration. On the val_S test set, it achieves a **63.0%** reduction in inference time (RT) and a 170.6% increase in throughput (SAMPLE/s), successfully realizing the ideal of being "both highly effective and highly efficient."

Table. 9. Efficiency Metrics under Different *Combination Granularity* Strategies

| Test_Set | Symbol | Dataset | MPT ↓ | RT ↓ s | SAMPLE/s ↑ | STEPS/s ↑ |
|---|---|---|---|---|---|---|
| val_sum | S*M*V | combo_D1.2 | 0.0095 | 1109.5285 | 0.239 | 0.12 |
|  | S*V+M | combo_D1SVM | **0.018** | 1436.4456 | 0.184 | 0.047 |
|  | M*V+S | combo_D1MVS | 0.0101 | **724.3830** | **0.366** | **0.184** |
| val_S | S*M*V | combo_D1.2 | **0.0069** | 596.6180 | 0.337 | 0.169 |
|  | S*V+M | combo_D1SVM | 0.0168 | 811.8776 | 0.248 | 0.063 |
|  | M*V+S | combo_D1MVS | 0.0245 | **220.4956** | **0.912** | **0.458** |

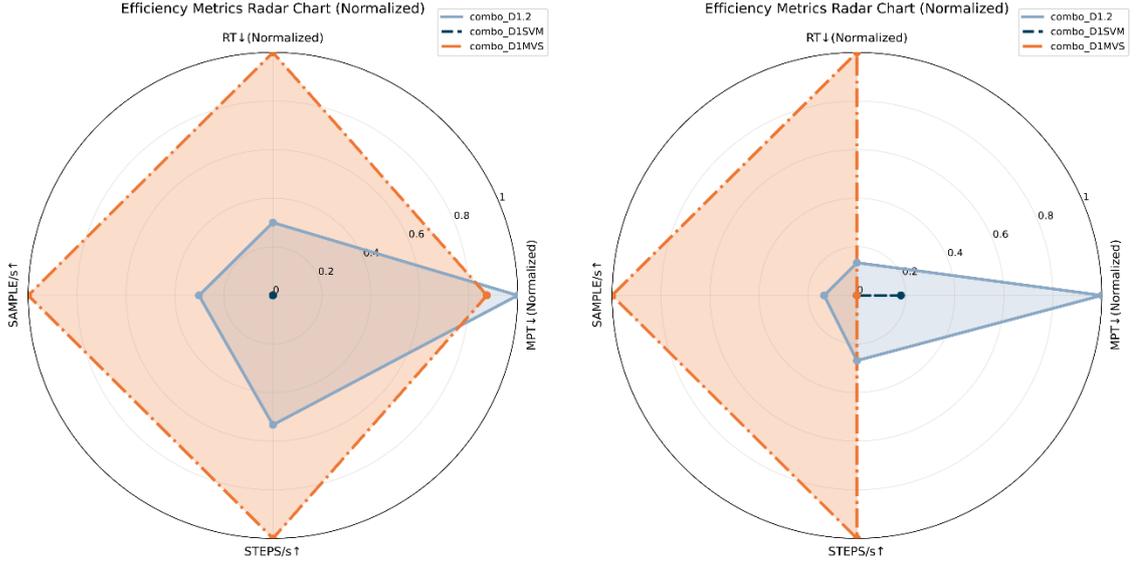

**Fig. 10**. Multidimensional Efficiency Comparison of *Combination Granularity* Strategies. (a) val_sum and (b) val_S test set performance across MPT, Runtime, SAMPLE/s, and STEPS/s.

This efficiency gain originates from its computationally parsimonious division of labor: intra-sample fusion of dynamic modalities (M*V) enables the model to share the vision encoder for temporal processing, thereby minimizing redundant computation, while inter-sample mixing of static images (+S) circumvents unnecessary feature extraction for this high-fidelity modality. In stark contrast, the S*V+M strategy exhibits comprehensively poor efficiency, a direct consequence of its misaligned modality organization that fails to leverage synergies. Fig. 10 provides a multidimensional visualization of these findings, clearly demonstrating that the M*V+S strategy occupies the largest area (indicating superior overall efficiency), whereas the S*V+M contour is severely contracted, offering compelling evidence for its practical deployment advantage.

**Summary of Combination Granularity Ablation**

The *combination granularity* ablation experiment uncovers a fundamental principle: the strategy for organizing modalities—i.e., *combination granularity*—is of equal importance to the choice of the modalities themselves. Even with an identical set of constituent modalities, the selection between intra-sample fusion (*) and inter-sample mixing (+) can precipitate order-of-magnitude performance differentials [37].

Our central finding is that dynamic modalities (multi-image M and video V), both of which encode temporal information, must be fused intra-sample (M*V) to unlock their synergistic potential. This fusion allows the model to learn their complementarity: the multi-image provides discrete phase context, while the video captures continuous dynamics, enabling the optimization of cross-modal representations via the alignment function a and fusion function g [38–40], thereby enhancing the model's understanding of process continuity [22]. Conversely, the static image (S) modality, which excels at instantaneous state perception, must be integrated via inter-sample mixing (+S). This approach preserves its high signal-to-noise ratio and prevents redundant interference, as the features of S and M are highly correlated (both being derived from V).

This "dynamic fusion, static mixing" (M*V+S) strategy effectively emulates the human cognitive architecture of modular information processing: dynamic sequences are jointly encoded to distill temporal consistency, while static snapshots are maintained as independent, high-fidelity anchors for action primitives. Empirical results confirm its absolute superiority in both prediction quality (BLEU-4 improved by 12.98×) and computational efficiency (inference time reduced by 63%). It decisively outperforms the full-fusion baseline (S*V*M, which suffers from redundant alignment) and partial fusion strategies (S*V+M, which is plagued by temporal misalignment).

As demonstrated in Fig. 3 and Fig. 9, its overall performance even eclipses that of the pure static image dataset [41], providing definitive proof that a structured *combination granularity* can orchestrate a true "1+1>2" synergistic effect.

**Deployment Recommendation**

The M*V+S strategy achieves an optimal Pareto balance between performance and efficiency, establishing it as the ideal configuration for the visual understanding core of cross-platform agents.

# System-Level Validation and Discussion of the Cross-Platform VLM Agent

This section details our system-level validation protocol and presents the results of feasibility pre-experiments.

**Prototype Feasibility Verification**

We begin by reporting the construction and initial successful execution of the minimum viable closed loop, which establishes an essential engineering baseline for subsequent system-level validation.

*Closed-loop Chain Establishment*

After fine-tuning for 30 epochs (without early stopping) using only the single-image dataset annotions_new2.1, we successfully executed the first complete closed-loop pipeline: "screenshot→Qwen2.5-VL inference→UI-TARS execution→interface refresh." This pipeline achieved an automatic run-through of a 6-step task flow on the GamePlatform B.

This milestone demonstrates three critical capabilities:

1) The VLM–executor interface is functionally viable;

2) Single-image data alone provides the minimal necessary signal for task execution;

3) The system possesses a foundational end-to-end iterative architecture.

*Capability Boundaries of the Prototype Stage*

Constrained by limited data volume and a single-modality design, the prototype system exhibited performance degradation in three representative failure modes: squad recognition, perception of subtle interface changes, and cross-platform scene differentiation. These failures indicate that the model has not yet acquired fine-grained visual–semantic alignment or robust long-range contextual reasoning capabilities. A detailed analysis of specific failure cases and their root causes is provided in Appendix E2.

**System-Level Validation and Performance Analysis**

Following the completion of our modality and *combination granularity* ablation studies, the Yanyun-3 model— fine-tuned for 27 epochs using the optimal M*V+S strategy (intra-sample fusion of multi-image and video, plus inter-sample mixing of single images)—was integrated with the UI-TARS executor for closed-loop testing across all three heterogeneous strategy game platforms.

A distinct game round was selected from each platform for evaluation, with comprehensive logs provided in Appendix E3. The summarized results are presented in Table.10.

**Table.10**. Summary of System-level Validation Outcomes across Platforms

| Platform | Scenario | Number of Tests | Number of Successes | Failure Attribution |
|---|---|---|---|---|
| GamePlatform A | Air/Ground Unit 1v1 | 100 | 30 | Icon recognition error [42], only mouse cursor changes after action execution |
| GamePlatform B | Ground 2v2 | 30 | 15 | Redundant actions [43], platform confusion |
| GamePlatform C | Ground 2v2 | 50 | 22 | Very small interface changes before and after action execution |

These results serve to establish Yanyun-3's foundational end-to-end operational capability, while also highlighting significant performance variance across platforms. Notably, on Platform B, where interface state changes are pronounced [41], the system achieved a 50% success rate (15/30), thereby empirically validating the critical role of explicit dynamic feedback in sustaining agent performance.

# Discussion

Grounded in our system-level validation, this section addresses two fundamental challenges: (1) the impact of multimodal organization strategies on agent performance boundaries, and (2) the ubiquitous bottlenecks and their underlying causes in vision-driven GUI automation. These challenges transcend the specific engineering context of this work and point toward general principles of multimodal embodied intelligence.

## *Combination Granularity*: A Structural Principle for Modality Organization

System-level validation demonstrates that the Yanyun-3 agent, employing the M*V+S strategy (intra-sample fusion of dynamic modalities plus inter-sample mixing of static images), can stably execute complex cross-platform operation flows, substantially outperforming the initial single-image-only prototype. Crucially, this improvement stems not from an increase in modality count, but from the structured organization of their intrinsic characteristics. This "dynamic fusion, static isolation" principle effectively emulates the human cognitive division of labor in complex interactive tasks: dynamic sequences are jointly processed to extract temporal context, while static snapshots serve as high-fidelity anchors for atomic decisions. Consequently, *combination granularity* transcends its role as a mere data construction technique; it functions as a mechanistic lever for enhancing multimodal collaboration efficiency. We posit that this principle holds significant transfer potential and warrants exploration in other embodied domains, such as robotic manipulation and autonomous driving.

## The Stalling Dilemma: A Fundamental Challenge in GUI Automation

Despite its strong performance in environments with pronounced interface changes (e.g., GamePlatform B), Yanyun-3 exhibits "stalling" on GamePlatform A and C—defined as the model's failure to generate the next instruction, thereby halting the operational flow. Our in-depth analysis identifies the root cause: post-action interface changes are often minimal (e.g., a mouse cursor shape change or a button highlight toggle), and the current VLM possesses limited perceptual acuity for such fine-grained dynamic differences [22].

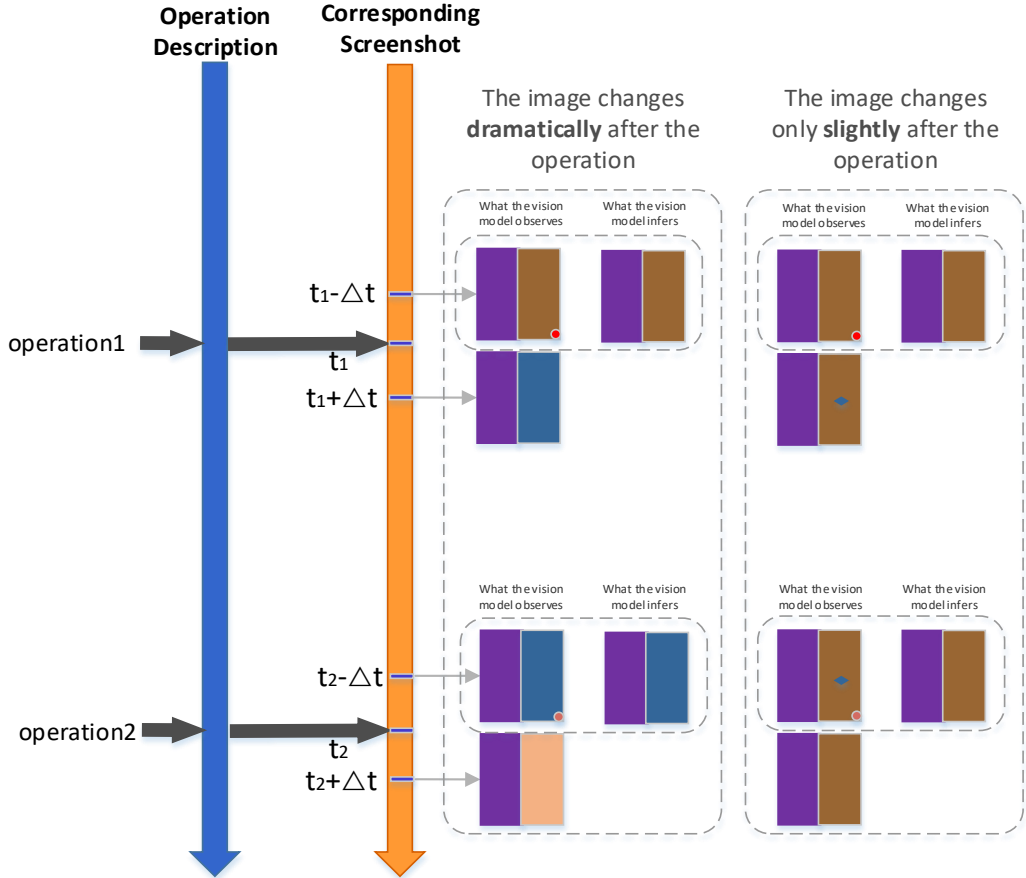

**Fig. 11**. Analysis of the Stalling Phenomenon. The current action is triggered by the pre-action screenshot. However, the visual delta between the post-previous-action and pre-current-action states is often negligible [44], leading to stalling. For instance, after a "resource allocation" action, the interface may only display a minuscule selection box or change the cursor to a "+" symbol—alterations that are both pixel-wise subtle and semantically sparse.

Human operators can overcome such ambiguity by leveraging memory of their action intent and expectations of feedback [45,46]. In contrast, purely vision-driven agents are critically dependent on explicit visual alignment [30,47]. When the inter-frame difference falls below the model's perceptual threshold, the system erroneously concludes that the "state is unchanged," failing to initiate the next reasoning cycle.

This phenomenon exposes a fundamental fragility inherent to pixel-based GUI automation in environments lacking explicit state signals. This bottleneck not only limits the robustness of strategy game agents but is also pervasive in operating system automation and office software assistants. To mitigate this, future work will focus on: a. developing attribute-level, fine-grained vision–language alignment mechanisms (e.g., disambiguating cursor state a from unit icon b in Fig. 12); b. integrating explicit state memory modules; and c. fusing programmatic semantic information (e.g., DOM trees, API logs) to reduce over-reliance on raw visual inputs.

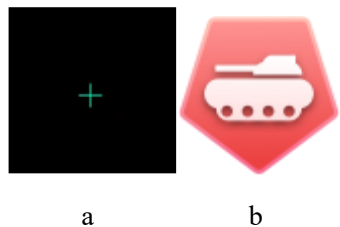

a        b

**Fig. 12**. Commonly Misidentified Visual Elements in Strategy Games. (a) The mouse cursor in its targeting state (used for resource allocation). (b) A Team A ground unit, capable of attacking Team B units and scouting adjacent hexes.

In summary, this study not only validates Yanyun-3's cross-platform feasibility but, through systematic discussion, also elucidates the mechanistic value of structured multimodal organization and identifies a core challenge in vision-driven automation, thereby offering both theoretical insights and concrete engineering pathways for the advancement of embodied intelligence. The principles revealed here extend beyond gaming: surgical robots require fusing dynamic video feeds while mixing static anatomical atlases; autonomous vehicles must fuse LiDAR sequences while mixing static map data. Our framework provides a general recipe for such multimodal organization.

## Conclusions

This study makes five principal contributions to the field:(1) A Formal Framework for Multimodal Data Organization: We introduce the concept of "*combination granularity*," which formally distinguishes between intra-sample fusion (*) and inter-sample mixing (+), establishing a systematic and principled methodology for VLM data construction.(2) Empirical Insights into Modality Synergy: Our ablation studies reveal that static images constitute the performance cornerstone, whereas dynamic modalities (multi-image, video) require intra-sample fusion to unlock their complementary potential. The proposed M*V+S strategy achieves a 12-fold improvement in BLEU-4 score and a 63% reduction in inference latency compared to full fusion. This finding directly challenges the prevailing heuristic in multimodal learning—that exhaustive intra-sample fusion is optimal—and demonstrates that structured organization is superior to naive stacking, thereby partially mitigating the visual cognition deficits of VLMs in temporal modeling [22].(3) A Deployable Embodied Agent Architecture: Yanyun-3 unifies perception, reasoning, and execution within a single, closed-loop pipeline, offering a practical and scalable solution for real-world applications such as military simulation and strategic decision support.(4) A Multi-dimensional Evaluation Benchmark: Moving beyond conventional accuracy metrics, our evaluation framework incorporates computational efficiency measures (e.g., MPT, SAM/s), establishing a comprehensive and reproducible benchmark for future work.(5) A New Paradigm for Multimodal Learning: We establish that strategic mixing is more effective than exhaustive fusion, thereby proposing a novel paradigm for multimodal learning that prioritizes the structured organization of information over its mere aggregation.

## Potential Impact and Outlook

The "*combination granularity*" framework and the Yanyun-3 agent transcend the specific domain of cross-platform strategy game automation. Our methodology offers a fundamental insight for the broader multimodal learning paradigm: the organization of modalities is as critical as the modalities themselves. This principle holds significant universal potential for a wide array of embodied intelligence tasks that hinge on the seamless coordination of static perception and dynamic reasoning, including robotic manipulation, autonomous driving, and general-purpose GUI automation.

The core contribution of this work is a transferable methodology, whose impact spans three key dimensions:

(1) Theoretical Paradigm Value: The formalization of "*combination granularity*" and the identification of the optimal "dynamic fusion, static mixing" (M*V+S) strategy [28] reveal that modality organization is a primary determinant of performance. This establishes a new theoretical paradigm for multimodal learning, moving beyond the simplistic aggregation of modalities. The principle is foundational and broadly applicable to any task that requires the integration of temporal dynamics with instantaneous static states.

(2) Technical Framework Value: We provide an end-to-end technical framework—from data curation to model fine-tuning—that directly addresses the pervasive challenge of "insensitivity to minimal interface changes" in interactive AI, as exemplified by the stalling phenomenon in Fig. 12. This framework offers a concrete blueprint for achieving robust perception and precise control in complex, dynamic environments, with direct relevance to embodied intelligence, robotic manipulation, and general GUI automation [11,29,47].

(3) Application Transfer Value: The validated Yanyun-3 architecture and its evaluation protocol chart a feasible pathway for the intelligent augmentation of specialized domains, such as civil decision support, and foreshadow its potential in realizing general-purpose human–computer interaction agents. For instance, in robotic manipulation, continuous video streams (encoding dynamic temporal context) could be integrated with key-state static snapshots (capturing instantaneous spatial configuration) using an M*V+S-inspired strategy to optimize contextual awareness. Furthermore, the proven "perception–reasoning–execution" closed-loop design and cross-platform generalization of Yanyun-3 lay a solid engineering foundation for the development of universal GUI automation agents capable of operating across diverse software ecosystems.

Building upon the limitations of the current study, we outline four promising avenues for future research to further unlock the potential of our framework:

(1) Enhancing Robustness to Minimal Interface Changes: To overcome the current agent's limited sensitivity to subtle visual cues (e.g., mouse cursor morphology)—a key bottleneck for reliable GUI automation—we will develop mechanisms for fine-grained perception and multi-source state representation. This will be pursued through three synergistic approaches: 1) constructing attribute-level, fine-grained vision–language representations [5,48] to enhance pixel-level change detection on dynamic screens; 2) integrating explicit state memory modules to enable reasoning based on action intent and historical context, thereby reducing dependence on immediate visual feedback [18]; and 3) fusing programmatic semantic information (e.g., DOM trees, API logs) to provide structured auxiliary signals that complement raw visual perception [47], fundamentally mitigating decision stalling induced by ambiguous interface states.

(2) Validating Generalization on Complex Platforms: To rigorously assess the generalization capacity of the "*combination granularity*" framework, we will evaluate it on more sophisticated commercial game environments (e.g., Hearts of Iron IV) that incorporate abstract knowledge and complex rule systems [49,50] Concurrently, we will benchmark the performance of diverse VLM backbones on these challenging interfaces to generate empirical guidelines for model selection, collectively advancing the field toward "general game agents."

(3) Innovating Spatio-Temporal Fine-tuning: Recognizing that current fine-tuning methods inadequately model the inherent spatio-temporal structure of multimodal data, we propose to develop STC-LoRA (Spatio-Temporal-Conditional LoRA). This novel method will leverage dedicated spatial and temporal LoRA adapters, coupled with dynamic rank adjustment, to achieve disentangled representation and adaptive fusion of spatio-temporal features [36,51]. This aims to more efficiently unlock the performance ceiling of models in complex, temporally extended decision-making tasks, offering new insights into parameter-efficient fine-tuning for multimodal foundation models.

(4) Extending to Broader Embodied Intelligence Domains: Finally, we will translate the "*combination granularity*" principle and the Yanyun-3 system architecture to embodied intelligence scenarios with richer physical interaction, such as robot visual manipulation and autonomous driving. This cross-domain validation will be crucial for establishing the universal applicability of our framework across the spectrum of embodied AI tasks.

## Methods

### System Architecture

Following an empirical investigation into the impact of various natural language instruction formats [52], [38] (see Appendix G for details) on the operational accuracy of UI-TARS, we established an instruction paradigm that enables precise execution of interface operations—such as clicking specific buttons or icons—on strategy game platforms. Building upon this, we constructed a multimodal dataset using Qwen2.5-VL-7B as the base model. This dataset comprises screen recording videos from three heterogeneous strategy game platforms (GamePlatform A, B, and C) along with the corresponding key-frame image–action instruction pairs [30,47] extracted therefrom. We performed parameter-efficient fine-tuning via QLoRA [53], endowing the model with the ability to recognize complex interface elements and generate precise, executable instructions. Upon integration with UI-TARS, the system forms a perception–reasoning–execution closed-loop pipeline [29,43], enabling automated operation through the following sequence: Screenshot→Qwen2.5-VL inference→Natural language action generation→UI-TARS parsing and execution→Environment refresh→New screenshot, thereby establishing a continuous operational loop [54].

The system comprises three core modules: VLM reasoning [18,35] (Qwen2.5-VL, which takes a screenshot as input and outputs a natural language instruction), executor (UI-TARS, which translates the instruction into concrete actions), and interface management [55] (an API layer dedicated to latency optimization). The overall architecture is illustrated in Fig. 13.

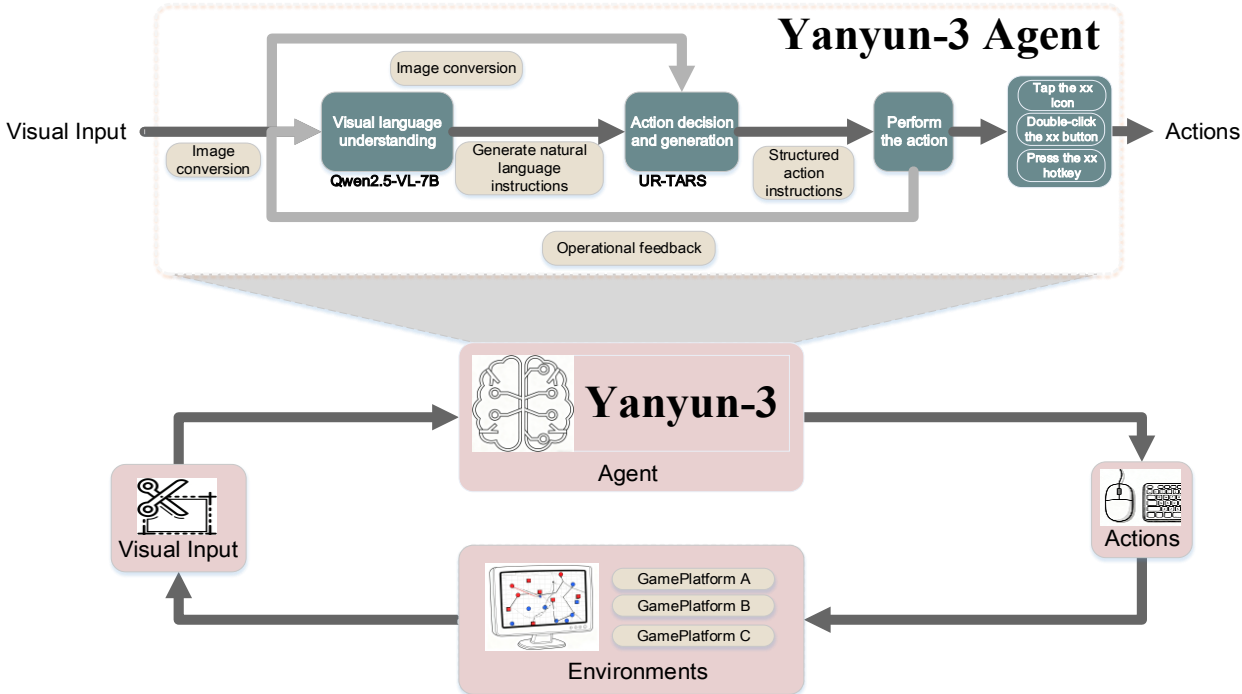

**Fig. 13**. Yanyun-3 System Architecture. Yanyun-3 executes learned actions within the strategy game environment. Upon any change in the game's displayed state, a new screenshot is captured and fed back to Yanyun-3, thus closing the loop for automated operation. The evaluation encompasses three heterogeneous strategy game platforms: GamePlatform A, GamePlatform B, and GamePlatform C. Further details of these environments are provided in Appendix A.

## Dataset Construction

This section details the multimodal dataset construction process and introduces two key conceptual definitions. A comprehensive description of the dataset design is provided in Appendix B.

**Modality and Sub-modality Definitions**

To account for the distinct information characteristics, data organization requirements, and model processing demands of strategy game tasks, we subdivide the visual modality into three image-based sub-modalities:

1) Static Image: Encodes instantaneous state information, aligning with action-level tasks. It provides a high-fidelity snapshot of the interface state immediately prior to an atomic action (e.g., clicking a specific button).

2) Multi-image Sequence: Encodes phase-level contextual information, corresponding to task-level objectives. By capturing transitions and dependencies across multiple frames, it supports reasoning over task phases (e.g., multi-step resource allocation sequences).

3) Video Clip: Represents continuous dynamic processes, mapping to scenario-level (or partial task-level) challenges. It preserves temporal dependencies and motion cues to facilitate reasoning about dynamic scenes (e.g., complex combat maneuvers).

This tripartite division reflects varying sampling densities along the temporal dimension [56], thereby establishing a controlled independent variable for our subsequent ablation studies.

**Base Datasets**

Static Image Dataset (annotions_new2.1): Each sample comprises a single interface screenshot paired with a corresponding natural language action instruction [19,57], specifically designed for single-step action recognition [58,41].

Multi-image Dataset (MI2.8.3): Structured around complete tasks, this dataset contains sequences of consecutive operation screenshots, thereby capturing the phased contextual information of a task.

Video Dataset (my_video_data): This dataset consists of short video clips [59,60] (each ⩽20 seconds in duration) that encapsulate continuous combat processes [32].

**Fused and Mixed Datasets and the Definition of *Combination Granularity***

Building upon the three base datasets and their derived sub-modalities, we construct various combined-modality datasets, which we categorize according to our proposed combination granularity framework. This strategy extends classical fusion paradigms (e.g., late fusion) [28] and optimizes modality complementarity by explicitly differentiating between intra-sample fusion (*) and inter-sample mixing (+). Table.11 delineates these two distinct combination approaches.

**Table.11.** *Combination Granularity*: Fusion vs. Mixing

| Aspect | Fusion (*) | Mixing (+) |
|---|---|---|
| Definition | Intra-sample integration of modalities within a single observation. | Inter-sample combination of observations across time or context. |
| Mathematical Representation | For a sample $x$ with modalities $x^{(m)}$, fusion yields a unified representation $z = g(a(f^{(m)}(x^{(m)})))$, where $f^{(m)}$ is feature extractor, a is alignment function, and g is fusion function. | For a dataset of N samples, mixing forms a mixed set $D = \bigcup_{i=1}^{N} \{(x_i^{(1)}, x_i^{(2)}, ..., x_i^{(M)})\}$, where each tuple combines modalities across samples. |
| Goal | Enable fine-grained cross-modal complementarity and synergistic information alignment at the sample level. | Emphasize statistical diversity and distributional variety across the training set, avoiding per-sample redundancy. |
| Example | Single Image * Video (combo_A1.1): Integrates a video clip and static images within one sample. | Single Image + Multi-image (combo_C1.1): Combines static image samples and multi-image sequence samples across the dataset. |

As summarized in Table 1, fused datasets prioritize fine-grained, intra-sample information complementarity, whereas mixed datasets emphasize inter-sample distributional diversity [19].

The distinction between intra-sample fusion and inter-sample mixing is not merely a design choice, but a principle grounded in the information structure of the modalities involved. Dynamic modalities—such as multi-image sequences (M) and video clips (V)—exhibit high intra-sample temporal mutual information: consecutive frames are strongly correlated, and their joint processing preserves action continuity and contextual coherence. In contrast, static snapshots (S) are inter-sample redundant (e.g., many frames share identical UI elements) yet intra-sample information-dense, encoding precise spatial configurations critical for atomic action execution.

In the context of strategy games, static elements (e.g., unit icons) demand high-precision recognition, while dynamic processes (e.g., combat flow) necessitate robust temporal modeling. Forcing the fusion of static and dynamic modalities within a single sample introduces feature misalignment: high-frequency static cues are overwhelmed by low-frequency motion patterns, degrading the signal-to-noise ratio. Conversely, mixing static samples preserves their semantic purity while enabling statistical generalization across diverse contexts—a strategy formalized below through an information-theoretic lens. Simple mixing, however, fails to establish the necessary alignment between phase context and dynamic cues.

This fundamental tension provides the physical motivation for our combination granularity design. To our knowledge, a formalized definition of combination granularity is absent from the existing multimodal literature. This work is the first to apply and rigorously evaluate this concept within the domain of strategy game automation through systematic ablation studies. A formal mathematical definition is provided in Appendix C.

**Annotation Methods and Key Strategies**

The data collection and annotation pipeline for strategy game tasks was executed as follows:

1) We recorded 82 operational videos from the target platforms.

2) Action timestamps [42] and corresponding natural language descriptions were manually annotated by reviewing the footage at 10× playback speed.

3) A Python script was employed to extract key-frame images based on these annotated timestamps.

4) The action descriptions and their associated screenshots were structured into JSON-formatted data entries of the form {image, instruction}, with analogous structures created for multi-image and video samples.

5) Key-frame selection criterion: The frame immediately preceding the execution of an action was selected as the input sample, ensuring that the model observes the pre-action interface state.

**Model Fine-tuning Strategy**

(1) Parameter-Efficient Fine-tuning Setup. To conduct our experiments under constrained GPU memory, we employed 4-bit QLoRA [53] for all fine-tuning runs, with the following hyperparameters:

1) Quantization: 4-bit, reducing VRAM consumption to approximately 7 GB.

2) Context length: 40,000 tokens, sufficient to accommodate high-resolution screenshots and lengthy instructions.

3) Per-device batch size: 1 (dictated by single-GPU memory limits).

4) Gradient accumulation: 16 steps, yielding an effective global batch size of 16.

5) Validation split: 10% of the training data, held out as a fixed validation set.

6) Logging frequency: Every 5 training steps.

7) Checkpoint frequency: Every 10 training steps.

8) LoRA configuration [61]: Rank r=2 , scaling factor α=32 , and dropout rate of 0.05.

(2) Early Stopping Criterion. The optimal number of training epochs was determined via an early stopping strategy to prevent overfitting:

1) Method: The training and validation losses (train_loss, eval_loss) were logged every 10 steps from trainer_log.jsonl to generate loss trend curves (see Appendix D).

2) Selection criterion: Training was halted at the epoch where the validation loss, after reaching its minimum, exhibited a sustained and significant increase, even as the training loss continued to decrease—a classic sign of overfitting.

The optimal epoch count for each dataset is summarized in Table.12.

Table.12. Statistics of Optimal Fine-tuning Epochs for Various Datasets

| Dataset | Optimal Epochs | Dataset | Optimal Epochs |
| --- | --- | --- | --- |
| annotions_new2.1 | 5 | combo_D1.1MVS | 27 |
| MI2.8.3 | 15 | combo_D1.1SVM | 10 |
| my_video_data | 9 | combo_D1.1SVMV | 8 |
| combo_A1.1 | 16 | MI2.7.1 | 7 |
| combo_B1.7 | 13 | MI2.7.2 | 8 |
| combo_C1.1 | 7 | MI2.7.3 | 7 |
| combo_C1.2 | 10 | MI2.7.4 | 8 |
| combo_D1.2 | 10 | MI2.7.5 | 15 |

## Ablation Experiment Design

### Experimental Objectives

Our ablation study is designed to address two core questions:

1) Modality Contribution: What is the individual and interactive impact of distinct visual modalities—static images, multi-image sequences, and videos—on VLM fine-tuning effectiveness?

2) *Combination Granularity*: Under a fixed set of modalities, how do the two proposed combination strategies, intra-sample fusion and inter-sample mixing, compare in terms of task performance, generalization, and computational efficiency?

For clarity, modality definitions and dataset compositions are detailed in **Dataset Construction** section.

### Experimental Protocol

We operationalize these objectives through two controlled experiments:

1) Modality Impact Experiment: We benchmark the performance of models trained on each single modality (S, M, V), their pairwise and full multimodal combinations, against a full-fusion baseline.

2) *Combination Granularity* Experiment: Holding the set of constituent modalities constant, we compare models trained using fusion versus mixing strategies.

### Test Set Design

To disentangle the model's capabilities along distinct cognitive axes, we employ two specialized test sets:

1) val_sum (multi-task test set): Comprising samples from all modalities, this set evaluates the model's capacity for comprehensive reasoning and generalization in complex, dynamic scenarios.

2) val_S (single-task test set): Containing only static image samples [22], this set isolates and assesses the model's proficiency in static interface element recognition and atomic action generation.

Both test sets feature identical action types and a 4-option multiple-choice question format. The sole difference lies in the input modality distribution. This orthogonal evaluation design enables a precise diagnosis of the specific source of any performance gain, moving beyond a monolithic aggregate score.

**Evaluation Metrics**

We assess model performance using two complementary metric suites:

1) Prediction Quality: BLEU-4 (predict_bleu-4), and the ROUGE family (predict_rouge: ROUGE-1, ROUGE-2, ROUGE-L). Higher scores indicate superior output quality.

2) Computational Efficiency: Model preparation time (MPT; predict_model_preparation_time), runtime per sample (Runtime; predict_runtime), samples processed per second (SAM/s; predict_samples_per_second), and steps per second (STEPS/s; predict_steps_per_second). Lower MPT and Runtime values denote greater efficiency, while higher SAM/s and STEPS/s values signify higher throughput.

# Ethical Considerations

Our research on cross-platform game operation automation raises important ethical considerations. While strategy games provide a controlled environment for studying embodied AI, we acknowledge the potential dual-use concerns when applied to military simulation scenarios. To address these concerns: (a) All training data underwent strict anonymization to remove player identifiers and sensitive game configurations; (b) We implement capability constraints that prevent autonomous execution beyond predefined safety thresholds; (c) Our open-sourced implementation includes explicit disclaimers against military applications. We advocate for responsible development frameworks where such technologies primarily serve educational purposes, game accessibility for disabled players, and stress-testing of user interface designs.

# Author contributions

Guoyan Wang: Data curation、Formal analysis、Investigation、Methodology、Software、Validation、Visualization、Writing-original draft.

Yanyan Huang: Conceptualization、Funding acquisition、Methodology、Project administration、Supervision、Writing-original draft、Writing-review & editing.

Chunlin Chen: Resources、Writing-review & editing.

Lifeng Wang: Investigation、Software、Writing-original draft.

Yuxiang Sun: Conceptualization、Funding acquisition、Resources、Methodology、Project administration、Writing-original draft、Writing-review & editing.

# Competing interests

The authors declare no competing interests.

# Supplemental Information

## Appendix A Environment

Here, the environment refers to the external environment where Yanyun-3 operates: GamePlatform A, GamePlatform B, and GamePlatform C. More adversarial game simulation environments will be incorporated later.

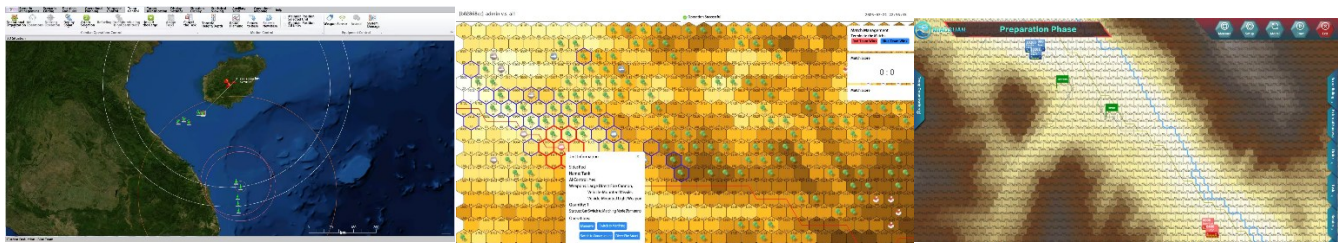

**Fig. A1**. Interface Diagrams of Various Environments

(a) GamePlatform A is a professional complex system simulation platform focused on multi-domain dynamic environment simulation. This platform can integrate land, sea, air, and electromagnetic dimensional elements, supporting rapid construction and real-time simulation of multi-layered mission scenarios. Its modular architecture allows flexible configuration of various simulation scenarios, providing technical support for intelligent decision-making research. The system supports "human-in-the-loop" open-loop full longitudinal command and control, with rich interface elements (often containing numerous operator icons, buttons, dropdown menus, input boxes, etc.), high operational precision requirements, and uneven interface changes after action execution.

(b) GamePlatform B is a strategy game Team A/B confrontation system independently developed by Nanjing University, built on a 3D Geographic Information System. It uses a hexagonal grid map, has simple interface elements, is easy to operate and learn, and exhibits large interface changes after action execution, making it highly readable.

(c) GamePlatform C platform was independently developed by the Intelligent Systems and Engineering Research Center of the Institute of Automation, Chinese Academy of Sciences. This strategy game system has been reconstructed based on cutting-edge artificial intelligence theories. It also uses a hexagonal grid map, with fewer interface elements than GamePlatform A and GamePlatform B. In small confrontation scenarios, hexagonal grid number labels are clear, facilitating automated operation.

## Appendix B Dataset Design Details and Performance

Yanyun-3's dataset originates from recorded gameplay videos on different strategy game platforms and key action frames extracted from these recordings. To map action frames to natural language, the dataset's modality *combination granularity* is divided into two types: intra-sample fusion and inter-sample mixing. Ultimately, nine types of datasets were formed: single image dataset annotions_new2.1, multi-image dataset MI2.8.3, video dataset my_video_data, single image * video dataset combo_A1.1, multi-image * video dataset combo_B1.7, single image * multi-image dataset combo_C1.2, single image * multi-image * video dataset combo_D1.2, single image * video + multi-image dataset combo_D1SVM, and multi-image * video + single image dataset combo_D1M*VS. Their composition structure is shown in Fig. B1:

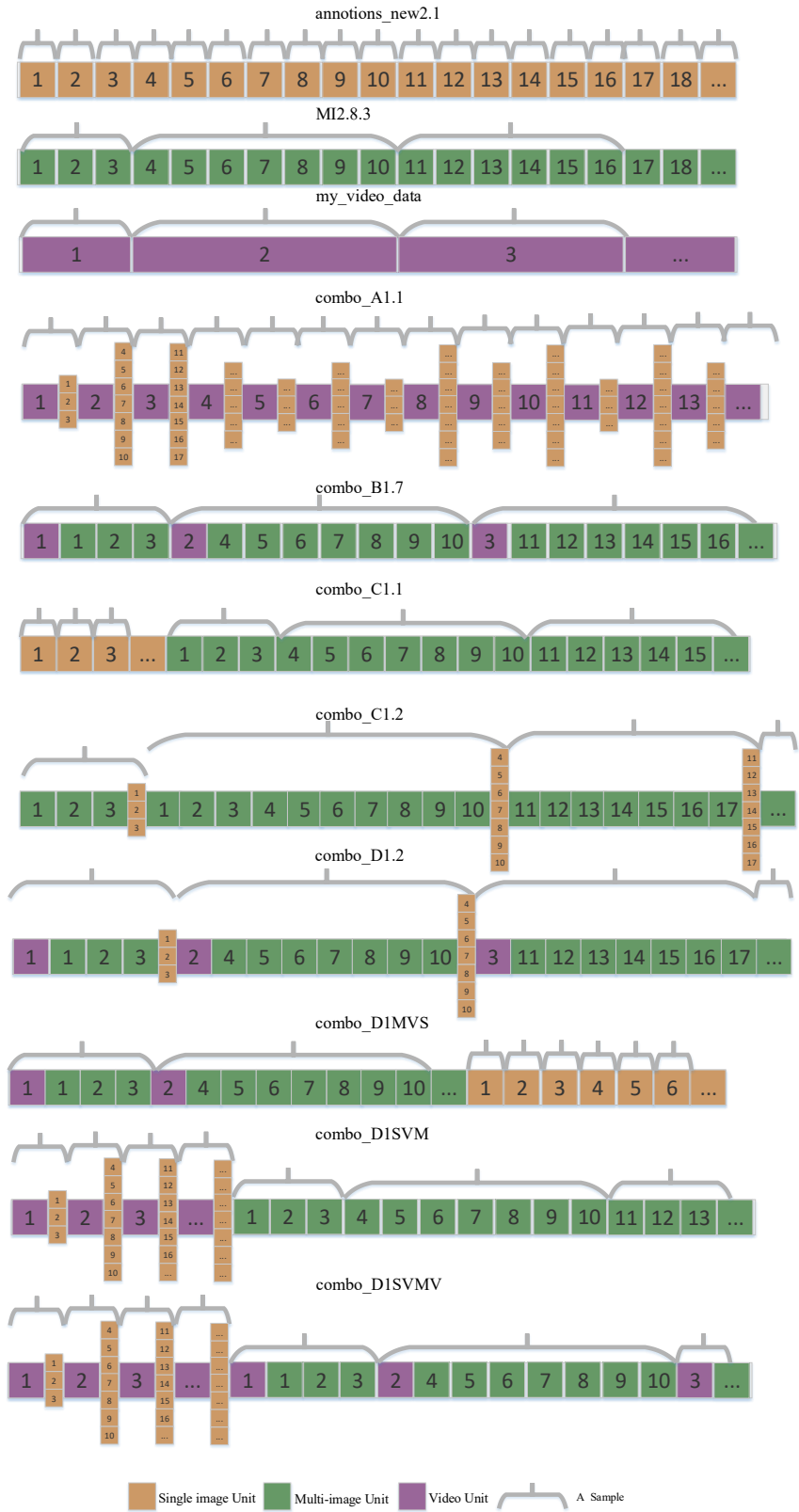

**Fig. B1**. Composition of Various Datasets. Yellow blocks represent single images, green represents multi-images, red represents video, and one bracket represents one sample.

## B1 Single Image Dataset

The single image dataset annotions_new2.1 (containing 1002 samples), denoted by symbol S, corresponds to single actions. The purpose of designing this dataset is to teach the large model to recognize static actions, such as:

click the red [Ground Unit] icon. Since single images map to actions, it belongs to the action-level dataset. It can learn the state at a certain moment but cannot capture the dynamic process of actions.

## B2 Multi-image Dataset

During the experiment, several versions of the multi-image dataset were formed, denoted by symbol M. During the ablation experiment, when fusing video with images, it was discovered that multiple images involved in one task could be placed within a single sample, forming version MI2.8.3 (containing 180 samples). Such a multi-image dataset is a task-level dataset. The MI2.8.3 dataset involves phased changes in actions.

## B3 Video Dataset

Initially, the recordings of operations were complete scenario recordings, hoping the large model could learn the entire scenario's operations. However, the current Qwen2.5-VL-7B base model cannot handle long videos for fine-tuning. Therefore, longer original recordings (e.g., the process of eliminating other sides on GamePlatform B) were broken down into task-level segments, while shorter ones (e.g., the resource allocation process on GamePlatform A) remained unchanged, forming a task-level video dataset my_video_data (containing 180 samples), denoted by symbol V. my_video_data contains the most complete information, involving continuous dynamic actions.

## B4 Single Image * Video Dataset

The single image * video dataset combo_A1.1 arranges video dialogues sequentially with multiple groups of single images from the video into one sample; multiple samples form this dataset, denoted by symbol S*V.

## B5 Multi-image * Video Dataset

The multi-image * video dataset combo_B1.7 forms one sample by combining video dialogues with a group of images from the video, with multiple samples forming this dataset, denoted by symbol M*V.

## B6 Single Image + Multi-image Dataset

Since single image and multi-image can be mixed, the entire sample set from the single image dataset annotations_new2.1 and the entire sample set from the multi-image dataset MI2.8.3 form the single image + multi-image dataset combo_C1.1, denoted by symbol S+M. As this dataset employs a mixing method with two modalities, it was not included in the ablation experiment, only its performance was recorded.

## B7 Single Image * Multi-image Dataset

By fusing single image data into multi-image dataset samples, the single image * multi-image dataset combo_C1.2 was formed, denoted by symbol S*M.

## B8 Single Image * Multi-image * Video Dataset

Here, single image data from the video * multi-image sample is added to the combo_B1.7 sample, forming combo_D1.2, denoted by symbol S*M*V.

## B9 Single Image * Video + Multi-image Dataset

Here, all samples from the single image * video dataset combo_A1.1 and all samples from the multi-image dataset MI2.8.3 are mixed to form combo_D1SVM, denoted by symbol S*V+M.

## B10 Multi-image * Video + Single Image Dataset

All samples from the multi-image video fusion dataset combo_B1.7 and all samples from the single image dataset annotations_new2.1 are mixed to form combo_D1MVS, denoted by symbol M*V+S.

# B11 Single Image * Video + Multi-image * Video

The single-image–video fused dataset combo_A1.1 and the multi-image–video fused dataset combo_B1.7 were mixed to form combo_D1SVMV, denoted symbolically as S*V+M*V. Since the video modality (V) is used twice in this combination, this dataset was not included in the ablation experiments; its performance was recorded only for reference. However, as shown in Table B1, the performance of S*V+M*V (i.e., combo_D1SVMV)—with BLEU-4 scores of 42.71 (on val_sum) and 62.41 (on val_S)—while substantially higher than that of the full-fusion baseline SMV (8.42 / 4.81), remains inferior to the MV+S strategy (47.79 / 62.41), particularly on the val_sum test set. This indicates that although increasing modality data volume (by duplicating V) can enhance performance, a well-designed *combination granularity* strategy (MV+S) is still superior.

# B12 Performance of Various Datasets on Different Test Sets

Performance heatmaps for each dataset on val_sum and val_S are shown below:

Table B1. Ablation Experiment Metric Statistics Table

| Metrics | 0 | S | V | M | S*V | M*V |
|---|---|---|---|---|---|---|
| BLEU-4(val2_sum) | 0.7789 | 34.82248038 | 1.900509811 | 11.68992981 | 12.29108528 | 3.129674717 |
| MPT(val2_sum) | 0.012 | 0.0094 | 0.0203 | 0.0095 | 0.0134 | 0.0185 |
| ROUGE-1(val2_sum) | 7.2446 | 51.49984 | 16.87441736 | 33.93001962 | 31.72986642 | 16.00892491 |
| ROUGE-2(val2_sum) | 0.5371 | 37.64772075 | 3.389051698 | 13.83404189 | 11.11771019 | 3.519746038 |
| ROUGE-L(val2_sum) | 3.5285 | 48.38567547 | 7.556666038 | 20.69638943 | 27.40327321 | 8.956122264 |
| RT(val2_sum) | 2828.3704 | 729.108 | 2360.4538 | 2306.4003 | 786.4008 | 2480.9239 |
| SAM/s(val2_sum) | 0.094 | 0.363 | 0.112 | 0.115 | 0.337 | 0.107 |
| STEPS/s(val2_sum) | 0.047 | 0.182 | 0.056 | 0.058 | 0.169 | 0.054 |
| BLEU-4(val2_S) | 0.5661 | 45.84147612 | 1.246093532 | 6.323039303 | 13.0366995 | 1.1214 |
| MPT(val2_S) | 0.0093 | 0.011 | 0.0123 | 0.0093 | 0.0142 | 0.0115 |
| ROUGE-1(val2_S) | 5.4935 | 60.33670448 | 12.97220547 | 28.130101 | 28.85093134 | 11.0532 |
| ROUGE-2(val2_S) | 0.4303 | 46.71890498 | 1.781195522 | 7.540502488 | 7.818989055 | 1.2318 |
| ROUGE-L(val2_S) | 2.2712 | 60.12011642 | 4.710976119 | 14.65818856 | 27.4008806 | 4.506 |
| RT(val2_S) | 1924.7092 | 177.7917 | 1414.4485 | 1542.4098 | 225.4404 | 1133.4459 |
| SAM/s(val2_S) | 0.104 | 1.131 | 0.142 | 0.13 | 0.892 | 0.177 |
| STEPS/s(val2_S) | 0.052 | 0.568 | 0.071 | 0.065 | 0.448 | 0.089 |
| Metrics | S+M | S*M | S*M*V | M*V+S | S*V+M. | S*V+M*V |
| BLEU-4(val2_sum) | 36.33932264 | 9.2374 | 8.4171 | 47.7851 | 10.5407 | 42.7121 |
| MPT(val2_sum) | 0.0262 | 0.007 | 0.0095 | 0.0101 | 0.018 | 0.0103 |
| ROUGE-1(val2_sum) | 55.4813917 | 31.0685 | 28.3595 | 65.2271 | 30.4147 | 60.4575 |
| ROUGE-2(val2_sum) | 40.55376642 | 11.1705 | 9.5328 | 53.2668 | 13.7064 | 47.0225 |
| ROUGE-L(val2_sum) | 50.7959483 | 18.7006 | 17.6586 | 61.0064 | 20.2766 | 56.3925 |
| RT(val2_sum) | 645.2474 | 1178.9005 | 1109.5285 | 724.383 | 1436.4456 | 306.2487 |
| SAM/s(val2_sum) | 0.411 | 0.225 | 0.239 | 0.366 | 0.184 | 0.922 |
| STEPS/s(val2_sum) | 0.206 | 0.113 | 0.12 | 0.184 | 0.047 | 0.233 |
| BLEU-4(val2_S) | 47.57232935 | 5.3338 | 4.8138 | 62.4146 | 4.3676 | 55.5289 |
| MPT(val2_S) | 0.0098 | 0.0071 | 0.0069 | 0.0245 | 0.0168 | 0.0126 |
| ROUGE-1(val2_S) | 62.0630398 | 26.6328 | 23.4552 | 74.8567 | 21.8946 | 68.5681 |
| ROUGE-2(val2_S) | 47.81669403 | 6.7652 | 4.9099 | 64.1477 | 6.1426 | 55.9845 |

| ROUGE-L(val2_S) | 61.30271294 | 13.2112 | 12.8416 | 74.5948 | 11.7394 | 68.3433 |
| RT(val2_S) | 186.3478 | 747.9945 | 596.618 | 220.4956 | 811.8776 | 93.5204 |
| SAM/s(val2_S) | 1.079 | 0.269 | 0.337 | 0.912 | 0.248 | 2.149 |
| STEPS/s(val2_S) | 0.542 | 0.135 | 0.169 | 0.458 | 0.063 | 0.545 |

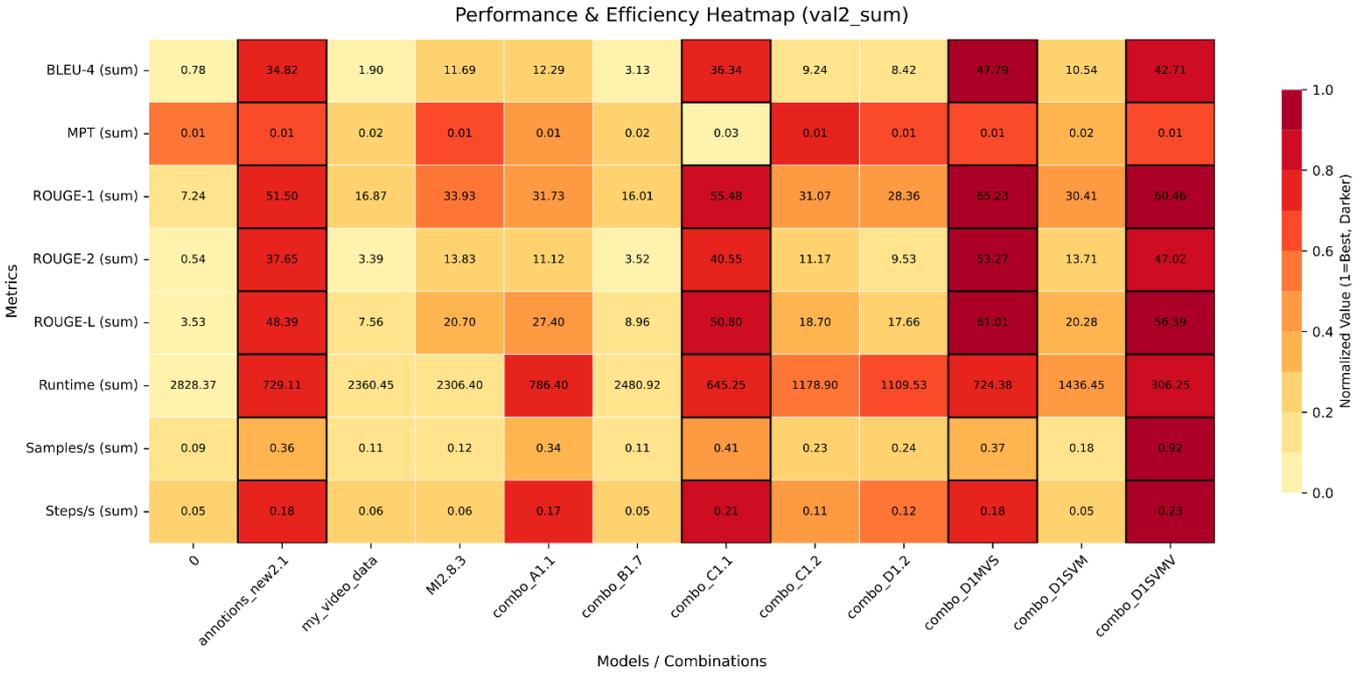

**Fig. B2**. Performance Heatmap of Various Datasets on val_sum

Performance radar charts for each dataset on val_sum are shown below:

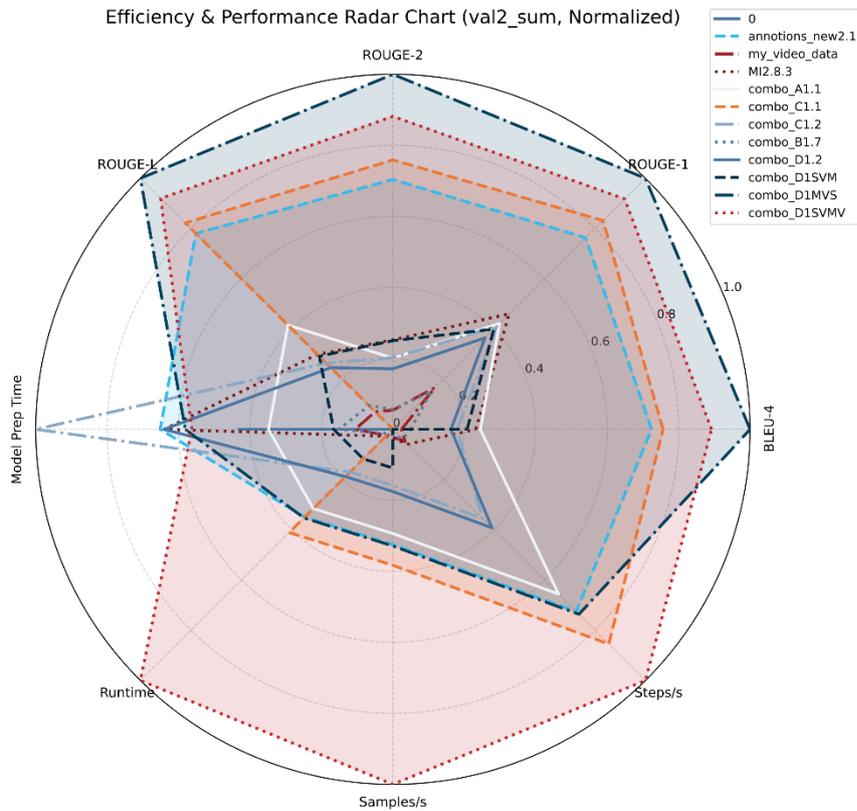

**Fig. B3**. Performance Radar Chart of Various Datasets on val_sum

Performance heatmaps for each dataset on val_S are shown below:

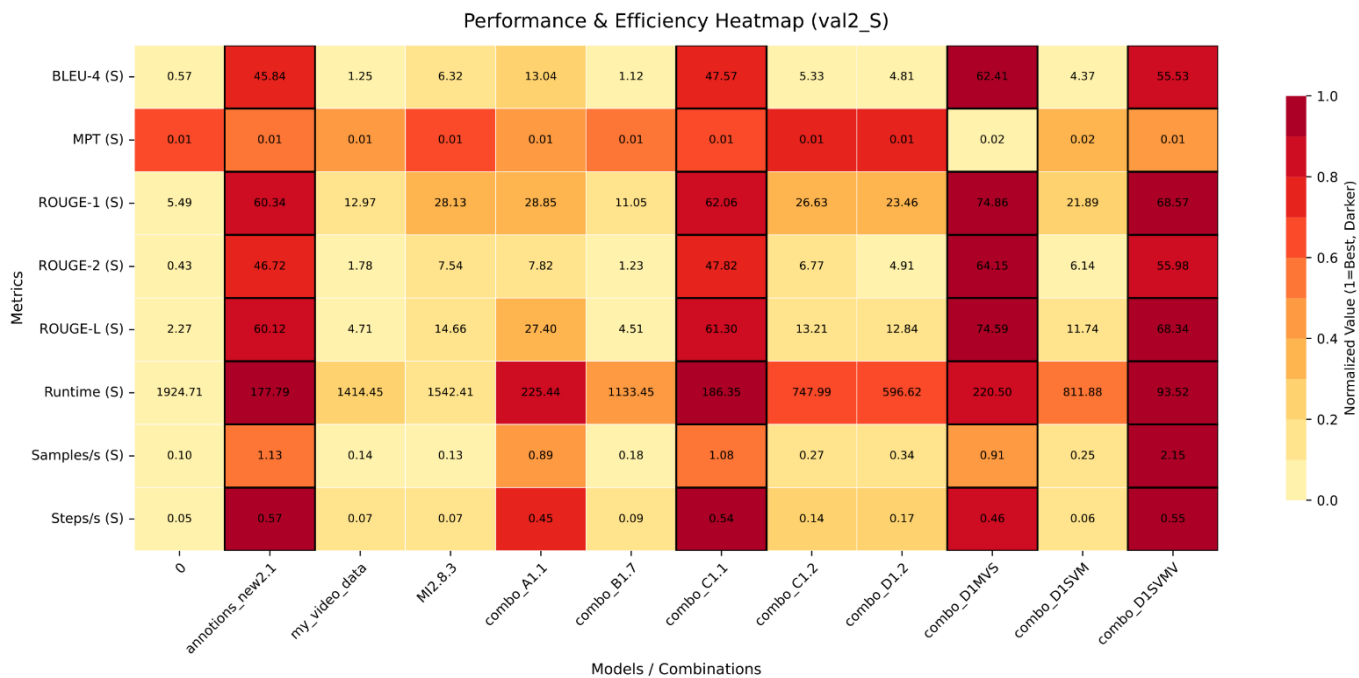

**Fig. B4**. Performance Heatmap of Various Datasets on val_S

Performance radar charts for each dataset on val_S are shown below:

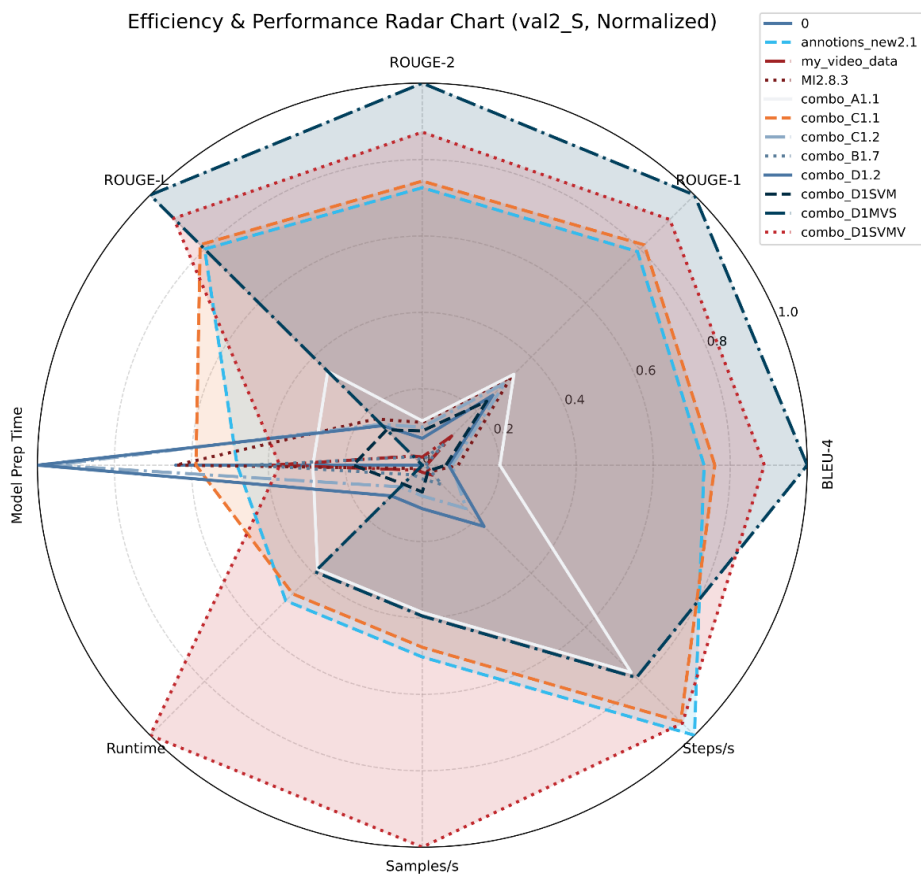

**Fig. B5**. Performance Radar Chart of Various Datasets on val_S

As illustrated by the respective portions of the training sets in Fig. B3 and Fig. B5, the combo_D1SVMVtraining set demonstrates superior overall performance across various metrics. However, as it utilizes the video modality twice, a direct and detailed comparison with other training sets is not conducted here. Nonetheless, this finding suggests a potential direction for future research.

## B13 Efficiency Ablation Experiment Supplement for Modal Ablation

Table B2. Multi-task Test Set Dataset Type Ablation Experiment Statistics for Efficiency

| Exp.NO. | S | M | V | MPT↓ (val_S) | PD | RT↓ (val_S) s | PD | SAM/s↑ (val_S) | PD | STEPS/s↑ (val_S) | PD |
|---|---|---|---|---|---|---|---|---|---|---|---|
| 1 | √ | √ | √ | 0.0095 | 0% | 1109.5285 | 0% | 0.239 | 0% | 0.12 | 0% |
| 2 | √ | √ | × | 0.0070 | 26.32 % | 1178.9005 | -6.25 % | 0.225 | 5.86 % | 0.113 | 5.83 % |
| 3 | √ | × | √ | 0.0134 | -41.05 % | 786.4008 | 29.12 % | 0.337 | -41.00 % | 0.169 | -40.83 % |
| 4 | × | √ | √ | 0.0185 | -94.74 % | 2480.9239 | -123.61 % | 0.107 | 55.23 % | 0.054 | 55.00 % |
| 5 | √ | × | × | 0.0094 | 1.05 % | 729.108 | 34.29 % | 0.363 | -51.88 % | 0.182 | -51.67 % |
| 6 | × | √ | × | 0.0095 | 0 % | 2306.4003 | -107.87 % | 0.115 | 51.88 % | 0.058 | 51.67 % |
| 7 | × | × | √ | 0.0203 | -113.68 % | 2360.4538 | -112.74 % | 0.112 | 53.14 % | 0.056 | 53.33 % |
| 8 | × | × | × | 0.0120 | -26.32 % | 2828.3704 | -154.90 % | 0.094 | 60.67 % | 0.047 | 60.83 % |

Table B3. Single-task Test Set Dataset Type Ablation Experiment Statistics for Efficiency

| Exp.NO. | S | M | V | MPT↓ (val_S) | PD | RT↓ (val_S) s | PD | SAM/s↑ (val_S) | PD | STEPS/s↑ (val_S) | PD |
|---|---|---|---|---|---|---|---|---|---|---|---|
| 9 | √ | √ | √ | 0.0069 | 0% | 596.618 | 0% | 0.337 | 0% | 0.169 | 0% |
| 10 | √ | √ | × | 0.0071 | -2.90 % | 747.9945 | -25.37 % | 0.269 | 20.18 % | 0.135 | 20.12 % |
| 11 | √ | × | √ | 0.0142 | -105.80 % | 225.4404 | 62.22 % | 0.892 | -164.69 % | 0.448 | -165.09 % |
| 12 | × | √ | √ | 0.0115 | -66.67 % | 1133.4459 | -89.97 % | 0.177 | 47.48 % | 0.089 | 47.34 % |
| 13 | √ | × | × | 0.0110 | -59.42 % | 177.7917 | 70.22 % | 1.131 | -235.61 % | 0.568 | -236.09 % |
| 14 | × | √ | × | 0.0093 | -34.78 % | 1542.4098 | -158.51 % | 0.130 | 61.42 % | 0.065 | 61.54 % |
| 15 | × | × | √ | 0.0123 | -78.26 % | 1414.4485 | -136.99 % | 0.142 | 57.86 % | 0.071 | 57.99 % |
| 16 | × | × | × | 0.0093 | -34.78 % | 1924.7092 | -222.61 % | 0.104 | 69.14 % | 0.052 | 69.23 % |

From the above two tables, it can be seen:

(1) Single image is the efficiency core. Single image configured alone has the shortest inference time and highest throughput (first in both test sets).

(2) Multi-image is a time-consuming burden. Removing multi-image (experiments 3/11) significantly improves efficiency in both datasets. Multi-image configured alone (experiments 6/14) has very poor efficiency.

(3) Video's impact on efficiency is unstable. Removing video does not significantly speed up in either test set, and some metrics are even worse (10/2). Video configured alone (7/15) is less efficient than the full modality.

(4) No modality performs worst. RT and throughput are the worst configuration in both datasets.

Conclusion:

(1) The single image dataset is the most efficient strategy, with the fastest inference speed and highest throughput.

(2) Multi-image is the main time-consuming modality; removing it can significantly speed up.

(3) Single image is an indispensable modality; omitting it drastically reduces efficiency.

(4) Video's impact on efficiency is less significant than single image and multi-image, but the video dataset is still inefficient.

## B14 Efficiency Ablation Experiment Supplement for *Combination Granularity*

Table E12 was condensed to yield B4 and B5.

**Table B4**. Comparative Experiment Statistics for Different Fusion/Mixing Datasets on the Multi-task Test Set for Efficiency

| Exp.NO. | S | M | V | MPT↓ (val_sum) | PD | RT↓ s (val_sum) | PD | SAMPLE/s↑ (val_sum) | PD | STEPS/s↑ (val_sum) | PD |
|---|---|---|---|---|---|---|---|---|---|---|---|
| 1 | * | * | * | 0.0095 | 0% | 1109.5285 | 0% | 0.239 | 0% | 0.12 | 0% |
| 17 | * | + | * | 0.018 | -89.47% | 1436.4456 | -29.46% | 0.184 | 22.98% | 0.047 | 60.83% |
| 18 | + | * | * | 0.0101 | -6.32% | 724.383 | 34.71% | 0.366 | -53.14% | 0.184 | -53.33% |

**Table B5**. Comparative Experiment Statistics for Different Fusion/Mixing Datasets on the Single-task Test Set for Efficiency

| Exp.NO. | S | M | V | MPT↓ (val_S) | PD | RT↓ s (val_S) | PD | SAMPLE/s↑ (val_S) | PD | STEPS/s↑ (val_S) | PD |
|---|---|---|---|---|---|---|---|---|---|---|---|
| 9 | * | * | * | 0.0069 | 0% | 596.618 | 0% | 0.337 | 0% | 0.169 | 0% |
| 19 | * | + | * | 0.0168 | -143.48 % | 811.8776 | -36.08 % | 0.248 | 26.41 % | 0.063 | 62.72 % |
| 20 | + | * | * | 0.0245 | -255.07 % | 220.4956 | 63.04 % | 0.912 | -170.62 % | 0.458 | -171.01 % |

From the above two tables, we know:

(1) Ranking is consistent across both test sets: M*V+S > full fusion (S*M*V) > S*V+M ② Reasons for M*V+S efficiency Multi-image and video fused within sample → Maximize utilization of multimodal temporal + keyframe complementary advantages, share visual encoding, reduce redundant computation. Single image appears independently → Static information extraction is done independently, avoiding redundant encoding resources wasted by fusing single image with multi-image.

(2) Reasons for S*V+M inefficiency Multi-image independent → Loses direct complementarity with video, needs separate visual encoding; Brings redundancy to input and attention computation, significantly increasing inference and loading time.

(3) MPT difference significance is minor Although MPT shows obvious differences (especially M*V+S highest in val_S), it accounts for a very small proportion of total time-consuming, so core conclusions are mainly determined by RT, SAMPLE/s, STEPS/s.

Conclusion: Under fixed three-modality set conditions, changing modality *combination granularity* significantly impacts inference efficiency. Analysis of both multi-task and single-task test sets consistently shows.

(1) M*V+S is the most efficient strategy, reducing inference time by 34.7% and 63.0% compared to full fusion, and increasing throughput by 53.1% and 170.6%, while also performing best in quality metric ablation.

(2) S*M*V has medium efficiency, stable quality.

(3) S*V+M shows increased loading and inference time and decreased throughput in both test sets, and should be avoided.

## B15 S*M and S+M Dataset Comparison Materials

In the *combination granularity* ablation experiments in *Combination Granularity* Ablation Experiment, the performance of three datasets with identical modalities but different combination granularities was compared. Actually, the two datasets S*M and S+M can also be compared. However, including these two datasets in the ablation would introduce changes in both modality and *combination granularity*, which does not meet the requirements of an ablation experiment. Therefore, the comparison is only made here.

Table B6. Prediction Quality Comparison of the 2 Datasets on Different Test Sets

| Test_Set | Dataset | Symbol | BLEU-4 ↑ % | ROUGE-1 ↑ % | ROUGE-2 ↑ % | ROUGE-L ↑ % |
|---|---|---|---|---|---|---|
| val_sum | combo_C1.1 | S+M | 36.34 | 55.48 | 40.55 | 50.80 |
|  | combo_C1.2 | S*M | 9.2374 | 31.0685 | 11.17 | 18.70 |
| val_S | combo_C1.1 | S+M | 47.57 | 62.06 | 47.82 | 61.30 |
|  | combo_C1.2 | S*M | 5.33 | 26.63 | 6.77 | 13.21 |

Table B7. Efficiency Comparison of the 2 Datasets on Different Test Sets

| Test_Set | Dataset | Symbol | MPT ↓ | RT ↓ s | SAMPLE/s ↑ | STEPS/s ↑ |
|---|---|---|---|---|---|---|
| val_sum | combo_C1.1 | S+M | 0.0262 | 645.2474 | 0.411 | 0.206 |
|  | combo_C1.2 | S*M | 0.007 | 1178.9005 | 0.225 | 0.113 |
| val_S | combo_C1.1 | S+M | 0.0098 | 186.3478 | 1.079 | 0.542 |
|  | combo_C1.2 | S*M | 0.0071 | 747.9945 | 0.269 | 0.135 |

From the above two tables, we know:

(1) Quality Performance

Regardless of the multi-task set (val_sum) or single-task set (val_S), S+M leads S*M in all BLEU/ROUGE metrics by a huge margin, with an even more exaggerated gap in the single-task set. Reason speculation: S*M intra-sample fusion forcibly aligns single image and multi-image, easily introducing visual information redundancy and semantic conflicts (especially since single image is often part of multi-image), causing model attention dispersion and alignment difficulties; whereas S+M mixing allows the model to learn features of the two modalities independently, resulting in purer training signals.

(2) Efficiency Performance

S+M's inference total time (RT) is significantly lower than S*M, with throughput increasing 2 to 4 times. This indicates that inter-sample mixing reduces the visual input volume per sample during inference, lightening the visual encoding burden and thus improving speed and throughput.

(3) Application Recommendation

If the goal is high quality and efficiency (especially in scenarios with high real-time inference requirements), for the combination of single image and multi-image modalities, the inter-sample mixing (S+M) strategy should be prioritized over intra-sample fusion (S*M).

Intra-sample fusion may be more suitable for modalities with strong complementarity and low redundancy (e.g., video+multi-image), not for highly correlated static modalities (single image+multi-image).

# Appendix C Formalization of Dataset *Combination Granularity*

## C1 Basic Definitions and Modal Correlation Logic
### C1.1 Modal Hierarchy and Derivative Relationships

For the three types of modalities involved in cross-modal mutual information calculation—Static Image (S), Multi-image Sequence (M), and Video (V)—a hierarchical derivative relationship and mathematical representation are defined based on the dataset characteristics that "static images are derived from video key frames and multi-image sequences are collections of video key transition frames":

1) Video Modality (V): Dynamic Scene Carrier

Defined as the "complete recording carrier of dynamic information in game scenes", denoted as

$$V = \{v_1, v_2, ..., v_{N_V}\}, \text{ where:}$$

① $v_k$: The $k$-th video sample (corresponding to one game scene segment), $k \in [1, N_V]$, and $N_V = 180$ in the experiment;

② Each $v_k$ contains $T_k$ consecutive frames ($T_k \geq 5$), divided into key transition frames and ordinary frames:

(i) Key transition frame set: $F_k^V = \{f_{k,1}^V, f_{k,2}^V, ..., f_{k,K_k}^V\}$, where $K_k$ is the number of key frames (3–6 frames, determined by the dynamic complexity of the scene, such as frames of actions like weapon switching and vehicle movement);

(ii) Ordinary frame set: $G_k^V = \{g_{k,1}^V, ..., g_{k,T_k-K_k}^V\}$, which are transition frames without core actions in the scene;

③ The scene overlap degree $\omega(v_k, v_l) \in [0,1]$ is defined to represent the proportion of the intersection of scene features between $v_k$ and $v_l$:

$$\omega(v_k, v_l) = \frac{|\text{Feat}(v_k) \cap \text{Feat}(v_l)|}{\min(|\text{Feat}(v_k)|, |\text{Feat}(v_l)|)}$$

Among them, Feat($v$) is the set of scene features of video $v$ (including map coordinates, action labels, and unit status), and $\omega_{\text{avg}} \leq 0.3$ in the experiment (average overlap degree $\leq 30\%$);

④ Information integrity: 100% (including dynamic continuity and static details), making it the highest-granularity modality.

2) Multi-image Sequence Modality (M): Subset of Video Key Frames

Defined as the "collection carrier of video key transition frames" and the dynamic core subset of the video modality, denoted as $M = \{m_1, m_2, ..., m_{N_M}\}$, where:

① $N_M = N_V = 180$ (1:1 scene alignment with video samples);

② The $k$-th multi-image sample $m_k$ and the $k$-th video sample $v_k$ satisfy the derivative relationship: $m_k = F_k^V$ (that is, $m_k$ is the complete set of key transition frames of $v_k$ without any frame omission);

③ Frame order and content consistency: $m_k = \{f_{k,1}^M, f_{k,2}^M, ..., f_{k,K_k}^M\}$, and $f_{k,i}^M = f_{k,i}^V$ ($i \in [1, K_k]$, the timing and content of multi-image frames are completely consistent with video key frames);

④ Scene overlap transmission: If $v_k$ overlaps with $v_l$ ($\omega(v_k, v_l) > 0$), then $\omega(m_k, m_l) = \omega(v_k, v_l)$ (the overlap degree of multi-image sequences is inherited from the corresponding videos);

⑤ Information integrity: 60%–80% (only retains dynamic transition information and eliminates redundancy of ordinary frames).

3) Static Image Modality (S): Disassembled Subset of Multi-image Frames

Defined as the "full disassembly carrier of multi-image sequence frames" and the frame-level subset of the multi-image modality, denoted as $S = \{s_1, s_2, ..., s_{NS}\}$, where:

① $N_S = \sum_{k=1}^{N_M} K_k = 1002$ (the total number of static images is the sum of the number of key frames of all multi-image samples);

② The $k$-th multi-image sample $m_k$ is disassembled into a static image subset $S_k = \{s_{t_k+1}, s_{t_k+2}, ..., s_{t_k+K}\}$, where:

(i) $t_k = \sum_{i=1}^{k-1} K_i$ (the starting index of static images corresponding to the $k$-th multi-image sample, $t_1 = 0$);

(ii) Frame content consistency: $s_{t_k+i} = f_{k,i}^M$ ($i \in [1, K_k]$, the content of static images is completely consistent with multi-image frames, and they are disassembled in the order of multi-image frames);

③ Scene overlap inheritance: The overlap degree $\omega(s_j, s_t)$ between static image $s_j \in S_k$ and $s_t \in S_l$ is equal to $\omega(m_k, m_l)$ (inherited from the corresponding multi-image samples);

④ Information integrity: 15%–30% (only retains single-frame static information without dynamic correlation information), making it the lowest-granularity modality.

### C1.2 Core Operations of Combination Granularity: Fusion and Mixing

Combination granularity distinguishes between intra-sample fusion (denoted as "*") and inter-sample mixing (denoted as "+"). The formal definitions and applicable scenarios of these two types of operations are as follows:

1) Intra-sample Fusion (Fusion, *)

① Definition

Integrate multi-modal data within a single sample to generate a unified feature representation. It is suitable for complementary alignment between dynamic modalities (such as M and V), with the goal of retaining temporal continuity and cross-modal collaborative information.

② Mathematical Representation

Assume that a sample $x$ contains data from $M$ modalities $x^{(1)}, x^{(2)}, ..., x^{(M)}$ (such as key frame sequences of M and video segments of V). The fusion process consists of three steps:

(i) Feature extraction: Map to the feature space through a modality-specific extractor $f^{(m)}$:

$$h^{(m)} = f^{(m)}(x^{(m)})$$

Among them, $h^{(m)}$ is the feature of the $m$-th modality (such as the frame sequence feature $h^{(M)}$ of M and the video temporal feature $h^{(V)}$ of V);

(ii) Modal alignment: Map multi-modal features to a shared space through an alignment function $a$ to eliminate differences in dimensions and distributions:

$$h_{align}^{(m)} = a(h^{(m)})$$

In the experiment, the alignment method is "feature dimension unification" (S, M, and V are all mapped to 1024 dimensions);

(iii) Fusion generation: Generate a unified representation $z$ through a fusion function $g$ (such as attention weighting and element-wise addition):

$$z = g(h_{align}^{(1)} * h_{align}^{(2)} * ... * h_{align}^{(M)})$$

③Applicable Scenarios

(i) Fusion between dynamic modalities (M*V): M provides phase information of discrete key frames, and V provides continuous temporal dynamics. After fusion, the "phase-dynamics" collaborative relationship is retained, corresponding to the dataset combo_B1.7 in the experiment;

(ii) Fusion between static and dynamic modalities (S*V): S provides a high signal-to-noise ratio static anchor, and V provides dynamic context. After fusion, the collaboration between "static positioning and dynamic reasoning" is optimized, corresponding to the dataset combo_A1.1 in the experiment.

2) Inter-sample Mixing (Mixing, +)

①Definition

Merge sample sets of different modalities at the dataset level without changing the modal composition of a single sample. It is suitable for combination of static and dynamic modalities (such as S and M*V), with the goal of retaining the high-purity information of static modalities and avoiding interference from dynamic redundancy.

②Mathematical Representation

Assume that the sample set of modality $X$ is $D_X = \{x_1, x_2, ..., x_{N_X}\}$, and the sample set of modality $Y$ is $D_Y = \{y_1, y_2, ..., y_{N_Y}\}$. The mixed dataset is:

$$D_{X+Y} = D_X \bigcup D_Y = \{(x_1,t_1),...,(x_{N_X},t_{N_X}),(y_1,t'_1),...,(y_{N_Y},t'_{N_Y})\}$$

Among them, $t$ is the sample label (action instruction), and it is required that the labels of samples in the same scene are consistent (for example, the labels of S and M*V samples in the same scene are both "click the red tank").

③Applicable Scenarios

(i) Static-dynamic mixing (M*V+S): First perform intra-sample fusion of M and V (to retain dynamic collaboration), then perform inter-sample mixing with S (to retain the static purity of S), corresponding to the optimal dataset combo_D1MVS in the experiment;

(ii) Mixing between static modalities (S+M): Merge S and M as independent sample sets respectively, which is suitable for scenarios where it is necessary to enhance the diversity of static samples, corresponding to the dataset combo_C1.1 in the experiment.

## C1.3 Information Theory Foundation: Mutual Information Formula with Overlap Correction

Based on the Gaussian mutual information framework and combined with scene overlap characteristics, the cross-modal mutual information calculation is corrected to quantify the strength of information correlation between modalities.

1) Core Formula

Assume that the features of modality $X$ (such as S) and $Y$ (such as M/V) follow Gaussian distributions $N(\mu_X, \Sigma_X)$ and $N(\mu_Y, \Sigma_Y)$, and their joint distribution is $N(\mu_{X,Y}, \Sigma_{X,Y})$. The mutual information considering scene overlap is:

$$I(X;Y) = \frac{1}{2}\log_2\left(\frac{|\Sigma_X| \cdot |\Sigma_Y|}{|\Sigma_{X,Y}|}\right) \times (1 - \frac{\omega_{avg}}{2})$$

Where:

① $\Sigma_X$: Weighted covariance matrix of modality $X$ (considering differences in sample contribution, with a weight of 0.2 for static images, 0.8 for multi-image sequences, and 1.0 for videos);

② $\Sigma_{X,Y}$: Joint covariance matrix of $X$ and $Y$;

③ $|\cdot|$: Matrix determinant;

④ $(1-\omega_{avg}/2)$: Scene overlap correction factor ($\omega_{avg}$ is the average scene overlap degree, taken as 0.3 in the experiment, and the correction factor is 0.85).

2) Theorems and Corollaries

①Theorem C1.1 Mutual Information Ordering of Modal Derivative Relationships

Based on the derivative relationship of "V→M→S", the cross-modal mutual information satisfies: $I(S;V) > I(S;M)$.

②Proof:

(i) S is a subset of M, and M is a subset of V. Therefore, the feature overlap degree between S and V is higher than that between S and M, that is, $|\Sigma_{S,V}| < |\Sigma_{S,M}|$;

(ii) Substituting into the formula, we get $\log_2(\frac{|\Sigma_S||\Sigma_V|}{|\Sigma_{S,V}|}) > \log_2(\frac{|\Sigma_S||\Sigma_M|}{|\Sigma_{S,M}|})$, and the correction factors are the same. Thus, $I(S;V) > I(S;M)$, which corresponds to $I(S;V) = 1.51$ bits and $I(S;M) = 1.24$ bits in the experiment.

②Corollary C1.1 Mutual Information Upper Bound of Scene Overlap

The upper bound of mutual information decreases as the overlap degree increases:

$$I(X;Y) \leq \frac{1}{2}\log_2\left(\frac{|\Sigma_X|\cdot|\Sigma_Y|}{|\Sigma_{X,Y}|}\right)$$

The equality holds when $\omega_{avg} = 0$ (no overlap). In the experiment, due to $\omega_{avg} = 0.3$, the actual mutual information is 15% lower than that without overlap.

## C2 Composite Combination Strategies and Mathematical Representations

Based on basic operations (*, +), three core composite strategies are defined to cover the construction logic of all datasets in the experiment:

### C2.1 Full Intra-sample Fusion (S*M*V*)

1) Definition

Perform full intra-sample fusion of the three modalities S, M, and V, that is, a single sample contains static frames, key frame sequences, and video segments simultaneously, corresponding to the dataset combo_D1.2 in the experiment.

2) Mathematical Representation

$$z = g\left(a(f(S)) * a(f(M)) * a(f(V))\right)$$

Among them, $f(S), f(M)$, and $f(V)$ are the feature extraction results of S, M, and V respectively.

3) Applicable Limitations

Due to high redundancy between S and M/V (S is a frame subset of M), feature confusion is likely to occur after fusion. In the experiment, its BLEU-4 score (8.42%) is much lower than that of the optimal strategy (47.79% for M*V+S).

### C2.2 Partial Fusion + Inter-sample Mixing (S*V+M)

1) Definition

First perform intra-sample fusion of S and V (to retain static-dynamic collaboration), then perform inter-sample mixing with M (to supplement multi-image phase information), corresponding to the dataset combo_D1SVM in the experiment.

2) Mathematical Representation

$$D_{S*V+M} = \{g(a(f(S_k)) * a(f(V_k)))\}_{k=1}^{180} \cup \{f(M_k)\}_{k=1}^{180}$$

3) Applicable Limitations

The temporal correlation between M and V is broken (M is not fused with V), resulting in fragmented dynamic information. In the experiment, its BLEU-4 score (4.37%) on the val_S test set is only 7% of that of the optimal strategy (62.41%).

### C2.3 Dynamic Fusion + Static Mixing (M*V+S)

1) Definition

First perform intra-sample fusion of M and V (to retain dynamic collaboration), then perform inter-sample mixing with S (to retain static purity), corresponding to the optimal dataset combo_D1MVS in the experiment.

2) Mathematical Representation

$$D_{M*V+S} = \{g(a(f(M_k)) * a(f(V_k)))\}_{k=1}^{180} \cup \{f(S_{k,i})\}_{k=1,i=1}^{180,K_k}$$

Among them, $S_{k,i}$ is the $i$-th static image of the $k$-th multi-image sample.

3) Advantage Verification

①Prediction quality: BLEU-4 = 47.79% on the val_sum test set, which is 5.7 times that of the full fusion strategy; BLEU-4 = 62.41% on the val_S test set, which is 12.98 times that of the full fusion strategy;

②Computational efficiency: The inference time (220.49s) on the val_S test set is 63% lower than that of the full fusion strategy (596.62s), and the sample throughput (0.912 SAM/s) is increased by 170%.

## C3 Constraints

To ensure the effectiveness and computational reliability of the combination granularity strategy, the following constraints must be satisfied:

### C3.1 Weak Scene Independence Constraint

1) The scene overlap degree $\omega(v_k, v_l)$ between any two video samples $v_k$ and $v_l$ is $\leq 0.5$ (no overlap in core dynamic information);

2) For sample pairs with an overlap degree $\omega > 0.3$, only one sample is retained for covariance calculation (the sample with a larger number of key frames is selected);

3) For sample pairs with an overlap degree $0 < \omega \leq 0.3$, they participate in the calculation with a weight of $w = 1 - \omega$ (the higher the overlap degree, the lower the weight).

### C3.2 Label Consistency Constraint

1) The labels of S, M, and V samples in the same scene are completely consistent, that is:

$$t(s_j) = t(m_k) = t(v_k), (s_j \in S_k \text{ and } m_k \text{ corresponds to } v_k)$$

2) The labels of samples in overlapping scenes are consistent, that is, if $\omega(v_k, v_l) > 0$, then $t(v_k) = t(v_l)$ (to avoid label conflicts).

### C3.3 Derivative Consistency Constraint

1) Frame order consistency: The frame order of M is consistent with the key frame order of V, and the disassembly order of S is consistent with the frame order of M;

2) Frame number matching: The number of frames $K_k$ of the $k$-th multi-image sample is completely equal to the number of samples in the corresponding static image subset $S_k$ ($|S_k| = K_k$);

3) Feature correlation: The features of S and the corresponding M/V satisfy $Corr(f(S_{k,i}), f(M_k)) > 0.8$ (Pearson correlation coefficient, to ensure that the derivative relationship can be quantified).

## C4 Symbol Table

Table C1. Symbol Table

| Symbol | Meaning | Value/Range |
|---|---|---|
| $V, v_k$ | Video modality set, $k$-th video sample | $N_V = 180$ |
| $F_k^V$ | Key transition frame set of the $k$-th video | $|F_k^V| = K_k = 3 \sim 6$ |
| $M, m_k$ | Multi-image modality set, $k$-th multi-image sample | $N_M = 180, m_k = F_k^V$ |
| $S, s_j$ | Static image modality set, $j$-th static image sample | $N_S = 1002$ |
| $S_k$ | Static image subset disassembled from the $k$-th multi-image sample | $|S_k| = K_k$ |
| $t_k$ | Starting index of static images corresponding to the $k$-th multi-image sample | $t_k = \sum_{i=1}^{k-1} K_i$ |
| $\omega(a,b)$ | Scene overlap degree between sample $a$ and $b$ | $[0,1], \omega_{avg} \leq 0.3$ |
| $Feat(v)$ | Scene feature set of video $v$ | Including map coordinates, action labels, etc. |
| $*$ | Intra-sample fusion operation | - |
| $+$ | Inter-sample mixing operation | - |
| $\Sigma_X$ | Weighted covariance matrix of modality $X$ | $d \times d, d = 1024$ |
| $I(X;Y)$ | Mutual information between modality $X$ and $Y$ | In the experiment, $I(S;M) = 1.24$ bits, $I(S;V) = 1.51$ bits |

## C5 Application Examples

Based on the above definitions, the combination granularity construction logic of the core datasets in the experiment is as follows:

(1) annotions_new2.1(S): A static image modality dataset generated by disassembling 180 multi-image samples, where each static image sample corresponds to $s_{t_k+i} = f_{k,i}^M$;

(2) MI2.8.3(M): A multi-image modality dataset, where each sample $m_k = F_k^V$ and corresponds to video $v_k$ in a 1:1 ratio;

(3) my_video_data(V): A video modality dataset, where each sample $v_k$ contains $T_k$ frames, and the key frame set $F_k^V$ is used to generate $m_k$;

(4) combo_D1MVS(M*V+S): First perform intra-sample fusion of $m_k$ and $v_k$ ($m_k * v_k = g(a(f(m_k)) * a(f(v_k)))$), then perform inter-sample mixing with S, which is the optimal strategy in the experiment.

# Appendix D Detailed Determination of Optimal Fine-tuning Epochs for Various Datasets

Due to the excessive amount of statistical data, only the line graphs generated from extracting data from trainer_log.jsonl are presented here. According to the loss trends of each dataset on the training and validation sets, the optimal fine-tuning epochs are determined.

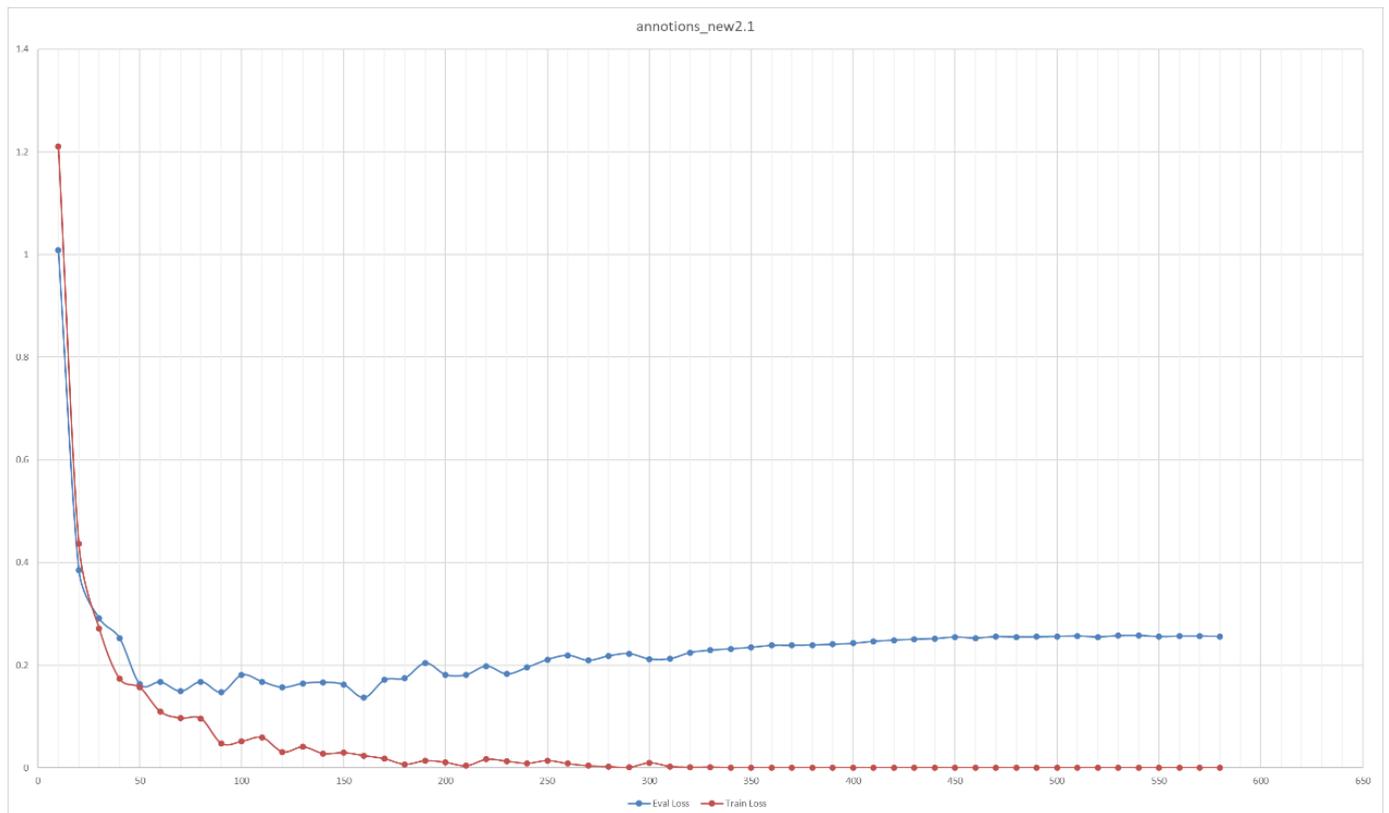

annotions_new2.1

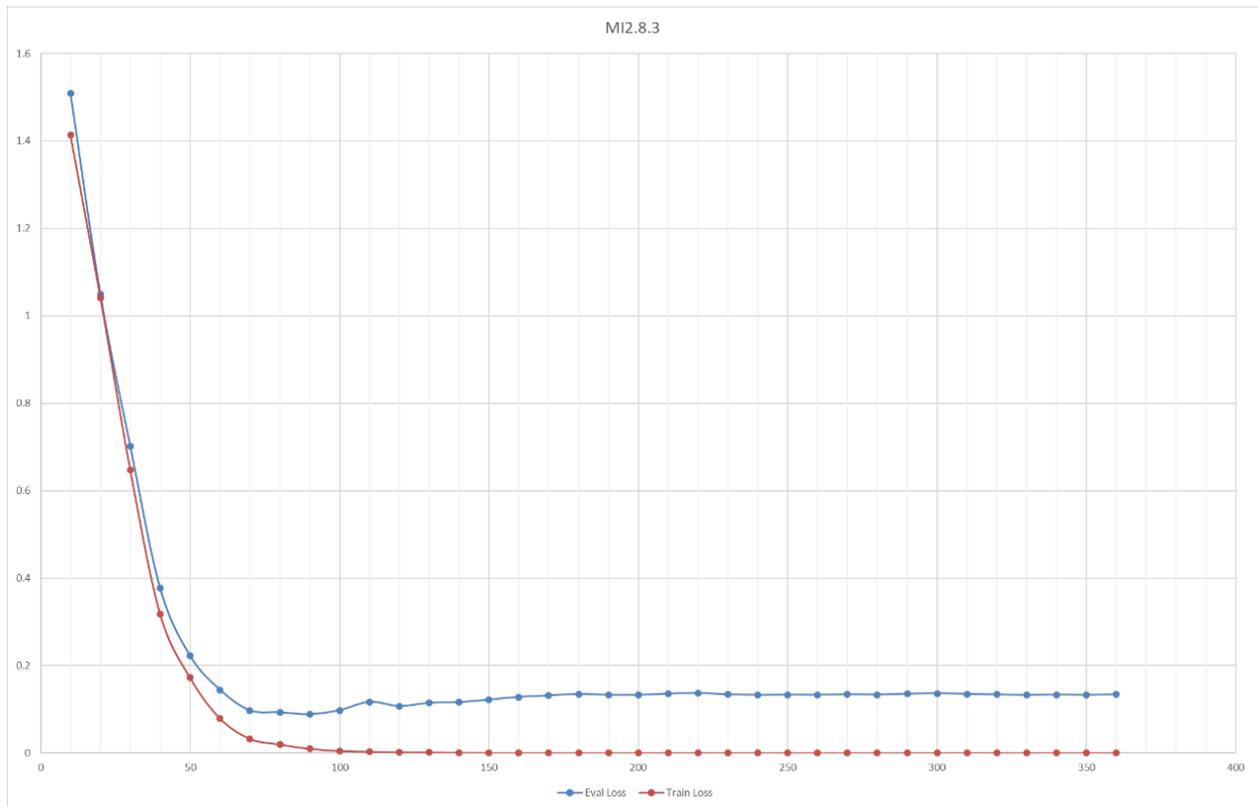
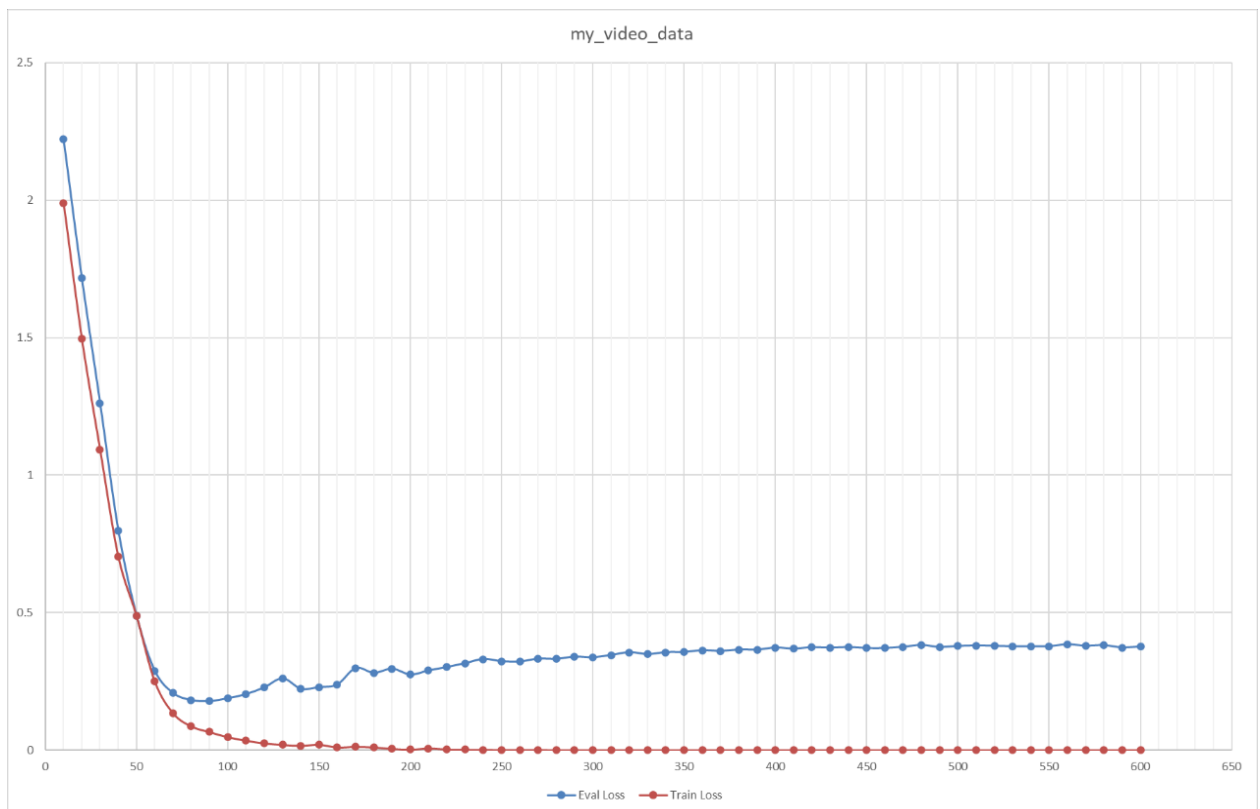

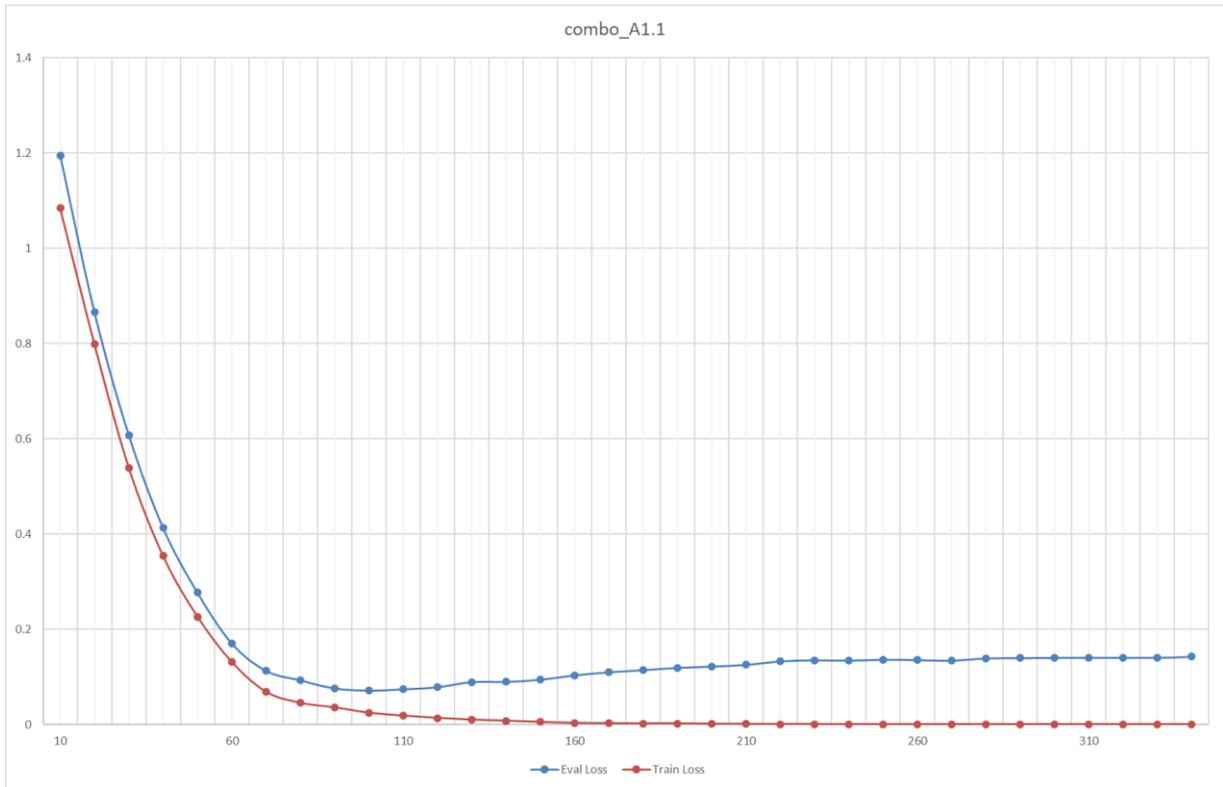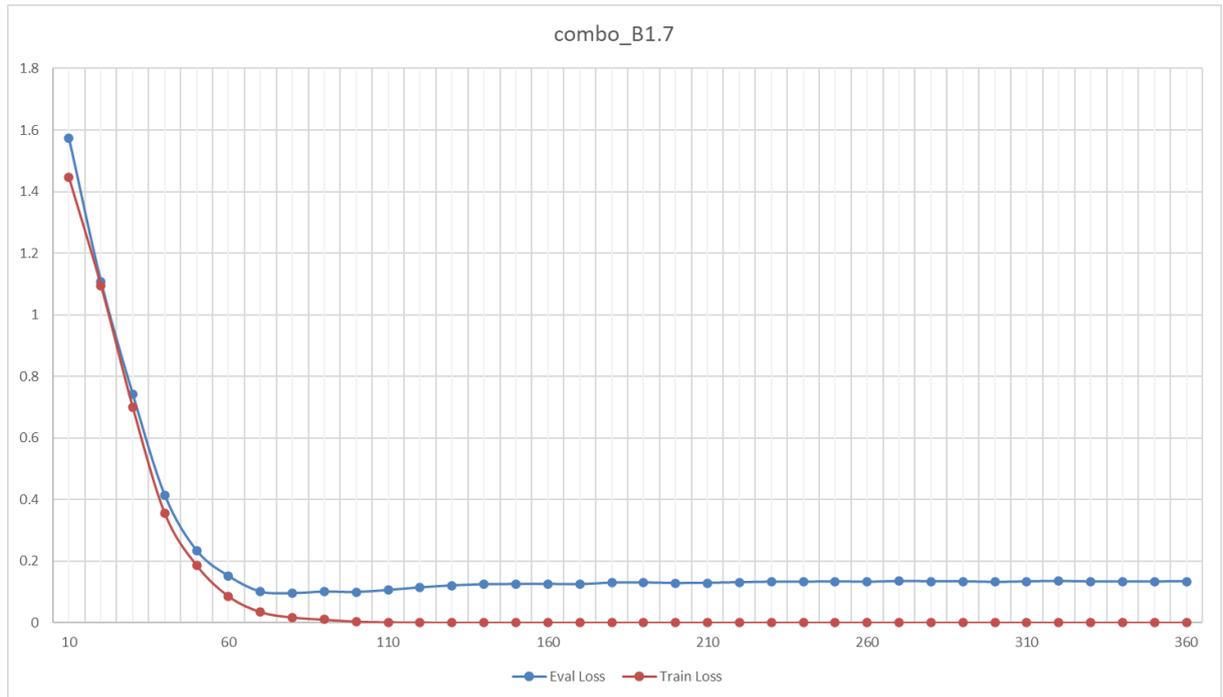

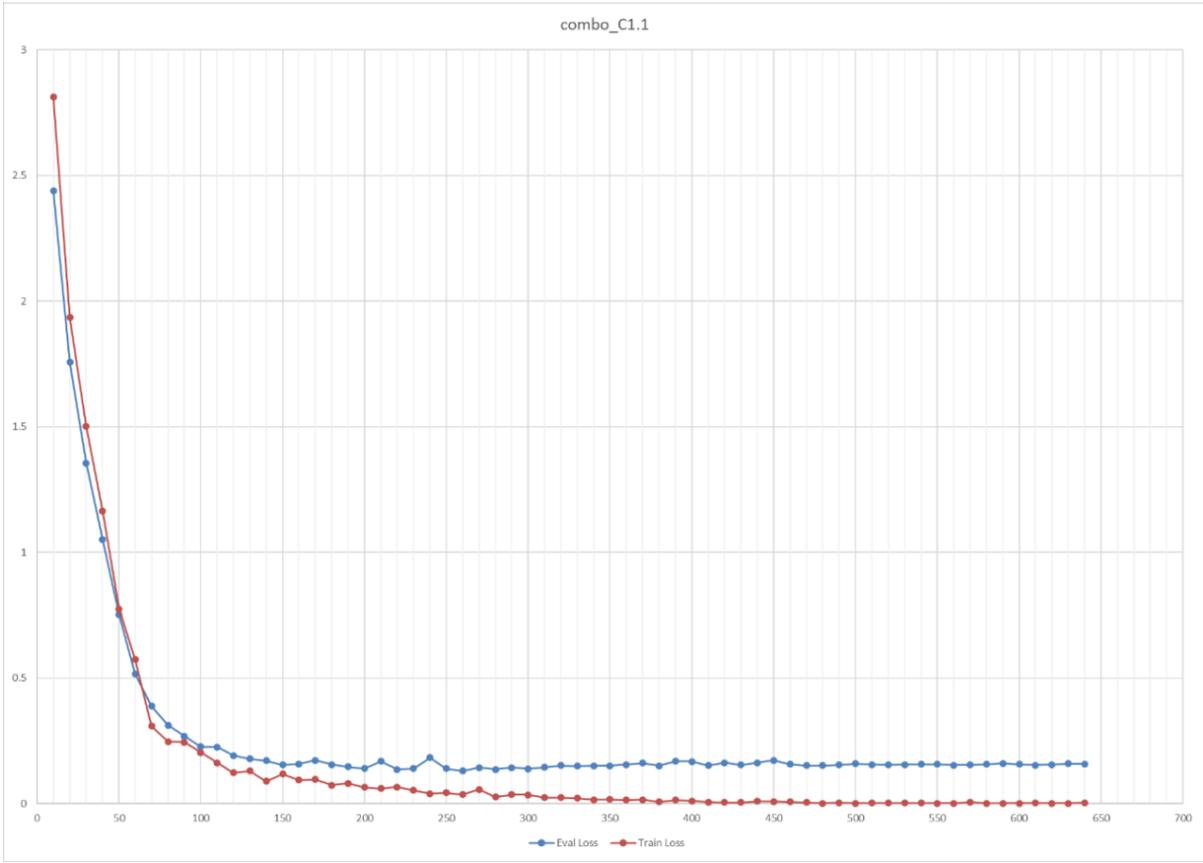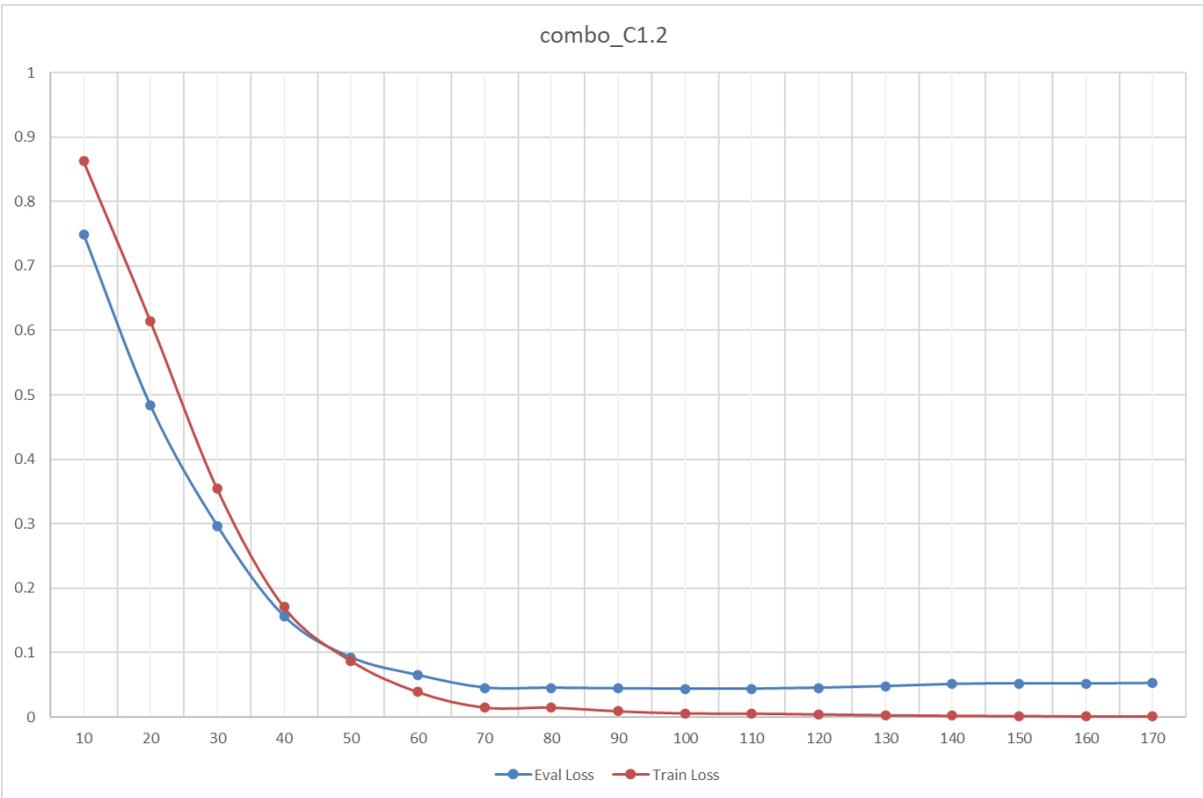19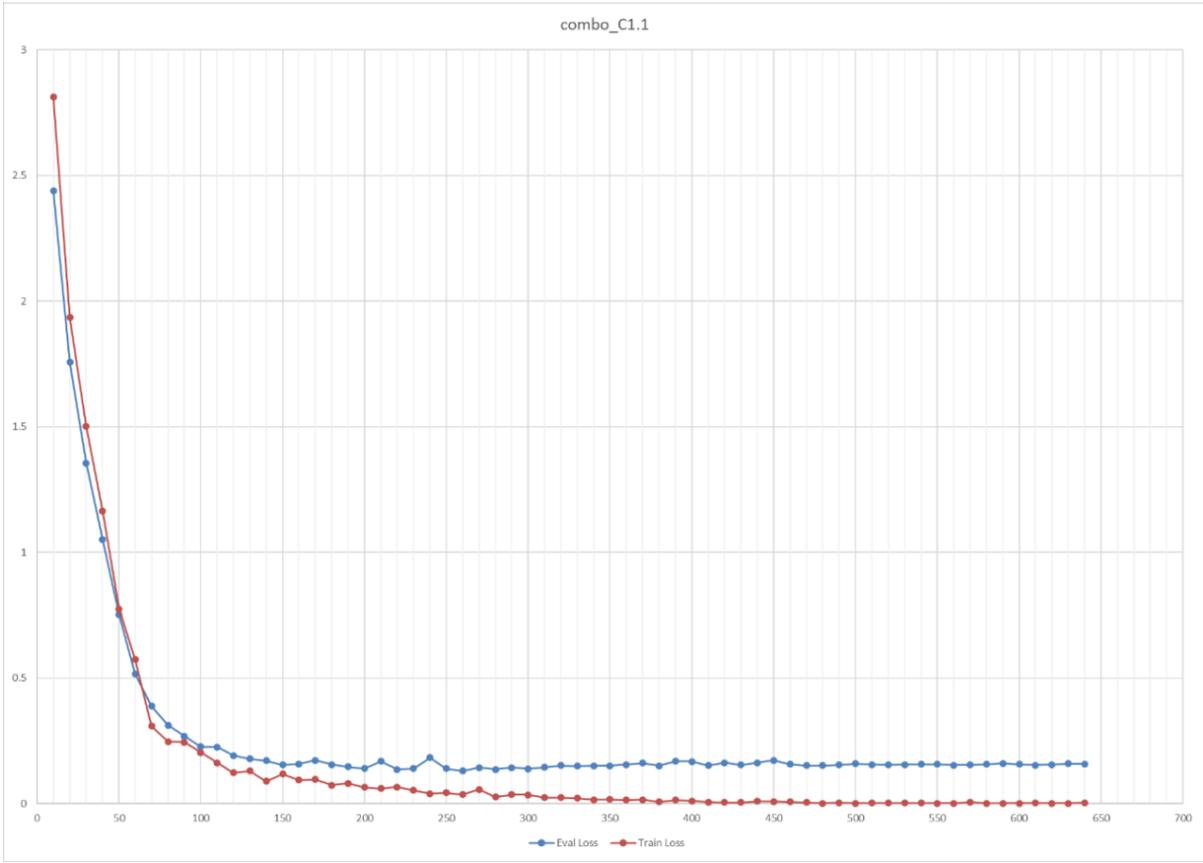
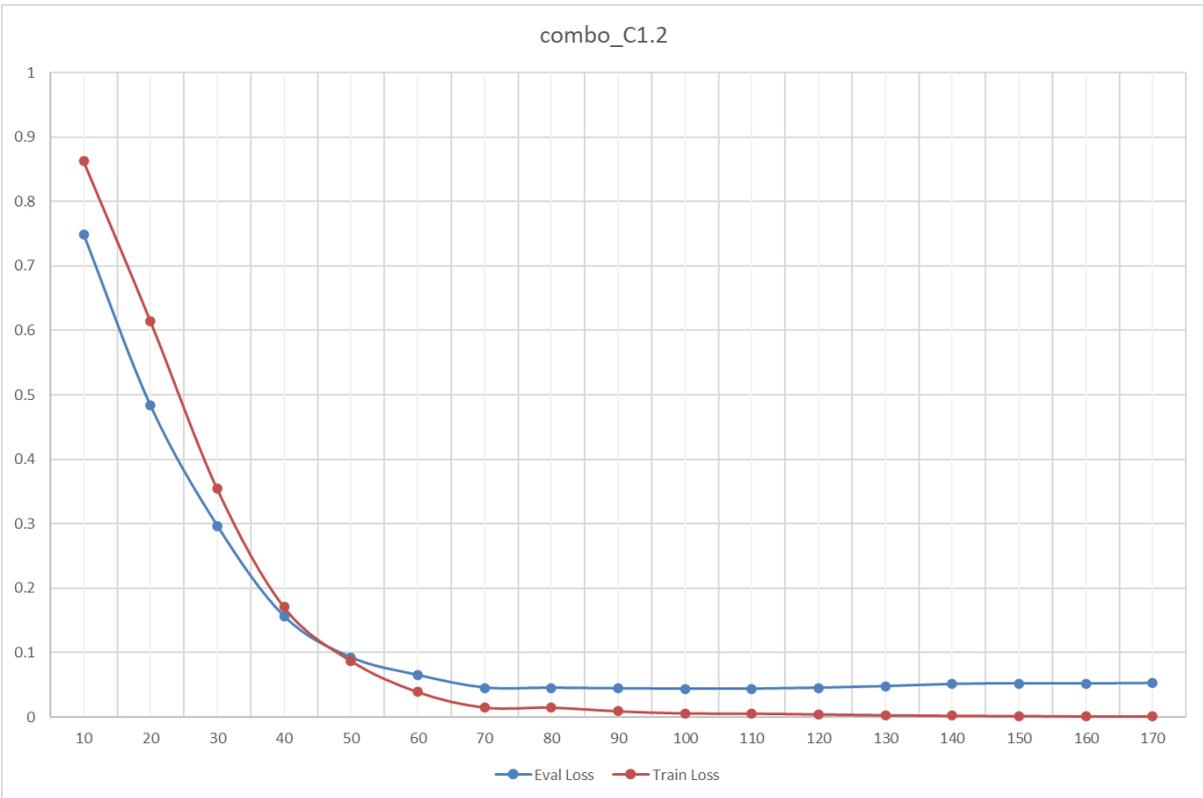

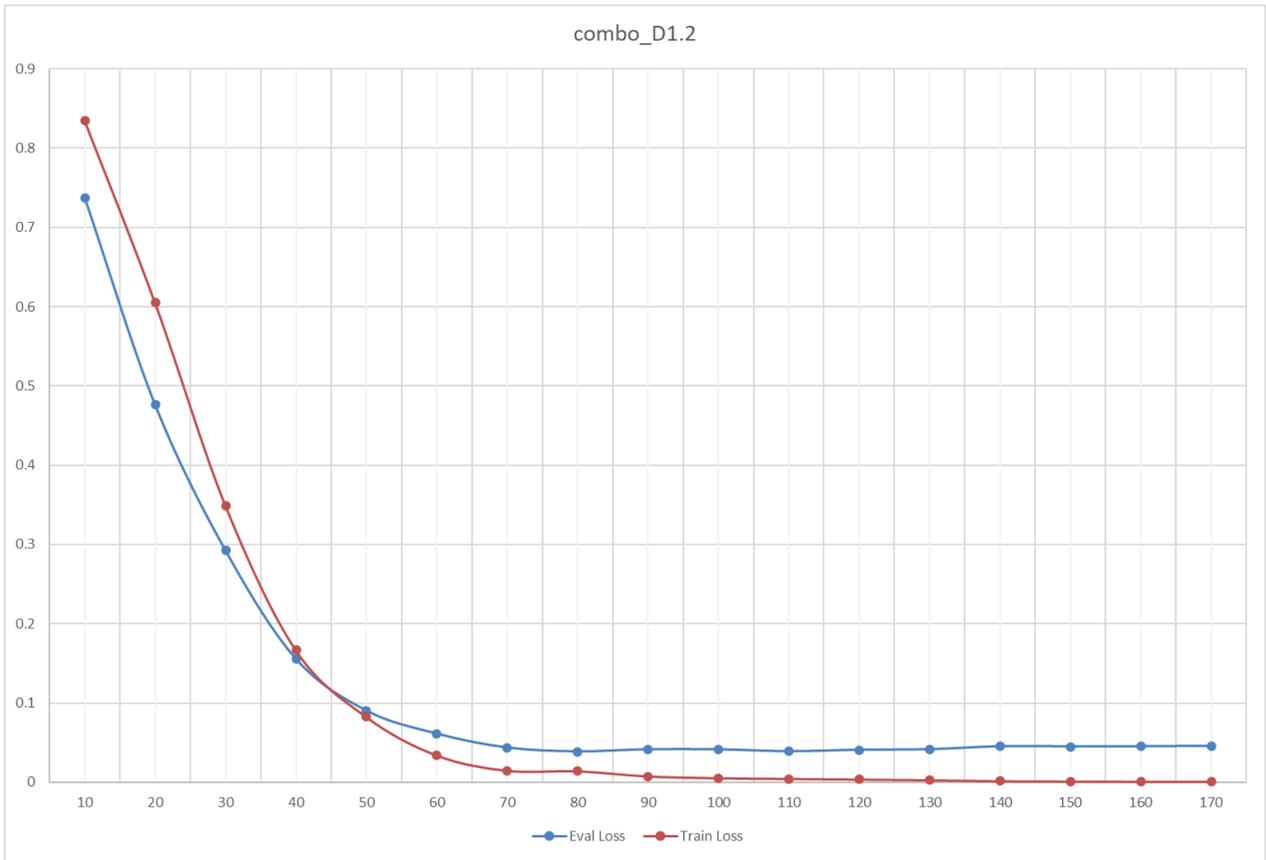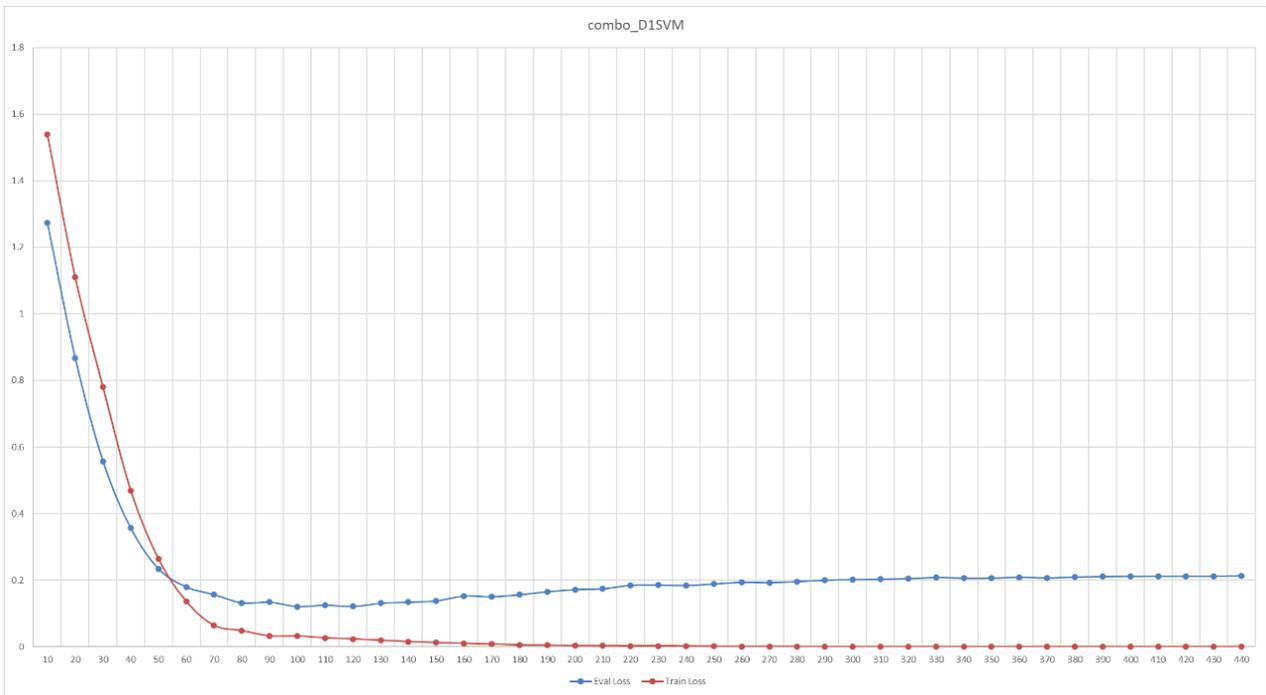

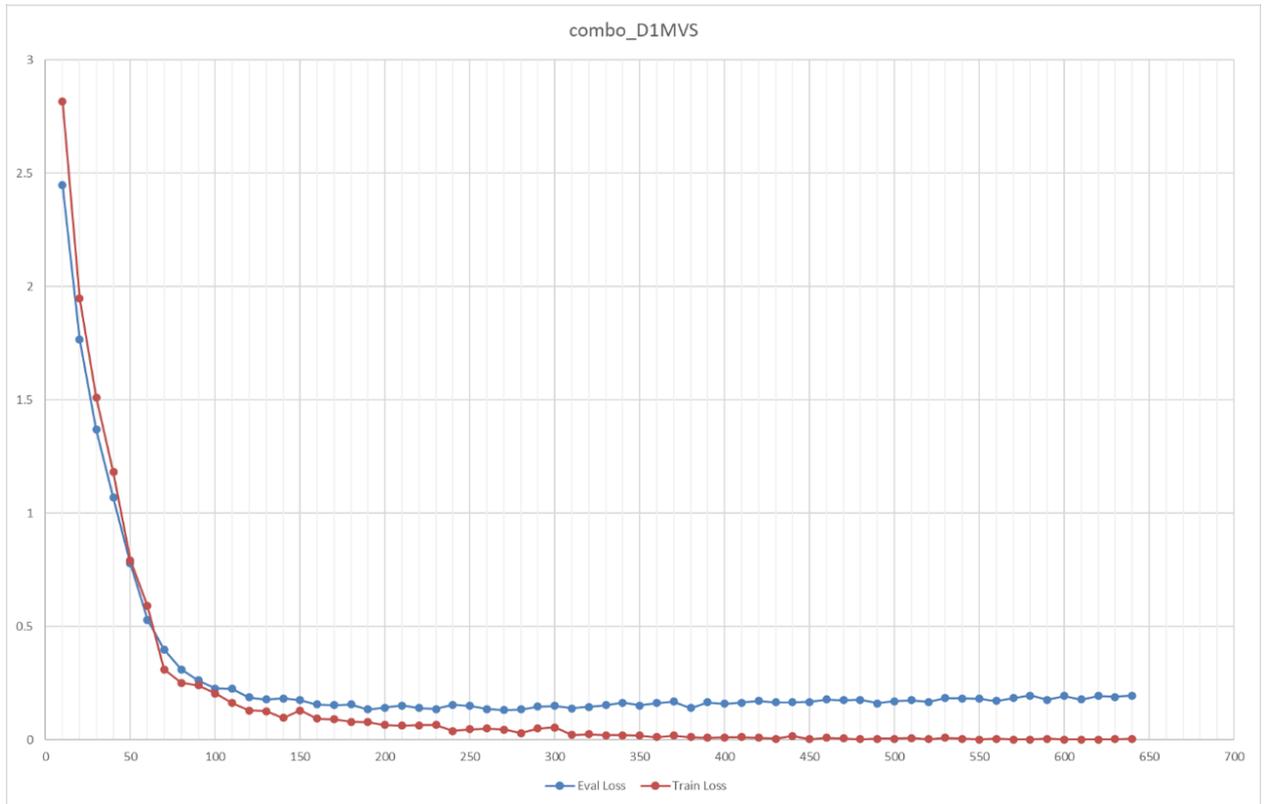
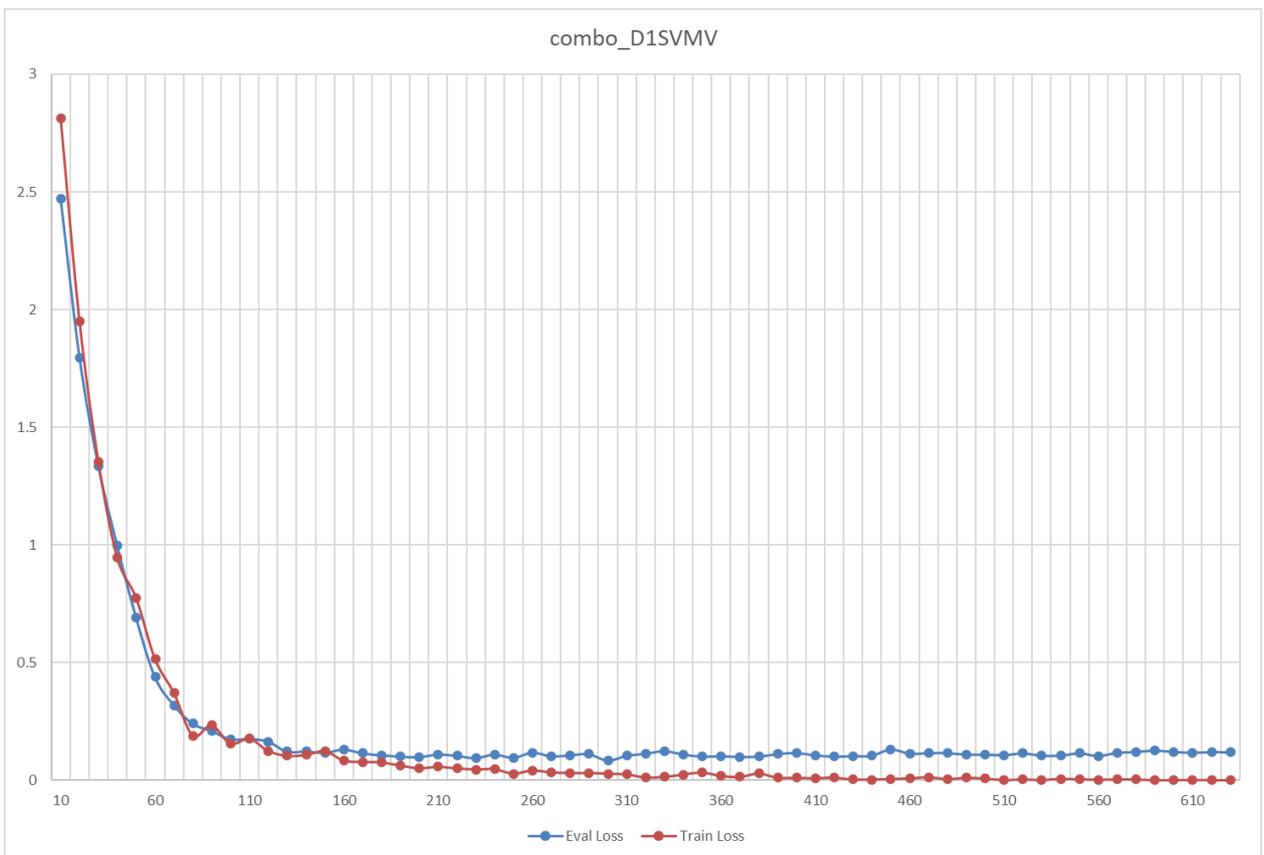

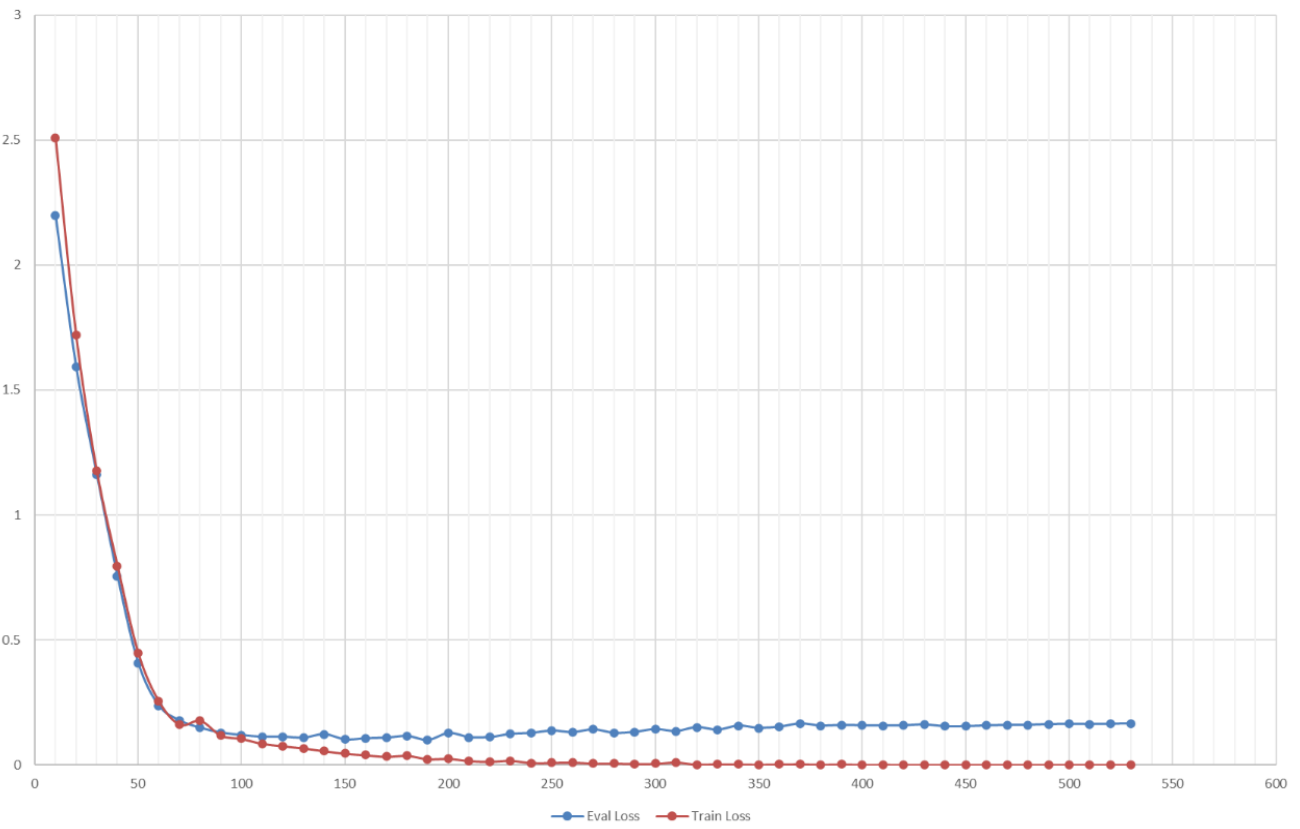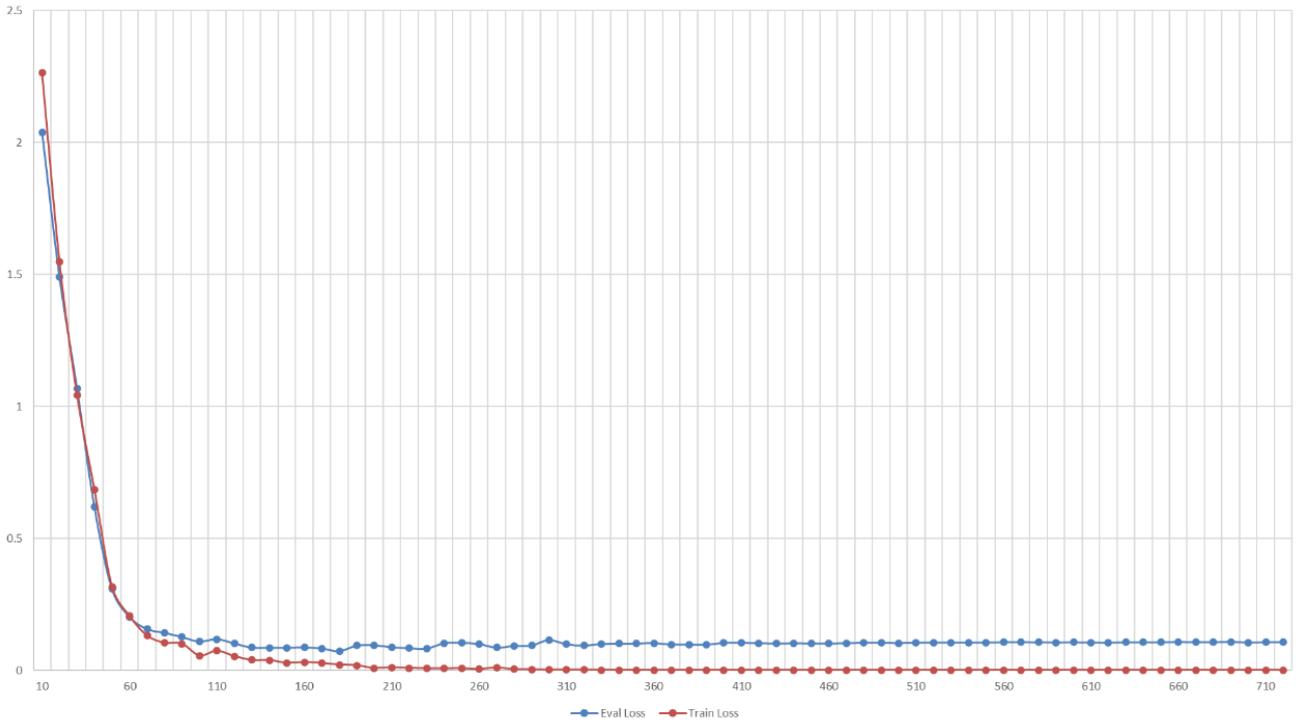

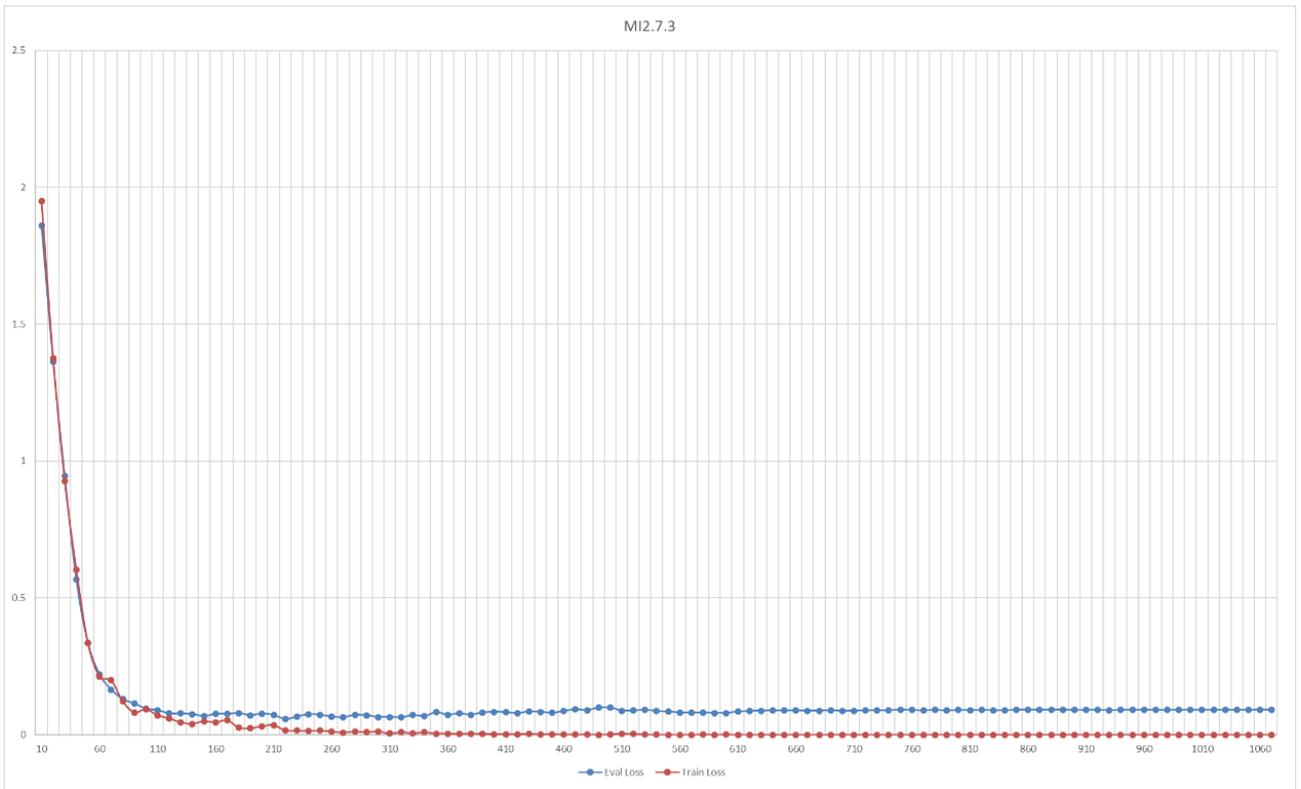
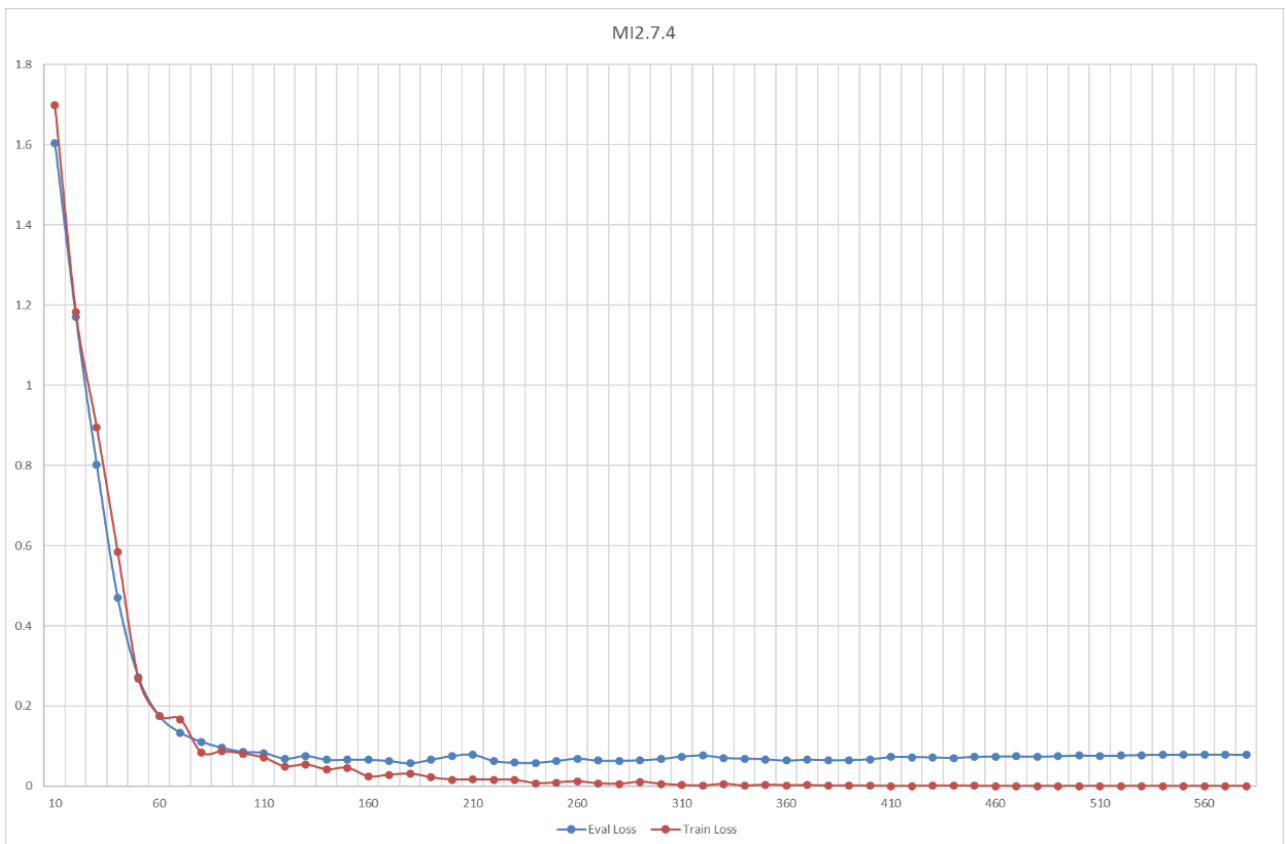

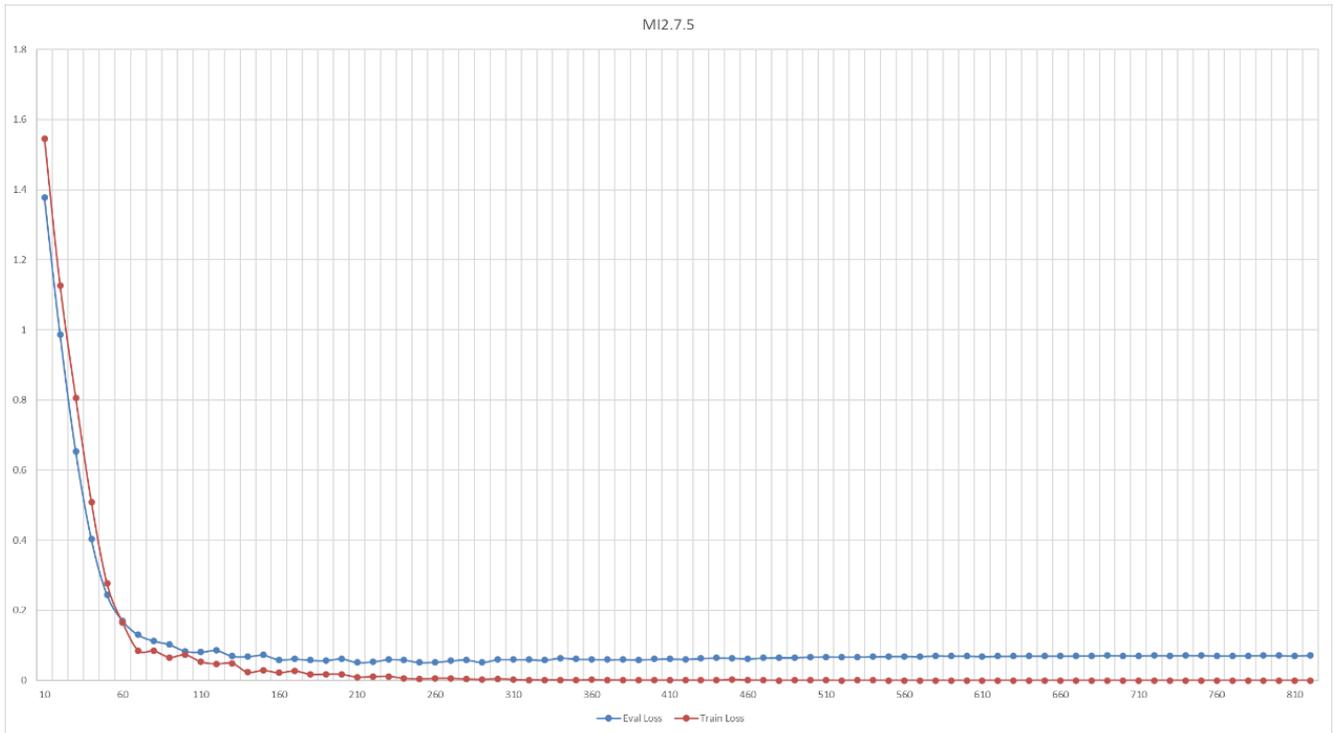

**Fig. D1**. Trends of Train Loss and Eval Loss vs. Steps for Various Datasets. Determine optimal fine-tuning steps and epochs based on the trend of Eval Loss vs. Train Loss in each graph.

# Appendix E Experimental Records

The prototype system validation phase refers to testing after fine-tuning with one modality dataset (annotions_new) for 30 rounds (without determining the optimal number of rounds) and before designing other datasets, connecting all components including the image understanding unit, execution unit, environment, etc. The system-level validation phase experiment refers to the overall experiment after comparing and screening through modal and *combination granularity* ablation experiments, selecting the combo_D1MVS dataset, which showed high prediction indicators, for fine-tuning for 27 rounds to obtain the large model as the image understanding layer.

### E1 Prototype System Validation Program Execution Log Snippet

Presented below is the English translation of the native program log output generated by the operation within the GamePlatform B environment.

D:\ProgramData\Anaconda\envs\python3.10\python.exe F:\VLM\UI-TARS-main\codes\test_Qwen2.5vl_url\Qwen_UITARS2.py

- 🚀 Welcome to the Qwen2.5-VL + UI-TARS Automated System
- 📌 Please enter the task goal; the program will automatically analyze the interface and execute operations
- 📌 Enter 'q' or 'quit' to exit the program
- 📌 Please enter the task instruction: Analyze the image and provide the next mouse action
- 🔄 Automatically looping the task: Analyze the image and provide the next mouse action
- 🔄 Executing Step 1
- 📷 Taking screenshot...
- 🧠 Calling Qwen2.5-VL to get task instruction...
- 🧠 Adjusting image size...
- 🖼 Adjusted image size: (1366, 768)

- 🧠 Converting image to Base64...
- 🧠 Calling Qwen2.5-VL model...
- HTTP Status Code: 200
- 💬 Qwen2.5-VL Output Instruction: Click [Mark as Direct Fire]
- 🧠 Calling UI-TARS model...
- 🤖 UI-TARS Model Reply: From the current page, I need to click the "Direct Fire" button located mid-left, which is a blue rectangle with white text "Direct Fire", to mark it as Direct Fire.
- Action: click(start_box='(257,671)')
- 👉 Clicking coordinates: 493, 725
- 🕐 Waiting for interface update...
- 🔄 Executing Step 2
- 📷 Taking screenshot...
- 💬 Calling Qwen2.5-VL to get task instruction...
- 🧠 Adjusting image size...
- 🖼 Adjusted image size: (1366, 768)
- 🧠 Converting image to Base64...
- 🧠 Calling Qwen2.5-VL model...
- HTTP Status Code: 200
- 💬 Qwen2.5-VL Output Instruction: Click "Large Caliber Direct Fire Cannon"
- 🧠 Calling UI-TARS model...
- 🤖 UI-TARS Model Reply: From the current page, I need to click the "Large Caliber Direct Fire Cannon" button to select this tool. From the interface, this button is located at the bottom of the tool information window, displayed as a blue button. I need to click the "Large Caliber Direct Fire Cannon" button to select this tool, which will switch the tool to "Large Caliber Direct Fire Cannon" mode.
- Action: click(start_box='(257,666)')
- 👉 Clicking coordinates: 493, 719
- 🕐 Waiting for interface update...
- 🔄 Executing Step 3
- 📷 Taking screenshot...
- 💬 Calling Qwen2.5-VL to get task instruction...
- 🧠 Adjusting image size...
- 🖼 Adjusted image size: (1366, 768)
- 🧠 Converting image to Base64...
- 🧠 Calling Qwen2.5-VL model...
- HTTP Status Code: 200
- 💬 Qwen2.5-VL Output Instruction: Click Operator Information Box
- 🧠 Calling UI-TARS model...
- 🤖 UI-TARS Model Reply: From the current page, I need to click the Operator Information Box to complete the task.
- Action: click(start_box='(231,402)')
- 👉 Clicking coordinates: 444, 434
- 🕐 Waiting for interface update...
- 🔄 Executing Step 4
- 📷 Taking screenshot...
- 💬 Calling Qwen2.5-VL to get task instruction...
- 🧠 Adjusting image size...

· 🖼 Adjusted image size: (1366, 768)
· 🧠 Converting image to Base64...
· 🧠 Calling Qwen2.5-VL model...
· HTTP Status Code: 200
· 🧠 Qwen2.5-VL Output Instruction: Click Operator Information Box
· 🧠 Calling UI-TARS model...
· 🤖 UI-TARS Model Reply: From the current page, I need to click the Operator Information Box. From the interface, the Operator Information Box is located on the right side of the screen, displaying relevant information about the current match. I need to click this information box to view detailed content.
· Action: click(start_box='(893,190)')
· 👉 Clicking coordinates: 1715, 205
· 🕐 Waiting for interface update...
· 🔄 Executing Step 5
· 📷 Taking screenshot...
· 🧠 Calling Qwen2.5-VL to get task instruction...
· 🧠 Adjusting image size...
· 🖼 Adjusted image size: (1366, 768)
· 🧠 Converting image to Base64...
· 🧠 Calling Qwen2.5-VL model...
· HTTP Status Code: 200
· 🧠 Qwen2.5-VL Output Instruction: Click [Fire]
· 🧠 Calling UI-TARS model...
· 🤖 UI-TARS Model Reply: From the current page, I am in a war strategy game interface. According to the task requirements, I need to click [Fire]. From the interface, I need to click a position on the map to perform the Fire operation. I should click a suitable Fire position on the map to complete the Fire operation.

It can be seen that during operation on GamePlatform B, the instruction action "Click [Fire]" originating from GamePlatform C appeared.

# E2 Prototype System Validation Record

This experiment was conducted after implementing the loop operation code Qwen_UITARS2.py to verify the 3-platform 1. screenshot; 2. send image to Qwen2.5vl; 3. return mouse instruction; 4. send mouse instruction and image to UI-TARS; 5. return specific action coordinates; 6. UI-TARS operates mouse, etc., loop process.

Table E1. Detailed Prototype System Validation Record

| Exp. Platform, Scenario | Number of Exp. | Experimental Phenomena | Summary |
|---|---|---|---|
| GamePlatform A, Air Unit aircraft scenario 1v1 | 10 | 1. Stagnation was encountered after the execution of the "click manual allocation" operation. Subsequently, the screenshot captured for Qwen2.5-VL analysis failed to be recognized, resulting in the inability to generate the next action of "click the cyan-colored [Aircraft]". (However, at the initial stage, the action "click the cyan-colored [Aircraft]" could be performed even when the affiliation could not be distinguished; what is the reason for | 1. The page changes were not obvious. 2. The affiliation could not be distinguished. |

| | | | | | | |
|---|---|---|---|---|---|---|
| | | | this?)<br>2.When positioned at the Team A station, at the beginning of the scenario, the cyan-colored [Aircraft] was directly clicked. | | | |
| GamePlatform B, Ground Combat 2v2 | 3 | 1、This experiment involved 6 phases<br><br>| 1 | 2 | 3 | 4 | 5 | 6 |<br>|---|---|---|---|---|---|<br>| Normal | Normal | Normal | Normal | Normal | Normal |<br><br>All phases could be executed sequentially.<br>2.Although "clicking of the Operator Information Panel" appeared in the fine-tuning dataset, it should not have been present in this specific workflow, constituting a redundant action.；<br>3.Qwen2.5-VL returned the mouse command "click [Fire]" designed for the GamePlatform C platform.<br>4. After Qwen2.5-VL returned "click the blue button of [Direct Fire]", actions of "click [Toggle Concealment]" and "click [Maneuver]" were performed by UI-TARS. Furthermore, after clicking the [Maneuver] button, an action of clicking [Cancel] was subsequently performed.<br>5. While positioned at the Team A station, clicking on the Team B operator icon was attempted to conduct direct fire against the Team A.<br><br>| Summary of the Above Phenomena | Dataset | Sg. A Dt | Sg. B Dt | Sg. C Dt | Sg. A If | Sg. B If | Sg. C If |<br>|---|---|---|---|---|---|---|---|<br>| 1. Click Operator Info Panel | Y | N | Y | N | N | Y | Y |<br>| 2. Click [Fire] | Y | N | N | Y | N | N | Y |<br>| 3. Click [Toggle Concealment] | N | N | N | N | N | Y | N |<br>| 4. Click [Maneuver] | Y | N | N | Y | N | Y | Y |<br>| 5. Click [Cancel] | N | N | N | N | N | Y | N |<br><br>This table is used for statistical purposes regarding actions performed by the large model that were not specified in the dataset but are present in the interface.(In the above table Sg is Strategy Game,Dt is dateset，If is Interface.) | 1.The entire workflow could be successfully executed.<br>2.Redundant actions were observed, specifically the "clicking of the Operator Information Panel" action during combat.<br>3.The command "click [Fire]" output by Qwen2.5-VL is an instruction designed for the GamePlatform C platform. As the current scenarios in GamePlatform B and GamePlatform C are similar (both involve 2v2 ground units), Qwen2.5 did not effectively distinguish between the platforms. However, the fact that Qwen2.5-VL mistakenly identified the GamePlatform B scenario as a GamePlatform C scenario (where the four units in GamePlatform B are positioned Red NW, Blue SE, and in GamePlatform C Red W, Blue E) suggests a potential generalization capability for GamePlatform C scenarios with varied formations.<br>4.The actions of clicking [Toggle Concealment] and [Maneuver] occurred because the tool was in a cooldown phase after a shot was fired; the [Direct Fire] button was not present in the operator information interface. Consequently, UI-TARS selected and clicked a blue button within that interface. Because the dataset consists of images where the mouse is at "a certain location" mapped to JSON instructions to click "that location", clicking [Maneuver] while the mouse happened to be over the [Cancel] button resulted in a returned command to click [Cancel] (Does this indicate autonomy in the large model?). Verification is needed regarding whether Qwen2.5 output "click [Cancel]" at that time. All subsequent Python outputs must be backed up, and all automated platform operations must be screen-recorded. |

| | | | 5.Inability to distinguish between friendly and other sides was observed. Descriptions specifying affiliation (friendly/enemy) need to be added to the JSON data. |

# E3 System-Level Validation Record

The current experiment mainly records the continuity of action execution, yielding the table below:

Table E2.Detailed System-Level Validation Record

| Exp. Platform, Scenario | Number of Exp. | Experimental Phenomena | Summary |
|---|---|---|---|
| GamePlatform A, Air Unit Aircraft Scenario 1v1 | 60 | 1. Compared to the initial stage, through modifications to the dataset, stagnation was no longer encountered following the "click manual allocation" operation, but rather occurred after "clicking the cyan-colored aircraft" without resource being allocated (click x resource, double-click x resource), leading to the closure of the [Resource Allocation] interface.<br>2.Compared to the initial stage, the ability to distinguish between Red and Team Bs has been achieved, and the probability of incorrectly operating the B force station when initially positioned at the A force station at the start of the scenario has been reduced.<br>3.Experiments demonstrated that the accuracy rate of "clicking the [cyan-colored aircraft] icon" was lower when the cyan-colored aircraft's name was not displayed, compared to when it was displayed. Furthermore, when the cyan-colored aircraft's name (e.g., AIR UNIT A) was displayed and the instruction executed was "click the [cyan-colored aircraft] icon AIR UNIT A", there was a probability that the click would target the name "AIR UNIT A" in the upper right corner of the aircraft icon rather than the aircraft icon itself. | 1. Modifications to the dataset had a significant impact on the experiment's execution, particularly after altering certain types of images within the training set. A noteworthy point is the correspondence between screenshots and action descriptions. The screenshot corresponding to an "action description" should be the screenshot captured "immediately before this action is to be performed", but it requires a substantial difference between the pre-action and post-action images. Alternatively, the selected base model must possess sufficient capability to recognize more subtle image changes.<br>2.Compared to the prototype system stage, explicitly stating "you are currently Team A/Team B" in the dataset later on effectively reduced the issue of operating the wrong station.<br>3.Textual description significantly impacts the accuracy of action execution. Adding textual descriptions on top of color (cyan, red icon) and shape (aircraft, Ground Unit) descriptions can improve clicking accuracy. This is particularly evident when the image understanding component mistakes an "aircraft" icon for a "Ground Unit" icon. |

| GamePlatform A, Ground Unit Scenario 1v1 | 40 | Automated cyclic interaction was achievable, but with instances of stagnation, occasionally requiring human assistance. The actions performed by the Ground Units here, like the aircraft above, involved resource allocation. After clicking "manual selection", the cursor changed to a "+" symbol, which was relatively small. The image understanding component continued to perceive this as the initial state of selecting the attacking force, leading to the selection of "click the red [aircraft icon]" when the intended target was the Team B Ground Unit. | The entire workflow is based on image recognition. Lack of change or minimal change in the image will prevent the image understanding component from outputting accurate actions for UI-TARS to execute. This requires significant attention. |
|---|---|---|---|
| GamePlatform B, Ground Combat 2v2 | 30 | 1.This experiment involved 6 phases:<br><br>| 1 | 2 | 3 | 4 | 5 | 6 |<br>|---|---|---|---|---|---|<br>| Normal | Normal | Normal | Normal | Normal | Normal |<br><br>Sequential execution was possible.<br>2. Although "clicking the Operator Information Panel" appeared in the fine-tuning dataset, it should not have been present in this specific workflow, constituting a redundant action.<br>3.Qwen2.5-VL, on the GamePlatform B platform, returned the erroneous mouse command "click [Manual Fire]". The term "Manual" actually originated from the GamePlatform A platform, although UI-TARS executed the action of "clicking the [Direct Fire] button".<br>4.The probability of clicking the [Maneuver] button was reduced. | 1. The entire workflow could be successfully executed.<br>2.Redundant actions were observed, specifically the "clicking of the Operator Information Panel" action during combat.<br>3.There was a low probability of platform confusion.<br>4.Compared to the prototype system, the interface waiting time was currently increased to 8 seconds, preventing the execution of actions under conditions where the interface changed abruptly after the screenshot was uploaded. |
| GamePlatform C, Ground Combat 2v2 | 50 | Automation was achievable, but stagnation occurred during cycles. | Similar to the GamePlatform A platform, this was caused by minimal interface changes before and after actions, and insufficiently detailed recognition of subtle changes by the large model. |

## Appendix F Experiment to Improve Execution Mechanism Precision

First, let's introduce action response, as it is closely related to subsequent dataset design. Here, action response refers to the process of giving UI-TARS an instruction and having UI-TARS execute that instruction on the strategy game platform interface. UI-TARS is an open-source multimodal intelligent agent developed by ByteDance based on a visual-language architecture. Its 1.5 version achieved leading results in 7 benchmark tests including OSWorld and demonstrated breakthrough long-term reasoning capabilities in game scenarios. The system provides cross-platform application support, covering automated office work, software development testing, and other application scenarios. Here, we use the API deployed on Alibaba Cloud for UI-TARS-7B to implement operations. UI-TARS's work is mainly divided into two phases: first, the desktop version of UI-TARS tests various action instruction implementations, and second, calling the UI-TARS API(As UI-TARS is based on Apache License 2.0, all modifications in this study, including coordinate mapping optimizations, have retained the original copyright notice).

## F1 Testing Action Phase

In the initial experimental phase, various action instruction descriptions (e.g., "click [Red Ground Unit] icon" or "click red [Ground Unit] icon") were tested on the desktop version of UI-TARS to determine which description could enable more precise action execution, especially when hexagonal grid numbers are obscured by icons (GamePlatform C platform) or input boxes and buttons are blurry (GamePlatform A platform). This process is recorded to lay the foundation for dataset creation. The experimental process record is too long; details refer to Appendix G.

## F2 API Calling Phase

Compared to the desktop version of UI-TARS, calling the API has three advantages.

Compared to the desktop version of UI-TARS, calling the API offers 3 advantages:

(1) As the desktop version's mouse must move between the Strategy Game interface and the UI-TARS input box, partially obscuring the Strategy Game interface, the UI-TARS interface deployed on Alibaba Cloud was called instead.

(2) This approach also facilitates the batch transmission of action commands output by the fine-tuned large model to UI-TARS, rather than manually placing commands into the desktop version's input box.

(3) It allows for precision improvement of actions that cannot be accurately completed.

Due to errors in pixel coordinates between pre-screenshot and operation times caused by UI-TARS's process of scaling screenshots against the original image and then converting normalized coordinates back to true screen pixels, improvements were made to UI-TARS's processes of "converting normalized coordinates to true screen pixels" and "rounding to obtain the true screen coordinates (sx, sy)". Precision was improved by approximately 3-4 pixels (using 2360×1600 resolution as an example) by experimenting with modifications to the floating-point coordinate system.

A comparison is shown in Table 1:

**Table F1**. Difference in Integer vs. Floating-Point Pixel Variation

| Input | X Absolute Value | Y Absolute Value |
|---|---|---|
| Integer (500,500) | 1180px | 800px |
| Floating Point (498.25,512.75) | 1176.87→1177px | 820.40→820 px |

The current mapping method is linear:

$$x\_absolute = (screen\_width * relative\_x) / 1000$$
$$y\_absolute = (screen\_height * relative\_y) / 1000$$

Assuming a small error Δx exists in the model's output relative_x along the width, the final click position will exhibit this error::

$$\Delta x_{absolute} = \left| \frac{dx_{relative}}{dx_{absolute}} \right| \cdot \Delta x_{relative} = \frac{screen\_width}{1000} \Delta x$$

Therefore, larger screens result in larger errors.

A polynomial was introduced to approximate the true coordinate relationship:

$$x_{absolute} = a_0 + a_1 \cdot x_{relative} + a_2 x^2_{relative}$$

Optimal coefficients a0, a1, a2 were trained using multiple samples (x_relative, x_absolute).

The incorporation of sliding averages, derivative-based trend judgment, and edge detection reduced maximum error by 3–5 pixels. After this precision gain, the problem—failure to select input boxes when they are blurred (GamePlatform A platform)—was resolved.

## F3 Comparison Experiment Before and After Improving UI-TARS

This video contrasts the desktop UI-TARS application before and after the proposed refinement, exemplifying the achieved precision improvement.

In [Video Link 1](), UI-TARS repeatedly fails to accurately click the small up-triangle to the right of the input box used to increment resource allocation.

In [Video Link 2](), the enhanced method introduced in this study is applied, resulting in a measurable precision gain for UI-TARS.

# Appendix G Experiment to Improve Execution Mechanism Precision

To conduct this research, we first needed to confirm that UI-TARS could indeed execute a certain strategy game paradigm language. Therefore, we conducted instruction experiments on the three strategy game platforms with UI-TARS.

## G1 GamePlatform B Platform Instruction Experiment

Table G1. GamePlatform B Platform Command Experiment

| Exp.ID. | Experimental Objective | Mouse Language | Capability | Defect | Number of Exp. | Summary |
|---|---|---|---|---|---|---|
| 0 | Attempted implementation of movement for friendly targets | Click the pink unit, click 1315, click 1316, click 1317 | Capability to move units was demonstrated | Intermittent reliability, frequent failure to click the specific hexagonal grid where the number was located (In fact, UI-TARS likely does not recognize hexagonal grids) | | This area is highly worthy of investigation |
| 1 | Implementation of sustained attacks against a specific target | Click the pink unit in the center of the screen - select attacking force. Click the operation button "Direct Fire" button - prepare to attack. Select the button "Large Direct Fire Cannon" and click - select resource. Mouse-select the Team B unit at the top of the screen as the target - select target. Repeat the steps: prepare to attack, select tool, select target. | Sustained attacks against a single Team B target were achieved | Inability to attack remaining Team B targets after one was destroyed. ① Even against visible targets, effective engagement remained problematic. Without explicit designation as a viable target (i.e., a target marked with a red number), performance would be worse. ② Specifically, the "mouse-select Team B unit as target" step was problematic. | 5 | Positional descriptors ("center of screen", "top of screen", "bottom") are crucial and recognizable by the UR. Distinction must be made between "select and click" versus "click". Performance was |
| | | | | | | intermittent. |
| 2 | Implementation of precise commands | Click the "pink Ground Unit icon in the upper left corner" -- (select attacking force). Click the "blue button labeled [Direct Fire]" -- (prepare to attack). Click the "white button below labeled [Large Direct Fire Cannon]" (select equipment). Click the "blue Ground Unit icon marked with a red number" -- (select target). Repeat the process: prepare to attack, select equipment, select target. | Precision attacks were achieved. | Repeated attacks were not implemented. | 12 | Relatively stable. |
| 3 | Attempted precise and repeated attacks | Click the "pink Ground Unit icon in the upper left corner" -- select attacking force. Click the "blue button labeled [Direct Fire]" -- prepare to attack. Click the "white button below labeled [Large Direct Fire Cannon]" -- select equipment. Click the "blue Ground Unit icon marked with a red number" -- select target. Repeat the process: prepare to attack, select equipment, select target. | Repeated attacks were achievable. | After successfully striking one target, the system could not switch to engage another Team B target. | 12 | Stable. |

The successful execution of entire tasks using sequential commands on the GamePlatform B platform, achieved through descriptive modifications, led to the subsequent adoption of multiple sequential commands for the GamePlatform C platform.

## G2 GamePlatform C Platform Instruction Experiment

Difficulties were encountered during the execution of sequential commands, as detailed in the table below.

Table G2. GamePlatform C Platform Command Experiment Table

| Exp. ID. | Experimental Objective | Mouse Language | Capability | Defect | Number of Exp. | Summary |
|---|---|---|---|---|---|---|
| 0 | Attempted implementation of movement for friendly targets | Click the 0000 Heavy Ground Unit; click the [Maneuver] button; click the green [Primary] (control point); double-click the green [Primary] | Route planning was only achieved. | The double-click action could not be completed (a key issue for next-step resolution). A slower mouse speed was found incapable of resolving the double-click issue; despite being mentioned in the paper, double-click functionality remained inoperable. | 10 | The paper referenced left-click, right-click, and hotkeys. Consequently, an .ahk file was utilized to implement F1 as a substitute for left-click double-click. |
| 1 | Attempted implementation of movement for friendly targets | Click the 0000 Heavy Ground Unit; click the [Maneuver] button; click the green [Primary] (control point); press the F1 key at the green [Primary] location | Arrival at the green [Primary] location was achieved. | The unit arrived in the vicinity of the green flag, but not specifically on the hexagonal grid cell containing the flag's base. | 10 | Can specific numeric coordinates be reached? Alternatively, how can the flag denoting the primary control point be described clearly? |
| 2 | Implementation of precise clicking on the primary Key Point control point (Modified for more precise description) | A green flag labeled "Primary" is present on the map; please click the hexagonal grid cell where the flagpole is located. | Precise positioning of the primary control point location was achieved. | Integration with actual combat operations is required. Prior to arrival at the location, clicking the "Auto Capture" button should be explicitly instructed. | 5 | Performance remained unstable. |
| 3 | Implementation of stable Key Point control | Step 1: 1. Click the "0002 Team A Combat Vehicle" icon. 2. Click the icon labeled | Precise capture was achievable here. | However, a pause was required, necessitating the introduction of granular output, specifically the | 5 | Stable |

| | | [Maneuver] in the lower right. 3. Click the icon labeled [Auto Capture] on the same horizontal level. Pause Step 2: 1. Click [Secondary]. 2. Press the hotkey F1. | | segmented input of the large model's sequential output to UI-TARS. | | |
|---|---|---|---|---|---|---|
| Horizontal comparison of the seizure exploration process (which originally included two processes, seizure and attack, was broken down into steps on April 23, after which only mouse command exploration for seizure was conducted in this part, and finally the suitable ones were moved to the top). ||||||| 
| 4 | Click the overlapping Team A unit; click to select [0002 Team A Vehicle]. Click the [Auto Key Point Control] button. Click the [Maneuver] button. There is a green flag labeled "Primary" on the map, please click the hexagonal grid cell where the flagpole is located. Click 0717. Click the F1 hotkey. Click [0002 Team A Vehicle]. Click the [Locked Attack] button. Click the Team B operator. | 1.The "0002 Team A Combat Vehicle" icon is clicked. 2.The "Auto Capture" option in the lower right is clicked. 3.The "Maneuver" option is clicked. 4.A green flag labeled "Primary" on the map is identified, and the hexagonal grid cell containing its flagpole is clicked. 5."0717" is clicked. 6.The F1 hotkey is pressed. 7.The "0002 Team A Combat Vehicle" icon is clicked again. 8."Locked Attack" is clicked, followed by a click on the blue icon. 9.The actions of | 1.The "0002 Team A Combat Vehicle" icon is clicked. 2.The "Maneuver" option is clicked, followed by clicking the "Attack" option. 3.A green flag labeled "Primary" on the map is identified, and the hexagonal grid cell containing its flagpole is clicked. 4."0717" is clicked. The F1 hotkey is pressed. 5.The "0002 Team A Combat Vehicle" icon is clicked again. 6."Locked Attack" is clicked, followed by a click on the blue icon. 7.The actions of clicking "Locked Attack" and | 1.The "0002 Team A Combat Vehicle" icon is clicked. 2.The "Maneuver" option is clicked. 3.A green flag labeled "Primary" on the map is identified, and the hexagonal grid cell containing its flagpole is clicked. 4."0717" is clicked. 5.The F1 hotkey is pressed. 6.The "0002 Team A Combat Vehicle" icon is clicked again. 7."Locked Attack" is clicked, followed by a click on the blue icon. 8.The actions of clicking "Locked Attack" and clicking the blue icon are repeated. | 1.The "0001 Team A Combat Vehicle" icon is clicked. 2.The icon labeled "Maneuver" is clicked. | 1.The "0001 Team A Combat Vehicle" icon is clicked. 2.The icon labeled "Maneuver" is clicked. 3.A green flag labeled "Primary" on the map is identified, and the hexagonal grid cell containing its flagpole is clicked. |
|  | Repeat clicking the [Locked Attack] button and clicking the Team B operator. | clicking "Locked Attack" and clicking the blue icon are repeated. | clicking the blue icon are repeated. |  |  |  |
| 5 | 1.The "0001 Team A combat vehicle" icon is clicked. 2.The icon labeled "Maneuver" is clicked. 3.A green flag labeled "Primary" on the map is identified, and the hexagonal grid cell containing its flagpole is clicked. 4."0717" is clicked. | Clicks are performed on 0715, 0616, 0917, and 1818. (It was observed that accuracy improved with local zoom-in (2x)! Under constant screen magnification, combat vehicles were moved towards the center.) | 1.The "0001 Team A Combat Vehicle" icon is clicked. 2.The icon labeled "Seizure" is clicked. 3.The icon labeled "Maneuver" is clicked. 4.A green flag labeled "Secondary" on the map is identified, and the hexagonal grid cell containing its flagpole is clicked. 5."0717" is clicked. 6.The F1 hotkey is pressed. | Key Issue: Execution can absolutely be performed step-by-step. This issue is critically important! | The separation of combat and seizure on April 23, 2025, was a crucial step. |  |
| 6 | 1.The "0002 Team A combat vehicle" icon is clicked. 2.The icon labeled "Maneuver" is clicked. 3."0717" is clicked. 4.The F1 hotkey is pressed. 5.The "0002 Team A combat vehicle" icon is clicked. 6.The icon | 1.The "0002" icon is clicked. 2.The icon labeled "Maneuver" is clicked. 3.The "Secondary" option is clicked. 4.The F1 hotkey is pressed. (A comparison was made between Bing and XingYuan browsers; the XingYuan browser icons were larger.) | 1.The "0002 Team A Combat Vehicle" icon is clicked. 2.The icon labeled "[Maneuver]" in the lower right is clicked. 3."0910" is clicked. 4.The F1 hotkey is pressed. | 1.The "0002 Team A Combat Vehicle" icon is clicked. 2.The icon labeled "[Maneuver]" in the lower right is clicked. 3."0910" is clicked. 4.The F1 hotkey is pressed. | 1.The "0002 Team A Combat Vehicle" icon is clicked. 2.The icon labeled "[Maneuver]" in the lower right is clicked. 3."0909" is clicked. 4.The F1 key on the keyboard is pressed. | 1.The "0001 Team A Combat Vehicle" icon is clicked. 2.The icon labeled "[Maneuver]" in the lower right is clicked. 3."0909" is clicked. 4.A double-click is performed. |
|   | labeled "Attack" is clicked. 7.The Team B combat vehicle icon is clicked, and "0003" is selected and clicked. |   |   |   |   |   |
| 7 | 1.The "0001 Team A combat vehicle" icon is clicked. 2.The icon labeled "[Maneuver]" in the lower right is clicked. 3.The cursor is positioned over the surface of "0909", and the F1 hotkey is pressed. | 1.The "0001 Team A Combat Vehicle" icon is clicked. 2.The icon labeled "[Maneuver]" in the lower right is clicked. 3.The cursor is moved over the surface of "0909", and the F1 hotkey is pressed. | 1.The "0001 Team A Combat Vehicle" icon is clicked. 2.The icon labeled "[Maneuver]" in the lower right is clicked. 3."0909" is clicked. 4.A double-click is performed. (Success here was attributed to a mouse replacement; changing the mouse unexpectedly enabled successful double-click execution!) | 1.The "0001 Team A Combat Vehicle" icon is clicked. 2.The icon labeled "[Control]" in the lower right is clicked. 3.The icon labeled "[Maneuver]" on the same horizontal level is clicked. 4."0708" is clicked. 5.A double-click is performed. | 1.The "0001 Team A Combat Vehicle" icon is clicked. 2.The icon labeled "[Control]" in the lower right is clicked. 3.The icon labeled "[Maneuver]" on the same horizontal level is clicked. 4."0715" is clicked, followed by a double-click. | 1.The "0001 Team A Combat Vehicle" icon is clicked. 2.The icon labeled "[Control]" in the lower right is clicked. 3.The icon labeled "[Maneuver]" on the same horizontal level is clicked. 4.A double-click is performed on "0715" (double-clicking was also unstable). |
| 8 | 1.The "0001 Team A combat vehicle" icon is clicked. 2.The icon labeled "[Control]" in the lower right is clicked. 3.The icon | 1.The "0002 Team A Combat Vehicle" icon is clicked. 2.The "Attack" option is clicked. 3.The "Team B Combat Vehicle" is clicked. (Currently operational) | Step 1: 1.The "0147 Team A Combat Vehicle" icon is clicked. 2.The icon labeled "[Maneuver]" in the lower right is clicked. Step 2: 3."5079" is clicked. | 1.The "0147 Team A Combat Vehicle" icon is clicked. 2.The button labeled "[Move]" in the lower right is clicked. 3."5079" is clicked. 4.The F1 hotkey is pressed (Standard: Action: hotkey(f1)). | 1.The "0002 Team A Combat Vehicle" icon is clicked. 2.The icon labeled "[Maneuver]" in the | Step 1: 1.The "0002 Team A Combat Vehicle" icon is clicked. 2.The icon labeled "[Maneuver]" in the lower |

| | | labeled "[Maneuver]" on the same horizontal level is clicked. 4."0708" is clicked. 5.The F1 hotkey is pressed. | | 4.The F1 hotkey is pressed. | | lower right is clicked. 3.The icon labeled "[Auto Capture]" on the same horizontal level is clicked. Pause: 4.The "[Secondary]" option is clicked. 5.The F1 hotkey is pressed. | right is clicked. 3.The icon labeled "[Auto Capture]" on the same horizontal level is clicked. Pause Step 2: 4.The "[Secondary]" option is clicked. 5.The F1 hotkey is pressed. |

As indicated in the table above, after unsuccessful attempts with multiple sequential commands, an approach utilizing a single command for one minor task per execution was initiated on April 23, 2025.

## G3 GamePlatform A Platform Instruction Experiment

Due to the scenario distinction between aircraft and Ground Units on the GamePlatform A platform, the command experiment was divided into two parts.

**Table G3**.GamePlatform A Platform (Aircraft) Command Experiment Table

| Exp. ID. | Experimental Objective | Mouse Language | Capability | Defect | Success Count / Number of Exp. | Summary |
|---|---|---|---|---|---|---|
| 0 | Achievement of attacks on Air Unit aircraft is implemented. | The red [Air Unit aircraft] is clicked; the [force control] above is clicked; the [attack selection] is clicked; the [manual engagement] is clicked; the cyan [Air Unit aircraft] is selected and [AIR WEAPON] is clicked; the up arrow behind the [tool selection for the chosen target] button is clicked twice; the [tool selection for the chosen target] is clicked; the "x" in the upper right corner is clicked to close the window. | corner of the tool allocation interface is clicked to close the window. Clicking of the red [Air Unit aircraft], clicking of the [force control] above, clicking of the [attack selection], clicking of the [manual engagement], selection of the cyan [Air Unit aircraft] and clicking of [AIR WEAPON] are achieved. | The up arrow behind the [tool selection for the chosen target] button is clicked twice. This step cannot be implemented. The up arrow is too small and consistently fails to be clicked. The number "2" is written in the space after [tool selection for the chosen | 0/5 | Independent study of writing in the space is conducted. |
| | | | | target]. | | |
| 1 | Entry into the input box is achieved. | The "0" in the space behind the [tool selection for the chosen target] button is rewritten as "2". | Positioning can be achieved. | The span is too large to achieve. | 0/4 | The process of rewriting numbers can actually be decomposed into three steps: selection, deletion, and input. |
| | | The default number "0" in the input box behind the [tool selection for the chosen target] button is deleted, and the number "2" is entered. | The button is always selected instead of the input box behind it. | The defect lies in the positional description phrase "behind". | 0/3 | "Behind" should be changed to "to the right of". |
| | | The default number "0" in the input box to the right of the [tool selection for the chosen target] button is deleted, and the number "2" is entered. | The input box is selected. | The number "0" is not selected and rewritten. | 2/8 | Because selection of the input box does not determine whether the cursor is positioned before or after "0", Hotkey BackSpace or Hotkey Delete should be pressed after selection. |
| 2 | Selection of the input box and input of numbers are achieved. | The "0" in the input box to the right of the [tool selection for the chosen target] button is selected, the Hotkey BackSpace or Hotkey Delete is pressed, and the number "2" is entered. | Implementation cannot be achieved. | The first half of a sentence is always executed, selecting the input box to the right of the [tool selection for the chosen target] button. | 0/3 | "Do not click the [tool selection for the chosen target] button" is added at the front. |
| | | The [tool selection for the chosen target] button is not clicked; the "0" in the input box to the right of the [tool selection for the chosen target] button is selected, the Hotkey BackSpace or Hotkey Delete is pressed, and the number "2" is entered. | | It is not an inability to distinguish between the button and the input box, but rather that clicking occurs based on the number of characters matching the target description at the beginning, without regard to the | 2/7 | In reality, there is only this one input box in the interface. Perhaps less positioning information is needed, directly stating "delete the 0 in the input box and enter 2" or "rewrite the number 0 in the input box to number 2". |
| | | | | subsequent description. | | |
| | | The [tool selection for the chosen target] button is not clicked; the area to the right of the "0" in the input box to the right of the [tool selection for the chosen target] button is clicked, the Hotkey BackSpace or Hotkey Delete is pressed, and the number "2" is entered. | | | 1/7 | |
| | | The "0" in the input box is deleted, and "2" is entered. | | Positioning error; the front button is pressed. | 0/6 | |
| | | The input box to the right of the number "0" is clicked. | | Precise positioning is achieved! | 1/8 | |
| | | The input box to the right of the number "0" is clicked, and the "0" is changed to the number "2". | | | 0/8 | |
| | | The cursor is positioned to the right of the number "0" and clicked. | | Three times it was placed to the left of "0". | 0/3 | |
| | | The cursor is positioned to the right of the number "0" and clicked, the Hotkey Delete is pressed, and the number "2" is entered. | | The cursor was positioned to the right, and the Delete key could not complete the intended action. | 0/4 | |
| | | The cursor is positioned to the right of the number "0" and clicked, the Hotkey Delete is pressed, the Hotkey BackSpace is pressed, and the number "2" is entered. | | Placing the Hotkey Delete first, when positioning is to the right of "0"; placing the Hotkey BackSpace first, when positioning is to the left of "0". | 1/5 | |
| | | The cursor is positioned in the input box to the right of "0", the key combination Ctrl+A is pressed, and the number "2" is entered. | | Prolonged rotation with no response, but success can be achieved. | 2/5 | Extreme instability is observed. |
| 3 | Double-clicking | 1. The red [Air Unit aircraft] is clicked. 2. The [force control] | | Individual steps can all be implemented. | 1/7 | Modifying numbers and clicking the up |

| | | above is clicked. 3. The [attack selection] is clicked. 4. The [manual engagement] is clicked. 5. The cyan [Air Unit aircraft] is clicked. 6. The AIR WEAPON is clicked. 7. The AIR WEAPON is double-clicked. 8. The AIR WEAPON is double-clicked. 9. The "x" in the upper right corner of the tool allocation interface is clicked to close the window. | | Multi-step experiments are scheduled for tomorrow morning (20250512). | | arrow determine the quantity first, then volley fire. Now, double-clicking can be used to confirm firing one round first, then another round. But the steps are still too many; hotkeys should be used for greater convenience. |
|---|---|---|---|---|---|---|
| 4 | Manual tool allocation via Shift+F1 is achieved. | 1. The red [Air Unit aircraft] is clicked. 2. The hotkey Shift+F1 is pressed. 3. The cyan [Air Unit aircraft] is clicked. 4. The AIR WEAPON is clicked. 5. The AIR WEAPON is double-clicked. 6. The AIR WEAPON is double-clicked. 7. The "x" in the upper right corner of the tool allocation interface is clicked to close the window. | The entire workflow can be successfully executed, achieving resource allocation via Shift+F1. | 1.Clicks sometimes land on PL-10, still due to clicking upon seeing the letters 'PL'. Descriptions need precision. 2. Clicks landed on the red GamePlatform A icon more than three times. | 2/8 | 1. The resource click is changed to 12C; 2. The Air Unit aircraft name is added before the Air Unit aircraft icon. |
| | | 1. The red AIR UNIT B [Air Unit aircraft] icon is clicked. 2. The hotkey (Shift+F1) is pressed. 3. The cyan AIR UNIT A [Air Unit aircraft] icon is clicked. 4. The 12C is clicked. 5. The 12C is double-clicked. 6. The 12C is double-clicked. 7. The "x" in the upper right corner of the tool allocation interface is clicked to close the window. | The entire workflow can be successfully executed, achieving resource allocation via Shift+F1. | 1.The resource selection part still has inaccuracies at times, with clicks only occurring in the general area of the required resource, leading to successful allocation. 2. After Shift+F1 is pressed, it is easily lost, causing clicks on AIR UNIT A to become ordinary clicks. | 5/10 | 1. This issue is a warning; 2. This problem is attempted to be resolved without using hotkeys. |
| 5 | Double-clicking for resource | 1. The red AIR UNIT B [Air Unit aircraft] icon is clicked. 2. The [force control] above is clicked. 3. The [attack selection] | | Clicking [manual engagement] also has the aforementioned | 1/3 | Implementation can be achieved. |

|  | allocation is implemented, rather than reliance on number modification. | is clicked. 4. The [manual engagement] is clicked. 5. The cyan AIR UNIT A [Air Unit aircraft] icon is clicked. 6. The AIR WEAPON is clicked. 7. The AIR WEAPON is double-clicked. 8. The AIR WEAPON is double-clicked. 9. The "x" in the upper right corner of the tool allocation interface is clicked to close the window. |  | problem. The selection target's [+] becomes an ordinary click. Multi-step processes can be implemented. |  |  |

Following the successful implementation for aircraft, the command flow for aircraft was replicated within the Ground Unit scenario.

**Table G4**.GamePlatform A Platform (Ground Unit) Command Experiment Table

| Exp. ID. | Experimental Objective | Mouse Language | Capability | Defect | Success Count / Number of Exp. | Summary |
| --- | --- | --- | --- | --- | --- | --- |
| 0 | Hotkey-based implementation of resource allocation for Air Unit aircraft targets. | 1.The red AIR UNIT B [Air Unit aircraft] icon is clicked. 2.The hotkey (Shift+F1) is pressed. 3.The cyan AIR UNIT A [Air Unit aircraft] icon is clicked. 4."12C" is clicked. 5."12C" is double-clicked. 6."12C" is double-clicked. 7.The "x" button in the upper right corner of the tool allocation interface is clicked to close the window. | Achievement of resource allocation is implemented. | Unreliable performance is occasionally encountered. |  | The process is modeled after the Air Unit aircraft resource allocation procedure. |
| 1 | Hotkey-based implementation of resource allocation for Ground Unit targets. | 1. The red [Ground Unit] icon is clicked. 2.The hotkey Shift+F1 is pressed. 3.The blue [Ground Unit] icon is clicked. 4."Ground Weapon" is clicked. 5."Ground Weapon" is double-clicked. 6."Ground Weapon" is double-clicked. 7.The "x" button in the upper right corner of the tool allocation interface is clicked to close the window. | Achievement of resource allocation is implemented. | Unreliable performance is occasionally encountered. |  |  |
| 2 | Manual button-based implementation | 1. The red AIR UNIT B [Air Unit aircraft] icon is clicked. 2.The [Force Control] above is | Achievement of resource allocation is | Unreliable performance is |  | The process is modeled |

| | | of resource allocation for Air Unit aircraft targets. | clicked. 3.The [Attack Selection] is clicked. 4.The [Manual Engagement] is clicked. 5.The cyan AIR UNIT A [Air Unit aircraft] icon is clicked. 6.The "AIR WEAPON" is clicked. 7.The "AIR WEAPON" is double-clicked. 8.The "AIR WEAPON" is double-clicked. 9.The "x" button in the upper right corner of the tool allocation interface is clicked to close the window. | implemented. | occasionally encountered. | | after the Air Unit aircraft resource allocation procedure. |
|---|---|---|---|---|---|---|---|
| 3 | | Manual button-based implementation of resource allocation for Ground Unit targets. | 1. The red [Ground Unit] icon is clicked. 2.The [Force Control] above is clicked. 3.The [Attack Selection] is clicked. 4.The [Manual Engagement] is clicked. 5.The blue [Ground Unit] icon is clicked. 6.The "Ground Weapon" is clicked. 7.The "Ground Weapon" is double-clicked. 8.The "Ground Weapon" is double-clicked. 9.The "x" button in the upper right corner of the tool allocation interface is clicked to close the window. | Achievement of resource allocation is implemented. | Unreliable performance is occasionally encountered. | | |

## Appendix H Experiment to Improve Execution Mechanism Precision

To enable the large model to possess "memory" capability and action continuity, we attempted to design the MI2.7.1 dataset, which places two consecutive images within one sample (e.g., images 1 and 2 in one sample, images 2 and 3 in the next sample, etc.). Based on this version, we formed datasets MI2.7.2, MI2.7.3, MI2.7.4, and MI2.7.5 to verify the impact of this sliding sample on large model fine-tuning (since a single task involves at least 6 images, the maximum sliding range is 6).

The composition of the sliding datasets is shown in the figure below:

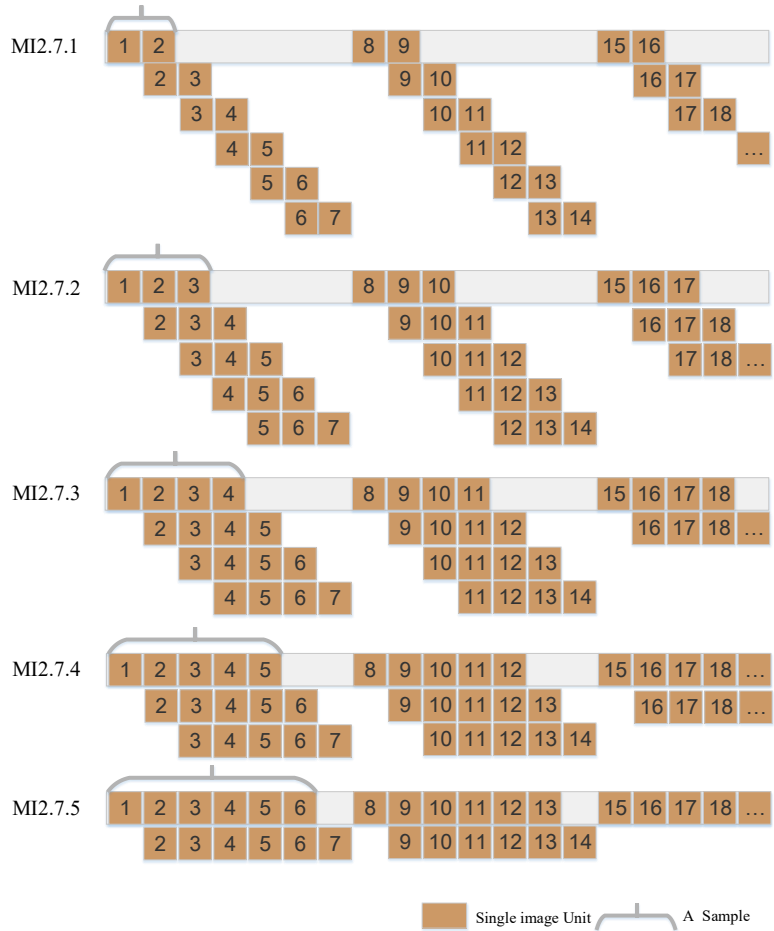

**Fig. H1**. Composition of Various Sliding Datasets. Khaki blocks represent single images.

**Table H1**. Performance of Various Sliding Datasets on Different Test Sets

| Metric | annotions_new2.1 | MI2.7.1 | MI2.7.2 | MI2.7.3 | MI2.7.4 | MI2.7.5 |
|---|---|---|---|---|---|---|
| bleu-4(val2_sum) | 34.82248038 | 37.2441 | 38.806 | 37.8204 | 34.7305 | 36.4469 |
| model_preparation_time(val2_sum) | 0.0094 | 0.0107 | 0.0071 | 0.0068 | 0.0073 | 0.0071 |
| rouge-1(val2_sum) | 51.49984 | 54.1944 | 56.582 | 53.9866 | 52.6548 | 55.0083 |
| rouge-2(val2_sum) | 37.64772075 | 39.5239 | 42.0351 | 41.2433 | 37.7835 | 40.1844 |
| rouge-l(val2_sum) | 48.38567547 | 51.1186 | 52.8173 | 50.5742 | 48.9572 | 51.3882 |
| runtime(val2_sum) | 729.108 | 749.8523 | 865.9963 | 351.8478 | 336.105 | 686.4905 |
| samples_per_second(val2_sum) | 0.363 | 0.353 | 0.306 | 0.753 | 0.788 | 0.386 |
| steps_per_second(val2_sum) | 0.182 | 0.177 | 0.154 | 0.378 | 0.396 | 0.194 |
| bleu-4(val2_S) | 45.84147612 | 48.9711 | 50.833 | 49.6123 | 46.0482 | 47.8819 |
| model_preparation_time(val2_S) | 0.011 | 0.0116 | 0.007 | 0.0071 | 0.0069 | 0.0074 |
| rouge-1(val2_S) | 60.33670448 | 63.947 | 64.3996 | 62.4827 | 60.8749 | 63.4773 |
| rouge-2(val2_S) | 46.71890498 | 49.1035 | 50.0518 | 50.3317 | 46.2407 | 48.5558 |
| rouge-l(val2_S) | 60.12011642 | 63.5776 | 64.2589 | 62.1727 | 60.4112 | 63.0456 |
| runtime(val2_S) | 177.7917 | 247.7992 | 333.2427 | 104.5855 | 99.8048 | 254.958 |
| samples_per_second(val2_S) | 1.131 | 0.811 | 0.603 | 1.922 | 2.014 | 0.788 |
| steps_per_second(val2_S) | 0.568 | 0.408 | 0.303 | 0.966 | 1.012 | 0.396 |

From the above table, performance indicator change line charts for each dataset can be generated.

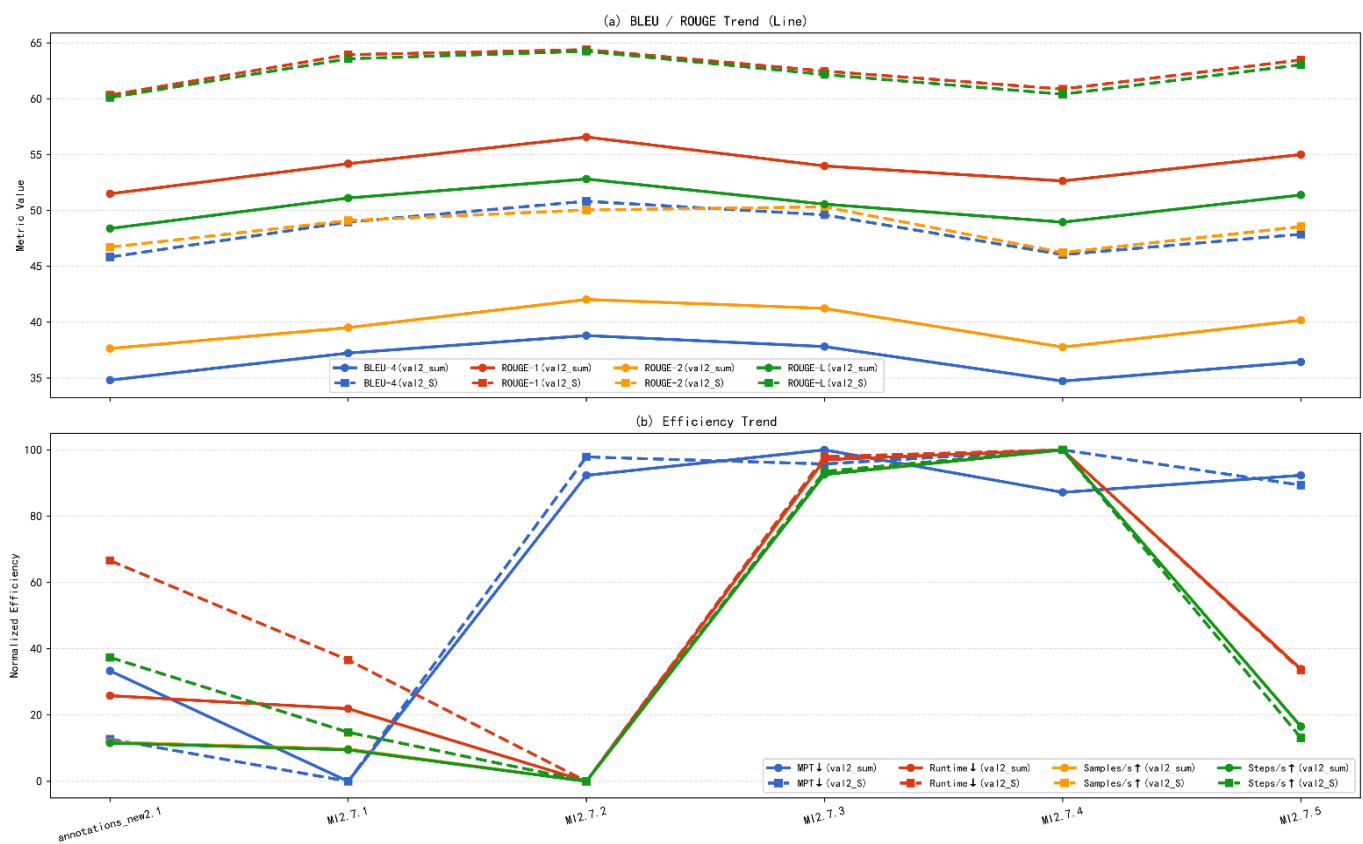

**Fig. H2**.Line Charts of Various Performance Metrics vs. Sliding Variation. The red Samples/s line in chart b is obscured by the purple Steps/s line and is not fully displayed.

From the above table, we know:

(a) Trade-off between quality and efficiency. Sliding 2 has the best quality in both test sets, but long runtime and low throughput; sliding 3 has quality slightly lower than sliding 2 but significantly better efficiency. Sliding 4 and 5 have the highest efficiency but a noticeable drop in quality.

(b) Optimal sliding range. When prioritizing quality, sliding 2 is the overall best quality choice; if sacrificing some quality for high-efficiency inference, sliding 3 is a more balanced choice.

(c) Excessive sliding range may have negative effects. Sliding 4 and 5 may introduce too much redundant or distant context, weakening the model's attention to adjacent frames' effective temporal signals, leading to quality degradation.

(d) Base dataset vs. sliding mode

All sliding versions (especially windows 2 and 3) outperform the initial single image annotions_new2.1 in quality, proving that placing previous and next frames within a sample enhances the model's temporal perception.

Conclusion:

(a) Best quality: MI2.7.2 (sliding window=2), performs outstandingly in BLEU-4 and ROUGE metrics, the preferred choice for precision-critical tasks.

(b) Best quality-efficiency trade-off: MI2.7.3 (sliding window=3), quality close to the best version, efficiency significantly better than sliding 2, suitable for scenarios requiring both quality and speed.

(c) Best efficiency: MI2.7.4 and MI2.7.5, significantly better inference speed and throughput than other versions, but noticeable quality drop, only suitable for speed-sensitive tasks where precision can be sacrificed.

(d) Excessive sliding range is detrimental: Too large a range (≥4) introduces more non-critical context, diluting the model's attention to current relevant frames.

# Appendix I Catastrophic Forgetting and Statistical Significance Testing

## I1 Catastrophic Forgetting Experiment

Initially, this study planned to perform sequential fine-tuning. To compare the impact of single-modality datasets and two-modality datasets on fine-tuning effectiveness, we later discovered catastrophic forgetting in the large model. Below are the records and verification of catastrophic forgetting.

Table I1. Catastrophic Forgetting Experiment Test Set Record

| Test_Set | Fine-Tuning Epochs | Test Result Files | Dataset | Model Files |
|---|---|---|---|---|
| val_S | 0 | valZN1 | Qwen2.5vl Base Model | None |
| val_V1 | 0 | valZN2 | Qwen2.5vl Base Model | None |
| val_S | 3 | valZN3 | annotions_new | train_2025-06-23-16-24-32 |
| val_V1 | 3 | valZN4 | my_video_data | train_2025-06-25-20-48-06 |
| val_S | 3+3 | valZN5 | annotions_new+my_video_data | train_2025-06-29-13-58-58 |
| val_V1 | 3+3 | valZN6 | annotions_new+my_video_data | train_2025-06-29-13-58-58 |
| val_S | 3+3 | valZN7 | my_video_data+annotions_new | train_2025-06-30-10-39-47 |
| val_V1 | 3+3 | valZN8 | my_video_data+annotions_new | train_2025-06-30-10-39-47 |

Forming the following indicator tables:

Table I2. Catastrophic Forgetting Experiment Indicator Record Table

| Metric | valZN1 | valZN2 | valZN3 | valZN4 | valZN5 | valZN6 | valZN7 | valZN8 |
|---|---|---|---|---|---|---|---|---|
| bleu-4 | 0.5661 | 0.8693 | 52.3385 | 0.9622 | 40.9341 | 33.1127 | 61.2251 | 0.7313 |
| model_preparation_time | 0.0176 | 0.0015 | 0.007 | 0.0073 | 0.0095 | 0.0073 | 0.0074 | 0.0093 |
| rouge-1 | 5.4935 | 10.4412 | 65.8618 | 14.0627 | 61.8618 | 64.6635 | 73.5454 | 29.6142 |
| rouge-2 | 0.4303 | 0.3411 | 52.0796 | 0.1417 | 43.7809 | 47.8489 | 63.3879 | 13.8833 |
| rouge-l | 2.2712 | 7.4627 | 65.6446 | 10.3493 | 53.834 | 51.1779 | 73.1383 | 15.3133 |
| runtime | 726.1088 | 160.1476 | 103.6174 | 88.4207 | 275.6491 | 83.9525 | 101.6286 | 61.1437 |
| SAM_per_second | 0.277 | 0.2 | 1.94 | 0.362 | 0.729 | 0.381 | 1.978 | 0.523 |
| steps_per_second | 0.139 | 0.2 | 0.975 | 0.362 | 0.366 | 0.381 | 0.994 | 0.523 |

According to the test result files, setting annotions_new as A and my_video_data as B, using bleu-4 as the metric, the following comparison result table is formed:

Table I3. Catastrophic Forgetting Indicator Result Comparison

| Dataset | Base Model (val_S) | Base Model (val_V1) | Annotions_new(val_S) | my_video_data(val_V1) |
|---|---|---|---|---|
| Score Notation | $S_0(A)$ | $S_0(B)$ | $S_1(A)$ | $S_2(B)$ |
| bleu-4 | 0.5661 | 0.8693 | 61.2251 | 0.9622 |
| Dataset | annotions_new+ my_video_data(val_S) | annotions_new+ my_video_data(val_V1) | my_video_data+ annotions_new(val_S) | my_video_data+ annotions_new(val_V1) |
| Score Notation | $S_3(A)$ | $S_3(B)$ | $S_4(A)$ | $S_4(B)$ |
| bleu-4 | 40.9341 | 33.1127 | 61.2251 | 0.7313 |

From the table, we know:

Only A trained: $S_1(A) = 52.3385$; A then B trained: $S_3(A) = 40.9341$

Absolute drop: $52.3385 - 40.9341 = 11.4044$; Relative retention rate: $40.9341/52.3385 \approx 78.2\%$.

Only B trained: $S_2(B) = 0.9622$; B then A trained: $S_4(B) = 0.7313$.

Absolute drop: $0.9622 - 0.7313 = 0.2309$; Relative retention rate: $0.7313/0.9622 \approx 76.0\%$.

Therefore, we can conclude:

(a) A task retention rate 78%, B task retention rate 76%, both significantly decreased (below 80% can be considered "significant forgetting").

(b) Catastrophic forgetting is confirmed, and the forgetting degree for tasks A and B is comparable.

(c) Sequential fine-tuning is not suitable for the current task; combined datasets should be used for VLM fine-tuning.

## I2 Statistical Validation of the Performance Decline (PD) Metric

To validate the statistical significance of the PD values reported in Table 10, we performed 10 independent inference runs for the M*V+S strategy on the val_S test set. In each run, only the random seed for the generative process was varied; all model weights, decoding parameters (temperature=0.7, top_p=0.9), and the test set remained fixed. The resulting BLEU-4 scores were:

**Table I4** Independent Reasoning Experiment Statistical Table

| Exp.NO. | BLEU-4 | Exp.NO. | BLEU-4 |
|---|---|---|---|
| 1 | 62.3642 | 6 | 63.4986 |
| 2 | 62.4146 | 7 | 62.6131 |
| 3 | 65.0266 | 8 | 62.9510 |
| 4 | 61.9159 | 9 | 62.5827 |
| 5 | 62.8407 | 10 | 62.6948 |

From these, we computed a mean score of 62.89 with a standard deviation of 0.93. Using the full-fusion baseline score of 4.81 as the null hypothesis, a one-sample t-test yielded a t-statistic of 198.56 with 9 degrees of freedom, corresponding to a p-value < 0.0001. This confirms that the massive performance gain (and the associated PD of approximately -1200%) is a statistically robust finding.